\documentclass[journal]{article}

\usepackage{cite}
   \usepackage[pdftex]{graphicx}

\usepackage{tikz,pgfplots}
\usetikzlibrary{spy}

\usetikzlibrary{calc}
\usepackage[top=70pt,bottom=90pt,left=90pt,right=90pt]{geometry}

\usepackage{amsmath}

\usepackage{algorithmic}

\usepackage{array}
\usepackage{multirow}
\usepackage[caption=false,font=footnotesize]{subfig}
\usepackage{amsmath,amssymb}
\usepackage{booktabs}
\usepackage{tabularx}

\usepackage{xspace}

\newcommand{\aquasi}{AQuaSI\xspace}
\newcommand{\eg}{e.\,g.\xspace}
\newcommand{\ie}{i.\,e.\xspace}
\newcommand{\wrt}{w.\,r.\,t.\xspace}
\newcommand{\etal}{\textit{et\,al.}\xspace}
\DeclareMathOperator*{\argmin}{argmin}
\DeclareMathOperator*{\argmax}{argmax}
\newcommand{\sign}[1]{\mathrm{sign}(#1)}
\newcolumntype{R}[1]{>{\raggedleft\arraybackslash\hspace{0pt}}p{#1}}
\newcolumntype{L}[1]{>{\raggedright\arraybackslash\hspace{0pt}}p{#1}}
\newcolumntype{C}[1]{>{\centering\hspace{0pt}}p{#1}}
\newlength \figureheight
\newlength \figurewidth
\renewcommand{\eqref}[1]{Eq.~\ref{#1}}
\usepackage{authblk}
\usepackage[title]{appendix}

\renewcommand{\vec}[1]{\ensuremath{\mathbf{#1}}}
\usepackage{graphbox} 

\usepackage{hyperref}
\begin{document}

\title{Adaptive Quantile Sparse Image (AQuaSI) Prior for Inverse Imaging Problems}

\author[1]{Franziska~Schirrmacher}
\author[2]{ Christian Riess}
\author[1]{ Thomas K\"ohler}

\affil[1]{Pattern Recognition Lab, Friedrich-Alexander-University (FAU) Erlangen-N\"urnberg, Erlangen 91058, Germany.  E-mail:\{franziska.schirrmacher, thomas.koehler\}@fau.de} \affil[2]{IT Security Infrastructures Lab, FAU Erlangen-N\"urnberg, Erlangen 91058, Germany.}

\date{}


\maketitle

\begin{abstract}
Inverse problems play a central role for many classical computer vision and
image processing tasks. Many inverse problems are ill-posed, and hence require
a prior to regularize the solution space. However, many of the existing priors,
like total variation, are based on ad-hoc assumptions that have difficulties to
represent the actual distribution of natural images. Thus, a key challenge in
research on image processing is to find better suited priors to represent
natural images.

In this work, we propose the Adaptive Quantile Sparse Image (AQuaSI) prior. It
is based on a quantile filter, can be used as a joint filter on guidance data,
and be readily plugged into a wide range of numerical optimization algorithms.
We demonstrate the efficacy of the proposed prior in joint RGB/depth upsampling,
on RGB/NIR image restoration, and in a comparison with related
regularization by denoising approaches.
\end{abstract}

%
%
%

%

\section{Introduction}
\label{sec:Introduction}

In computer vision and image processing, many tasks are modeled as inverse problems ranging from low-level vision to image analysis and scene understanding. Inverse problems refer to the inference of latent information from noisy observations, whereas both quantities are linked by a generative model. For instance, in image denoising, we are interested in obtaining clean images from noisy ones under a certain noise model. Despite widespread application domains, such formulations can be tackled from 
two complementary perspectives: \textit{Model-based}
approaches aim at inverting the generative model using iterative optimization. 
This includes classical Bayesian statistics~\cite{Babacan2011} or
the most recent compressive sensing methodologies~\cite{Ochs2013}.
\textit{Learning-based} approaches infer mappings in forward direction by
learning from exemplars. Here, dictionary-based methods~\cite{Elad2006}
constitute the classical approach, while more recently deep neural network
pushed the frontiers towards end-to-end learning, for example in
super-resolution~\cite{Kim2016} or motion deblurring~\cite{Wieschollek2017}.

An inverse problem is
oftentimes highly \textit{ill-posed}. This implies that its solution is not
necessarily unique and continuously related to the given observations. 
Turning ill-posed problems to well-posed ones requires \textit{priors} 
on the desired solution spaces. Such prior knowledge can either
be incorporated implicitly via learning from example data, or explicitly by
enforcing properties of a solution. The latter has the merit that it is
exclusively driven by available input data without relying on external training
data. 
It is achieved by a proper regularization of the problem, 
which can be steered by general assumptions like smoothness or sparsity of the 
desired solution. For instance, traditional models like total variation (TV) 
or related methods regularize with \textit{linear} filter operations (\eg, 
gradients or Wavelet transforms) using sparsity-inducing loss functions 
in that filter domain. Such priors are well understood and can be modeled 
in Bayesian form but, despite their flexibility, they 
are based on too simplistic design choices to reasonably constraining the 
solution space. For example, TV priors in image restoration
oftentimes cause staircasing artifacts that are typically not observed in natural images. 

Recently, powerful priors that build on image denoising 
methods have been demonstrated for their performance in inverse imaging. 
Plug-and-play (PnP) \cite{Venkatakrishnan2013} formulations, autoencoding priors 
\cite{Bigdeli2017}, as well as regularization by denoising (RED) \cite{Romano2016a} 
and variants thereof \cite{Sun2019,Mataev2019} are some manifestations 
of this approach.
Specifically, RED allows to relate 
\textit{non-linear} operators with image priors in analytical form. 
However, 
despite its practical success, Reehorst and Schniter \cite{Reehorst2018} showed that 
RED algorithms can only be related to specific priors under denoisers with
symmetric Jacobian. Unfortunately, this property does not hold true for
popular denoisers like BM3D or deep neural nets. Moreover, RED is based on a cross correlation 
between an image and its residual noise estimated by a denoiser, which is assumed 
to be white noise. This model can be translated into Tikhonov-like regularization also known as graph Laplacian in case of symmetric smoothing \cite{Chan2019}.
This is a delicate property as it would require a perfect denoiser to infer the 
residual noise. Current denoising engines can barely infer the residual in form 
of pure white noise, especially if observations are degraded 
by challenging noise types like Poisson or multiplicative models. This can limit the power 
of the prior in many applications.


\begin{figure}[!t]
	
	\centering
	

	\includegraphics[width=0.8\textwidth]{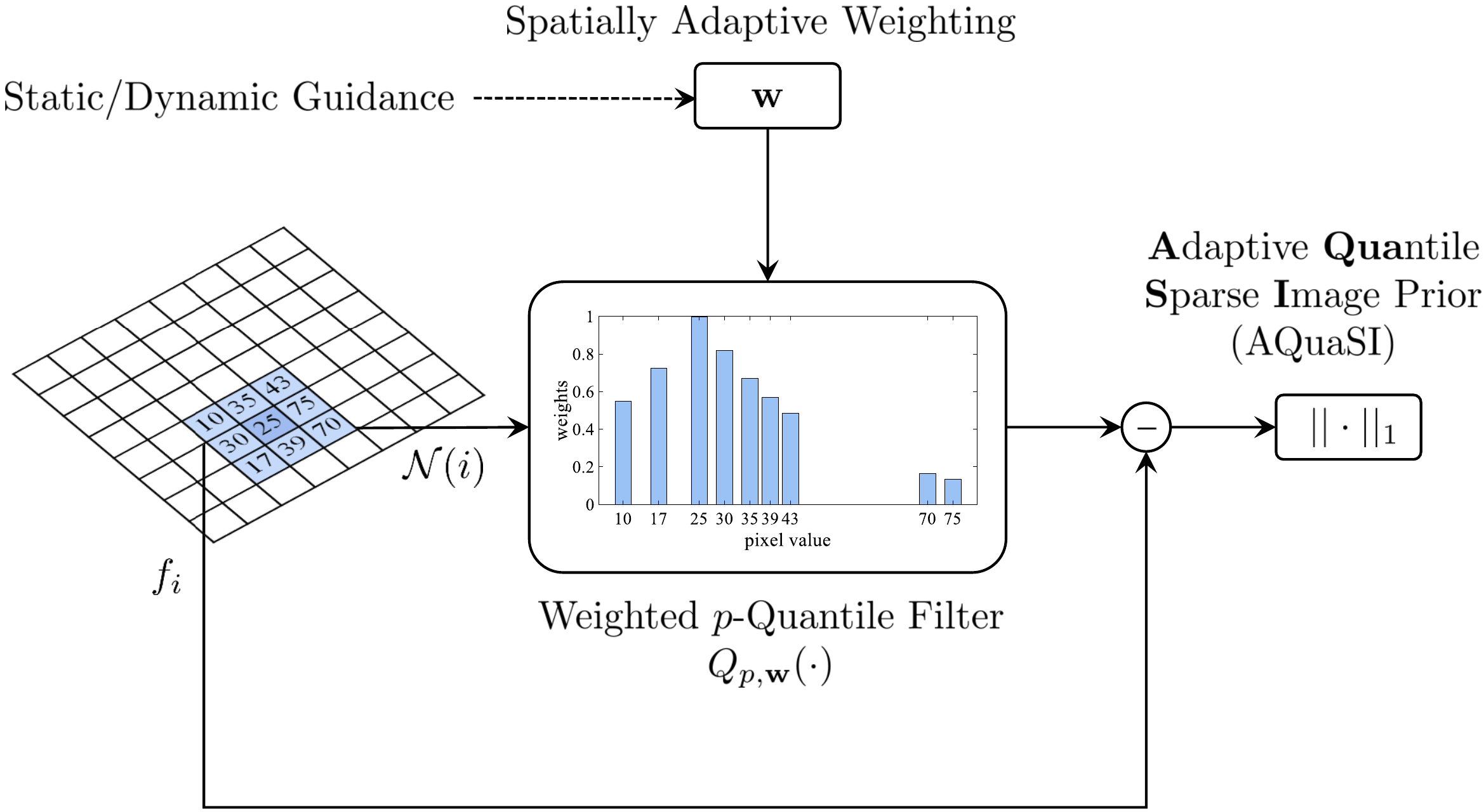} \\
	\subfloat[Noisy input]{\includegraphics[width=0.23\linewidth]{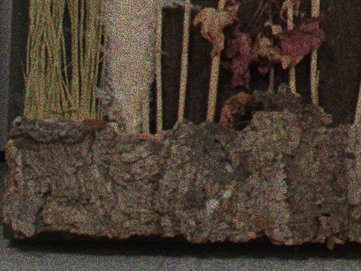}}\
	\subfloat[RED~\cite{Romano2016a} with DnCNN~\cite{Zhang2017beyond}]{\includegraphics[width=0.23\linewidth]{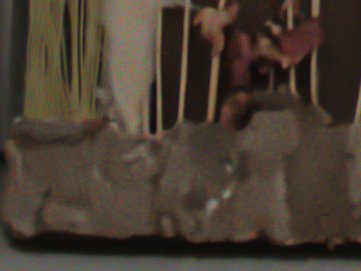}} \
	\subfloat[AQuaSI - \textit{Dynamic} guidance]{\includegraphics[width=0.23\linewidth]{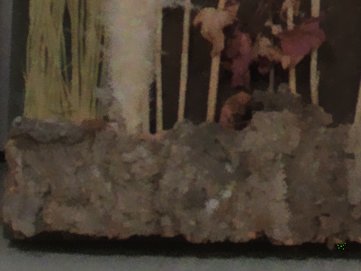}}\
	\subfloat[AQuaSI - \textit{Static} guidance]{\includegraphics[width=0.23\linewidth]{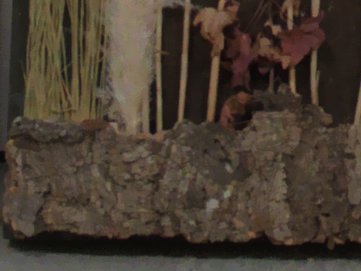}}	
	\caption{Top:Schematic representation of the \aquasi prior. Bottom: Denoising results on an image corrupted by multiplicative speckle noise. The weighted p-quantile filter is applied to an input image $\vec{f}_i$. Then, the $L_1$ norm is applied to the difference of the filtered output and the input image. The filter sorts the pixel values within a predefined neighborhood $\mathcal{N}(i)$ in ascending order. Weights are assigned based on the similarity to the center pixel and the intensity distribution in a guidance image. The guidance can be gained by external data (\textit{static} guidance) or from the input data itself (\textit{dynamic} guidance).}
	\label{fig:aquasi}
\end{figure}

In this paper, we pursue the goal of designing a more appropriate model and
propose the \textbf{A}daptive \textbf{Qua}ntile \textbf{S}parse \textbf{I}mage
(AQuaSI) prior -- a novel prior for ill-posed inverse imaging problems. The proposed 
\aquasi prior falls into the broad class of RED formulations 
with some crucial design choices: 1) We exploit the general class of 
\textit{weighted quantile filters} as a specific denoising type to build our model. Contrary to standard denoisers, such filters are particularly effective for heavy-tailed noise distributions differing from Gaussian models, \eg, impulsive noise \cite{Bovik2010}. In our framework, they can also be linearized, and hence very efficiently deployed for numerical optimization. 2) \aquasi assumes images to be fixed points
under weighted quantile filtering similar to autoencoding priors,
but with \textit{sparsity-promoting regularization} contributing to its robustness. In particular, we use $L_1$ norm regularization for fixed points and can relate this model to a prior distribution in 
the sense of Bayesian statistics.
In contrast, the graph Laplacian in RED is sensitive to denoising inaccuracy of the used filter. While both priors are expected to perform equally under ideal conditions like additive Gaussian noise, our sparse model gains in importance under challenging noise distributions. The advantages of these properties can be seen in Fig.~\ref{fig:aquasi} (left) for multiplicative speckle noise removal. Here, RED with DnCNN \cite{Zhang2017beyond} as denoiser fails to handle this noise distribution and overpenalizes the denoising inaccuracy in its regularization term, resulting in oversmoothing. In contrast, \aquasi with its weighted quantile filter and sparse regularization achieves superior structure preservation.  

The proposed model generalizes our previous paper on the QuaSI prior~\cite{Schirrmacher2018}. This earlier work proposes a
quantile filter as prior. In this work, we make it spatially adaptive 
by introducing filter weights.
This filter class also contains weighted median filtering~\cite{Ma2013,Zhang2014} 
as a special instance. As such, \aquasi enables regularization
via \textit{joint filtering} by calculating the filter weights from \textit{guidance data}. In this case, the pixel values within the filter kernel are sorted and weighted by values derived from the guidance data to calculate quantiles. The spatial weighting can be either derived from an external image, termed \textit{static} guidance, or from the given input image, termed \textit{dynamic} guidance. In contrast, QuaSI exploits uniform weights. A schematic representation of our prior is shown in Fig.~\ref{fig:aquasi} (right). 

We further show that \aquasi is
universally usable and how it can be deployed in different model-based
optimization approaches to solve typical inverse imaging problems\footnote{The
source code to our methods and supplementary material can be downloaded at: \url{https://github.com/franziska-schirrmacher/AQuaSI}}. In different applications, including non-blind image deblurring, joint upsampling of RGB/depth
images, and restoration of RGB/near infrared images, we demonstrate that our
prior improves upon hand-crafted priors that are driven by ad-hoc assumptions. 

This paper is organized as follows. We first mathematically define the inverse
problem and list related work on it in Sec.~\ref{sec:RelatedWork}. The AQuaSI
prior and its numerical optimization are presented in
Sec.~\ref{sec:FormulatingTheAQuaSIPrior} and
Sec.~\ref{sec:NumericalOptimizationWithAQuaSIPrior}, respectively.
Quantitative and qualitative experiments on three applications of AQuaSI are
presented in Sec.~\ref{sec:Applications}.  Finally, Sec.~\ref{sec:Conclusion}
concludes this work.

\section{Problem Statement and Related Work}
\label{sec:RelatedWork}

The focus of this work is on inverse problems in computer vision. An inverse problem aims at recovering a latent image $\vec{f} \in \mathbb{R}^n$ from measurements $\vec{g} \in \mathbb{R}^m$. In a maximum a-posteriori (MAP) formulation, the latent image maximizes the posterior distribution $p(\vec{f}|\vec{g})$ and can be inferred as the solution of an optimization problem using Bayes rule
\begin{equation}
	\label{eqn:invProbMap}
	\hat{\vec{f}} 
		= \argmax_{\vec{f}} p(\vec{f} | \vec{g})
		= \argmax_{\vec{f}} p(\vec{g}|\vec{f})p(\vec{f})\enspace.
\end{equation}
The prior $p(\vec{f})$ models the distribution of the latent image and the observation model $p(\vec{g}|\vec{f})$ denotes the probability of observing the measurement $\vec{g}$ from the latent image $\vec{f}$. Using the negative log-likelihood $-\log(p(\vec{f}|\vec{g}))$, \eqref{eqn:invProbMap} can be written as the energy minimization task
\begin{equation}
	\label{eqn:invProbDef}
	\hat{\vec{f}} = \argmin_{\vec{f}} \left\{ \mathcal{L}(\vec{f}, \vec{g}) + \lambda R(\vec{f}) \right\}\enspace,
\end{equation}
where $\mathcal{L}(\vec{f}, \vec{g}) = - \log(p(\vec{g}|\vec{f}))$ is a \textit{data term} to measure the
fidelity of an estimate $\vec{f}$ \wrt $\vec{g}$ under a given image formation
model, and $\lambda R(\vec{f}) = -\log(p(\vec{f}))$ is a \textit{regularization term} with weight $\lambda
\geq 0$ derived from prior knowledge on the solution $\hat{\vec{f}}$. In common frameworks for denoising, deblurring, or super-resolution, the data term is
$\mathcal{L}(\vec{f}, \vec{g}) = \|\vec{g} - \vec{W}\vec{f}\|_2^2$, where $\vec{W} \in
\mathbb{R}^{n \times m}$ is a task-dependent operator that models an
underlying linear image formation model. While the choice of the data term is
highly dependent on the application, it is oftentimes possible to define more
generic priors for the regularization term $R(\vec{f})$. The large body of
existing works on image priors can roughly be organized into priors on
sparsity, priors from deep learning, and priors from denoising.

\subsubsection{Natural Scene Statistics and Sparsity Priors} 
The vast majority of classical image priors p(\vec{f}) for \eqref{eqn:invProbMap} are derived from statistical models of natural scenes. Some common assumptions are to consider natural images as smooth, piecewise smooth, or piecewise constant signals under certain transforms. This includes the well known TV prior~\cite{Perrone2016} that has been widely adopted for image restoration. Many elaborated optimizers exist to solve inverse problems with a TV prior, but the underlying statistical assumption of a piecewise constancy of image intensities is in many applications too simplistic to appropriately model the complex nature of natural scenes. 

Further advances in the field of natural scene statistics led to
Hyper-Laplacian priors~\cite{Krishnan2009a,Kohler2015c,Pan2016} or normalized
scale-invariant priors~\cite{Krishnan2011}. Specifically for deblurring,
non-linear extreme value priors like dark~\cite{Pan2016a} or bright channels
\cite{Yan2017} have been proposed. Such approaches exploit sparsity of natural
images as a stronger prior for image restoration. Spectral models
\cite{Wang2013,Gu2014} are likewise heading towards this direction but exploit
low-rank properties of non-local similar patches. These are more suitable to
model natural scene statistics but require specific optimization strategies
like re-weighted minimization~\cite{Kohler2015c} or variable splitting
\cite{Pan2016}, which limits their flexibility. In contrast, we aim in this
work at designing an \textit{out-of-the box} image prior that is widely
applicable within different optimization schemes.

\subsubsection{Deep Learning-Based Priors}
The use of deep learning architectures like convolutional neural networks
(CNNs) is a different track to design the prior p(\vec{f}) in \eqref{eqn:invProbMap} and
has become popular to solve low-level vision problems. In early works,
classical approaches to sparse coding like~\cite{Yang2010} have been newly
interpreted via CNNs and learned in an end-to-end fashion from exemplars
\cite{Wang2015}. Other approaches are to integrate denoising networks to
regularize model-based optimization in variable splitting schemes
\cite{Zhang2017,Meinhardt2017}. However, these methods require learning from
large-scale example datasets, which are not readily available in many
real-world applications. In contrast, the proposed prior in this work is
\textit{unsupervised}.

Deep image priors~\cite{Ulyanov2018, Mataev2019} are one alternative, as prior
knowledge on natural images is extracted from network architectures instead of
example data.  In principle, this avoids the use of simplistic hand-crafted
features for regularization, but such priors are --- other than the proposed
prior --- difficult to deploy in existing model-based image restoration
frameworks.

\subsubsection{Transforming a Denoiser into a Regularizer}
Most closely related to our proposed approach are strategies to form the prior $p(\vec{f})$ in \eqref{eqn:invProbMap} via image denoising. PnP priors are one of these attempts and replace proximal operators in variable splitting algorithms like the alternating direction method of multiplier (ADMM) \cite{Venkatakrishnan2013,Chan2017} or primal dual optimization \cite{Ono2017} by invoking denoising algorithms. Moreover, variants of the iterative shrinkage/thresholding algorithm (ISTA) \cite{Daubechies2004} have been proposed to extend the PnP formulation towards non-linear \cite{Kamilov2017} and online optimization problems \cite{Sun2019o}. This implicitly models a prior on the solution space and have gained much popularity in signal recovery and image restoration. Chan \cite{Chan2019} provided theoretical insights to the actual performance of PnP by deriving relations to classical MAP inference and consensus equilibrium \cite{Buzzard2018} for the class of symmetric smoothing priors.

 In that sense, an alternative approach includes regularization by denoising (RED) as proposed by Romano \etal~\cite{Romano2016a}. Their model considers clean images as a fixed point of image denoising. In its core, RED extends the well known fixed point property of natural images as raised in related works on autoencoding \cite{Bigdeli2017} or mean-shift priors \cite{Bigdeli2017b} to general denoisers. This allows the integration of many advanced filters to solve inverse imaging problems under different optimization schemes and broadens the scope compared to PnP. In principle, RED can also be combined with other complementary models like deep image priors \cite{Mataev2019}. The used denoiser can be related to an analytical prior in a MAP framework provided that it has symmetric Jacobian \cite{Reehorst2018} -- a property that is, however, not fulfilled by many elaborated filters including those covered in this work. %
From a practical perspective, RED is based on Tikhonov regularization similar to related autoencoding priors as proposed in the work of Bigdeli \etal \cite{Bigdeli2017}. In \cite{Ahmad2019}, the authors showed that PnP can also be related to Tikhonov regularization. A direct comparison in~\cite{Chan2019} shows that RED is less sensitive to parameters compared to PnP, while PnP yields better performance in terms its mean squared error. However, overall Tikhonov regularization can be expected to be less robust to non-Gaussian noise, \eg, than $L_1$-based regularizer. The proposed \aquasi prior aims at increasing the robustness using a sparsity-promoting model without relying on strong assumptions like a symmetric Jacobian for its interpretation.

\section{Adaptive Quantile Sparse Image (AQuaSI) Prior}
\label{sec:FormulatingTheAQuaSIPrior}

The proposed \aquasi prior is a spatially adaptive model for natural images. It
is sensitive to common image degradations, and can be efficiently implemented
in various optimization frameworks. In this section, we first introduce the
prior itself and its behavior under common image degradations. Then, we show
its applicability in joint filtering, and finally present a linearization of the filter for
insertion into an optimizer. 

\begin{figure}[!tbp]
	\centering
	\footnotesize
	\setlength \figurewidth{0.2\textwidth}
	\setlength \figureheight{0.80\figurewidth}
	\begin{tabular}{llll}
	\subfloat{\includegraphics[width = 0.21\textwidth]{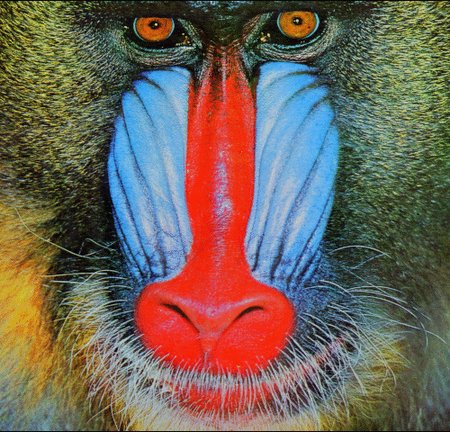}}&
	\subfloat{\includegraphics[width = 0.21\textwidth]{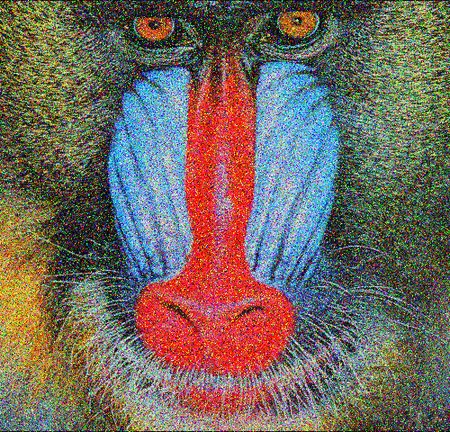}}&		
	\subfloat{\includegraphics[width = 0.21\textwidth]{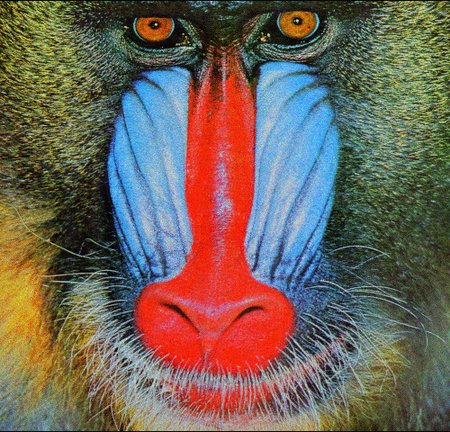}}&					
	\subfloat{\includegraphics[width = 0.21\textwidth]{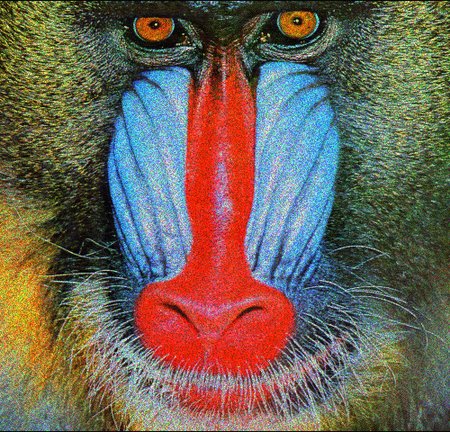}}\\
	\setcounter{subfigure}{0}
	\subfloat[Clean]{\includegraphics[width = 0.21\textwidth]{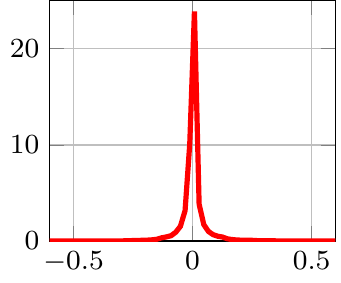}}&
	\subfloat[Gaussian / salt \& pepper]{\includegraphics[width = 0.21\textwidth]{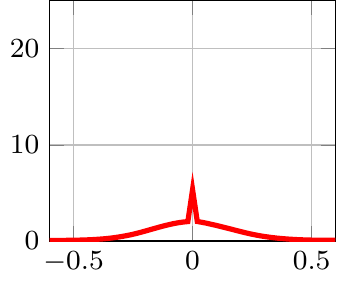}}&
	\subfloat[Poisson]{\includegraphics[width = 0.21\textwidth]{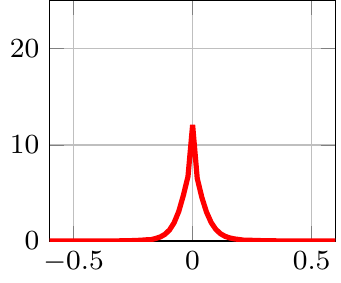}}&
	\subfloat[Speckle]{\includegraphics[width = 0.21\textwidth]{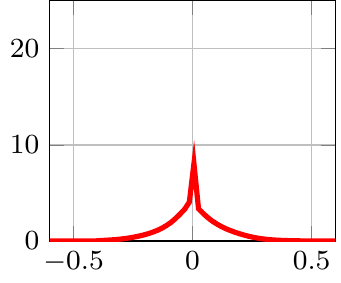}}\\
	\end{tabular}
	\caption{Influence of different images degradations to the behavior of the \aquasi prior. Top row: clean and degraded \textit{baboon} image using Gaussian, mixed Gaussian / salt \& pepper, Poisson, and speckle noise, respectively. Bottom row: empirical distribution of the quantile residual $\vec{r} = \vec{f} - Q_{p, \vec{w}}(\vec{f})$ for the different degradations. Notice that the distributions on clean images are strongly concentrated around 0, while the distribution of noisy images are very broad.} 
	\label{fig:noiseStudyImages}
\end{figure}

\subsection{Definition of the \aquasi Prior}

Turning denoising algorithms into an image prior for regularization  is an
emerging topic in image restoration. Our approach follows a similar train of
thought and is based on \textit{weighted quantile filtering} as a particular
denoiser.
We denote a weighted $p$-quantile filter with $0 \leq p \leq 1$ for an image
$\vec{f}$ as $Q_{p,\vec{w}}(\vec{f})$, where $\vec{w} \in \mathbb{R}^N$ are
spatially adaptive weights. The weighted quantile filter selects a pixel from a neighborhood based on two considerations: a) the intensity distribution in the neighborhood, b) the distribution of similarities of each pixel to the center pixel. That way, pixels that are very dissimilar to the center can be weighed down, and
effectively be ignored by the quantile filter.

Let $\vec{w}_i \in \mathbb{R}^n$ with $n =
|\mathcal{N}(i)|$ be the weights within the kernel $\mathcal{N}(i)$ of this
filter centered at the $i$-th pixel.
Then, we define a permutation of these weights, $\tilde{\vec{w}}_i =
m_i(\vec{w})$, such that the associated pixel intensities $m_i(f_{ij}) \in
\mathcal{N}(i)$ are sorted in ascending order.
%
The weighted
$p$-quantile is the element $[Q_{p,\vec{w}}(\vec{f})]_i = f_l$ with
\begin{equation}
	\label{eqn:weightedQuantileDef}
	w_{l} = m_i^{-1}(\tilde{w}_{i{k^*}})\enspace
\end{equation}
where $m_i^{-1}(\tilde{\vec{w}_i})$ maps the elements of the permutation $\tilde{\vec{w}_i}$ back to weights in the filter kernel $\vec{w}_i$ and the index $k^*$ is given by:
\begin{equation}
	\label{eqn:weightedQuantileDefPartTwo}
	k^* = \mathop{\mathrm{min}} k \quad \mathrm{s.t.} \quad \sum_{j = 1}^k \tilde{w}_{ij} \geq p \sum_{j = 1}^{n} \tilde{w}_{ij}\enspace.
\end{equation}

The general model in \eqref{eqn:weightedQuantileDef} and
\eqref{eqn:weightedQuantileDefPartTwo} depends on the percentage parameter $p$,
which is a hyperparameter to adjust the filter for specific applications. If
$p = 0.5$, we obtain the weighted median filter, which has been shown to be
particularly suitable for denoising under zero-mean
noise~\cite{Ma2013,Zhang2014}. 

Conceptionally, the proposed \aquasi prior follows the idea that a natural
image should be a fixed point under the weighted quantile filter. That is,
the \textit{quantile residual} of a clean image $\vec{f}$ is zero, i.\,e.
$\vec{f} - Q_{p,\vec{w}}(\vec{f}) = \vec{0}$. In order to apply this assumption for the
regularization of inverse problems, we define the prior with slight relaxation
according to the sparsity-inducing term
\begin{equation}
	\label{eqn:aquasiDef}
	R_{\mathrm{AQuaSI}}(\vec{f}) = \|\vec{f} - Q_{p,\vec{w}}(\vec{f})\|_1\enspace.
\end{equation}

Under the general MAP framework in \eqref{eqn:invProbMap}, we can relate this regularization term to 
the prior distribution $p(\vec{f}) = \exp(- \lambda R_{\mathrm{AQuaSI}}(\vec{f}))$.

\subsection{Impact of Image Degradations on the \aquasi Prior}
\label{sec:InfluenceOfImageDegradationsToTheAQuaSIPrior}

We validate the key assumptions behind our \aquasi prior on natural images that have been
subject to different types of degradations. Specifically, we present the
behavior of \aquasi for several noise distributions. We use the images from Set14~\cite{Zeyde2012} as \emph{clean} images, and simulate \emph{noisy}
counterparts using mixed Gaussian and salt \& pepper noise, Poisson noise, and speckle noise. Sample images are shown in the top row of
Fig.~\ref{fig:noiseStudyImages}. For each noise model, we compute the quantile residual
\begin{equation}
	\vec{r} = \vec{f} - Q_{p, \vec{w}}(\vec{f}) \enspace,
	\label{eqn:residual}
\end{equation}
with $p = 0.5$ and $\vec{w}=\mathbf{1}$ on a neighborhood of $n = 5^2$ pixels, which
corresponds to the usual median operator on a window of $5 \times 5$ pixels.
In the bottom row of Fig.~\ref{fig:noiseStudyImages}, histograms of the residuals are shown.

Two observations can be made from these distributions. First, the residual
distributions of noisy images are very broad. In contrast to that, residual
distributions on clean images are strongly concentrated around zero.
Second, also for clean images, there are a few notable outliers that
largely differ from zero.
Both observations together indicate that the proposed quantile residual is an
adequate model for natural images: the overall concentration around zero
indicates that the quantile is a good predictor for most pixels, and the
use of the sparsity-promoting $L_1$ norm ensures that the influence of the few
outliers in the clean image does not dominate the overall result.

\subsection{Joint Filtering}

\begin{figure}[!tbp]
	\centering
	
	\subfloat[Noisy color image]{
		\begin{tikzpicture}[spy using outlines={rectangle,orange,magnification=2.5, 
			height=2.5cm, width = 1cm, connect spies, every spy on node/.append style={thick}}]
		\node {\pgfimage[width = 0.25\textwidth]{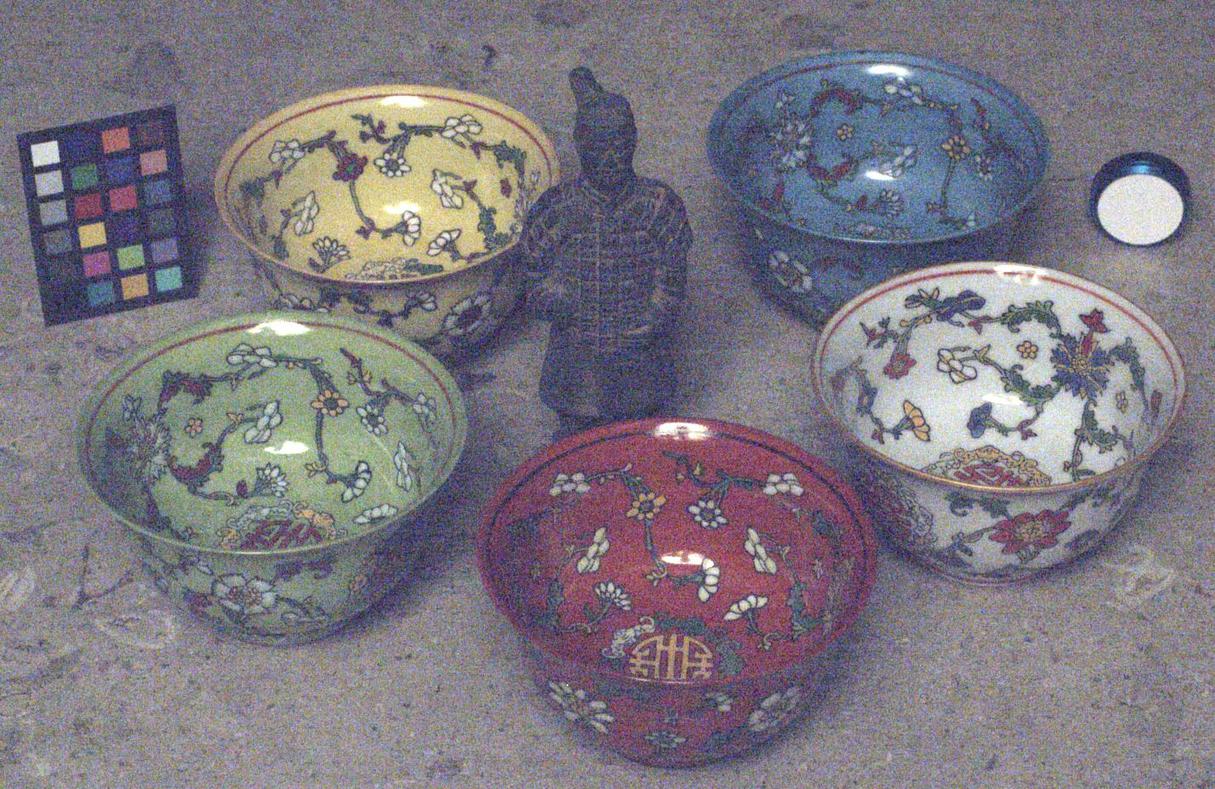}};
		\spy on (-1.5,-0.2) in node [left] at (-1.95, 0.0);		  
		\end{tikzpicture}\label{fig:bowlsRGB}}
	\subfloat[ Near-infrared image]{
		\begin{tikzpicture}[spy using outlines={rectangle,orange,magnification=2.5, 
			height=2.5cm, width = 1cm, connect spies, every spy on node/.append style={thick}}]
		\node {\pgfimage[width = 0.25\textwidth]{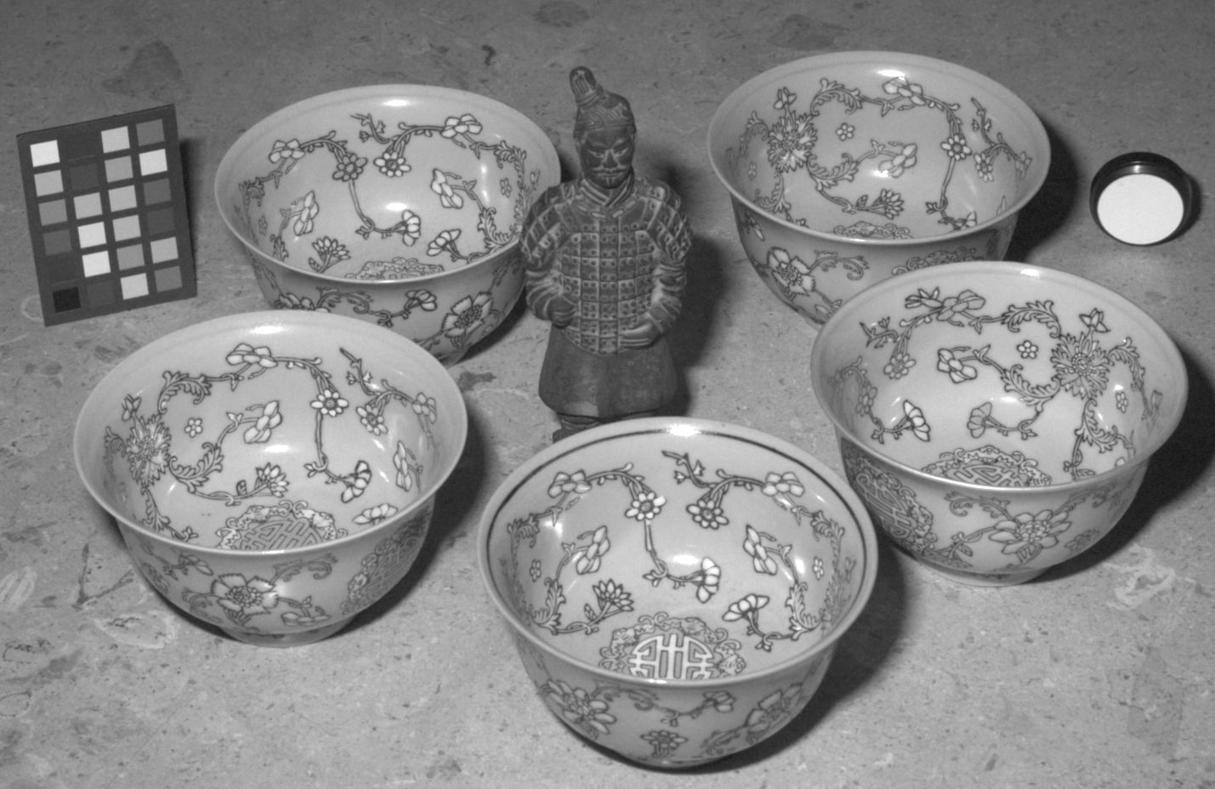}};
		\spy on (-1.5,-0.2) in node [left] at (-1.95, 0.0);			  
		\end{tikzpicture}\label{fig:bowlsNIR}}
	\subfloat[QuaSI \cite{Schirrmacher2018}]{
		\begin{tikzpicture}[spy using outlines={rectangle,orange,magnification=2.5, 
			height=2.5cm, width = 1cm, connect spies, every spy on node/.append style={thick}}]
		\node {\pgfimage[width = 0.25\textwidth]{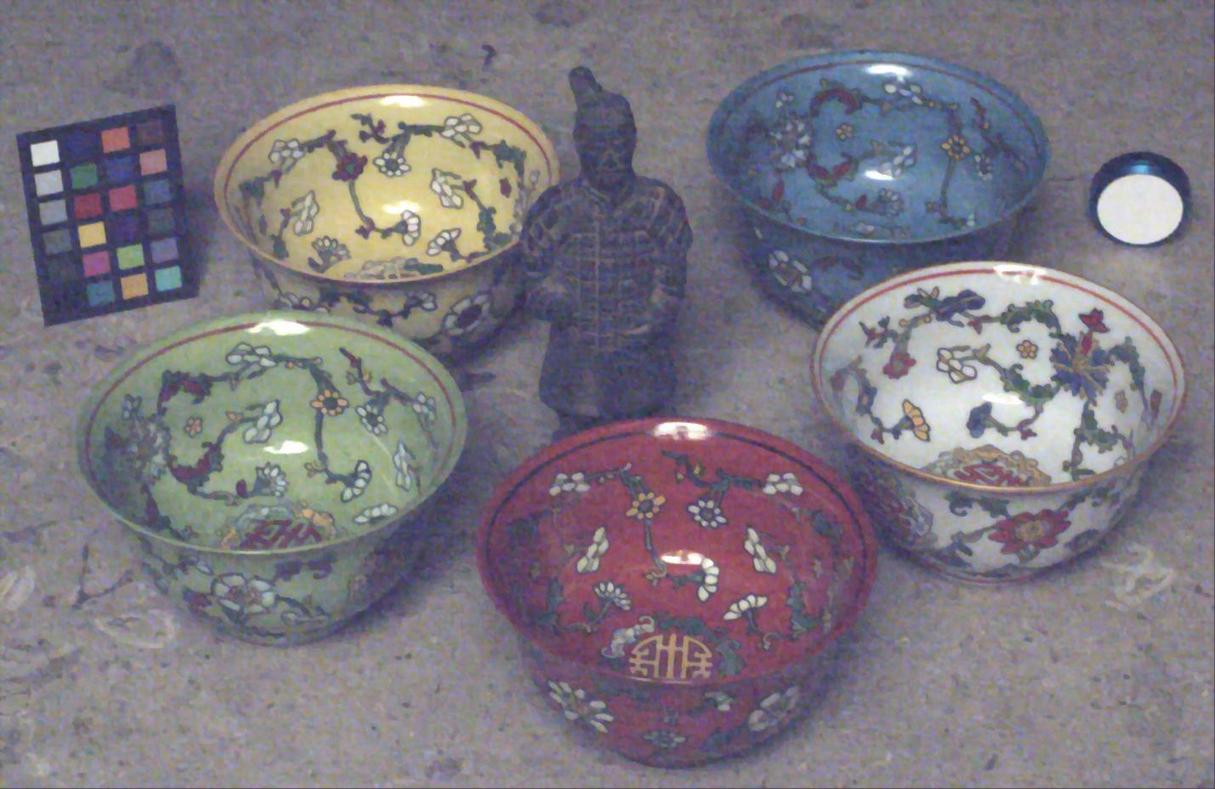}};
		\spy on (-1.5,-0.2) in node [left] at (-1.95, 0.0);				  
		\end{tikzpicture}\label{fig:bowlsQuaSI}}\\[-1.0ex]
	\subfloat[AQuaSI - \textit{Dynamic} guidance, fixed $\vec{f}$]{
		\begin{tikzpicture}[spy using outlines={rectangle,orange,magnification=2.5, 
			height=2.5cm, width = 1cm, connect spies, every spy on node/.append style={thick}}]
		\node {\pgfimage[width = 0.25\textwidth]{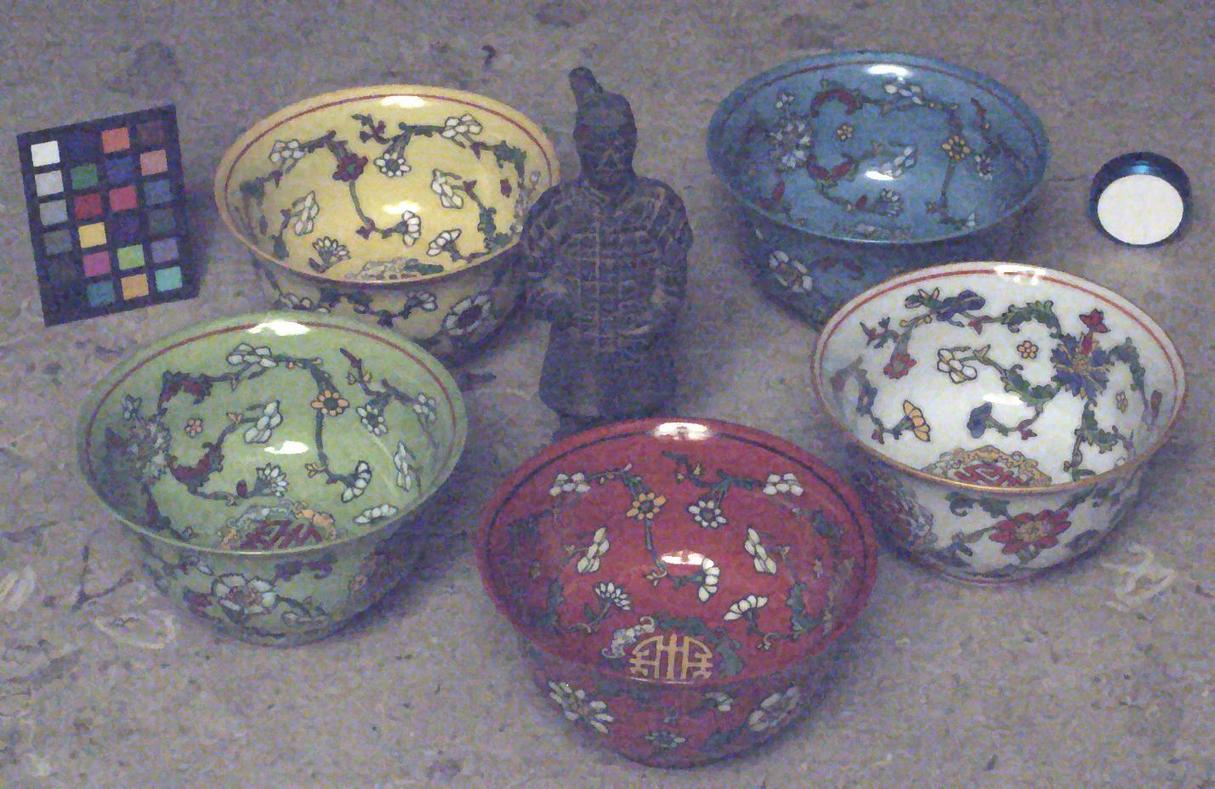}};
		\spy on (-1.5,-0.2) in node [left] at (-1.95, 0.0);				  
		\end{tikzpicture}\label{fig:bowlsAQuaSINI}}
	\subfloat[AQuaSI - \textit{Dynamic} guidance, updated $\vec{f}$]{
		\begin{tikzpicture}[spy using outlines={rectangle,orange,magnification=2.5, 
			height=2.5cm, width = 1cm, connect spies, every spy on node/.append style={thick}}]
		\node {\pgfimage[width = 0.25\textwidth]{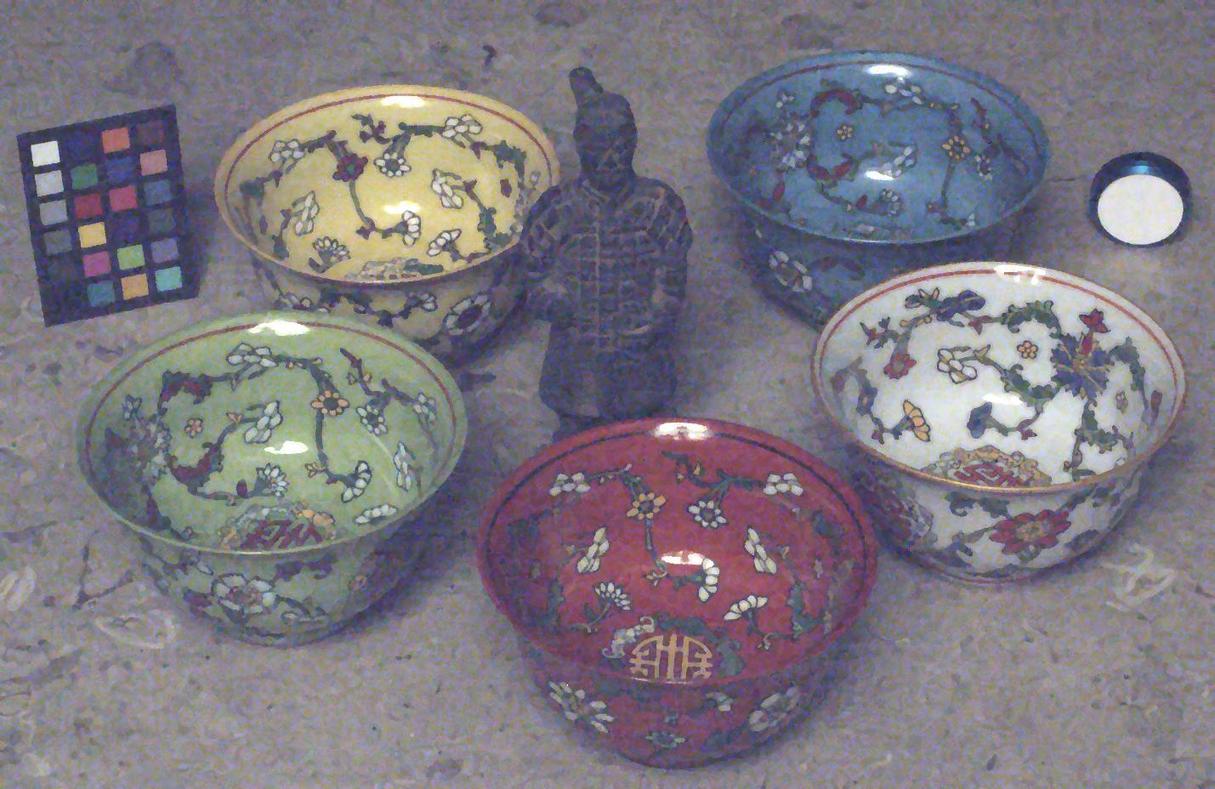}};
		\spy on (-1.5,-0.2) in node [left] at (-1.95, 0.0);				  
		\end{tikzpicture}\label{fig:bowlsAQuaSIII}}
	\subfloat[AQuaSI - \textit{Static} guidance]{
		\begin{tikzpicture}[spy using outlines={rectangle,orange,magnification=2.5, 
			height=2.5cm, width = 1cm, connect spies, every spy on node/.append style={thick}}]
		\node {\pgfimage[width = 0.25\textwidth]{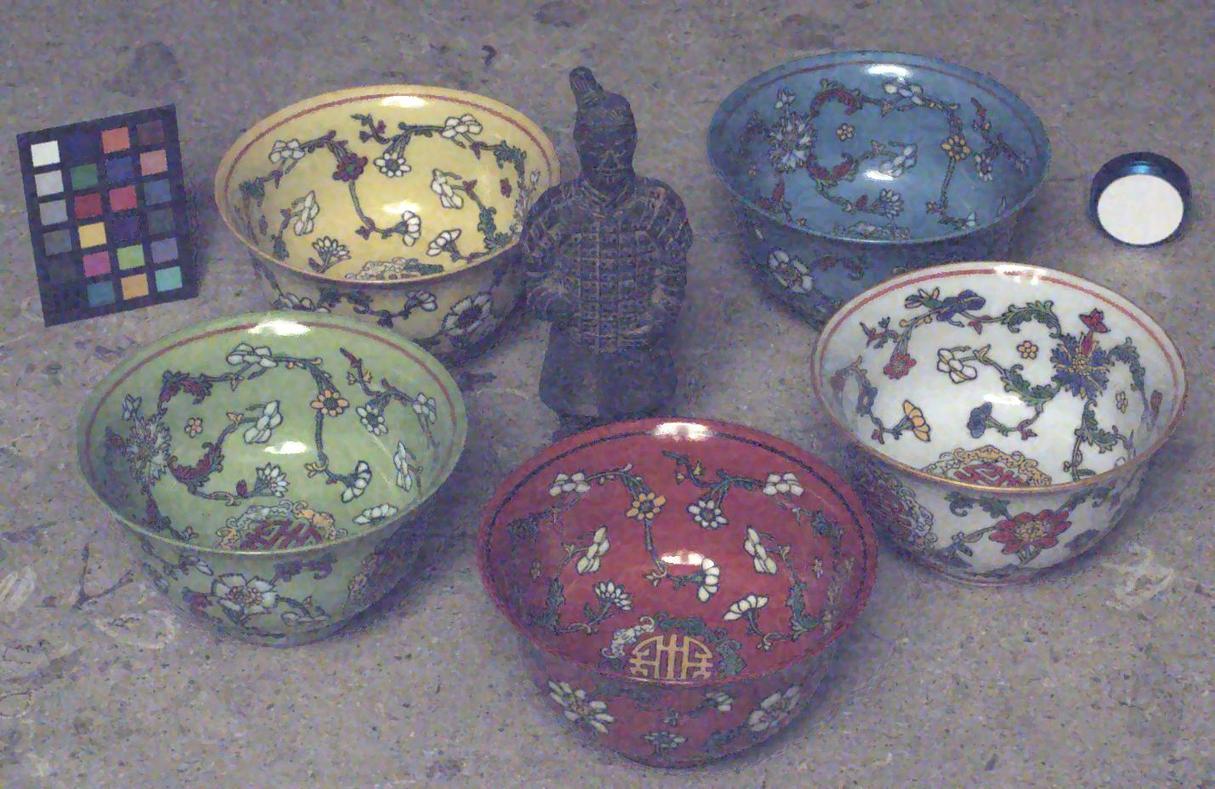}};
		\spy on (-1.5,-0.2) in node [left] at (-1.95, 0.0);			  
		\end{tikzpicture}\label{fig:bowlsAQuaSIGI}}	
	\caption{Denoising of the \emph{bowls} color image~\cite{Krishnan2009}
		using the ADMM optimization in combination with different variants of our prior. QuaSI \cite{Schirrmacher2018} is spatially non-adaptive and can only exploit color data for regularization. \aquasi is spatially adaptive and can be steered by color data (either the fixed input or an intermediate ADMM result) in a \textit{dynamic} guidance mode or by external near-infrared data in a \textit{static} guidance mode.  Notice the superior recovery of small image structures achieved by the spatially adaptive variants.}
	\label{fig:bowlsComparison}
\end{figure}

Similar to existing local~\cite{Zhang2014,He2013} or global~\cite{Ham2015} optimization-based
methods, the proposed prior facilitates \textit{joint filtering}
by steering the regularizer in \eqref{eqn:aquasiDef} with either internal or
external image features. This is achieved by controlling the weights $\vec{w}$
of the underlying weighted quantile operator. At the same
time, the non-uniform choice of weights make the proposed technique spatially
adaptive.

More specifically, for regularization of $\vec{f}$, the weight $w_{ij}$ for
the pixel $f_j$ \wrt the pixel $f_i$ is determined according to
\begin{equation}
	\label{eqn:weightDef}
	w_{ij} = G_{\sigma_w}\left( z_i, z_j \right)\enspace,
\end{equation}
where $z_i$ and $z_j$ are the respective pixels in a \textit{guidance image}
$\vec{z}$, and $G_{\sigma_w}\left( z_i, z_j \right)$ is a measure for their
spatial neighborhood. For instance, we can use an isotropic Gaussian kernel
\begin{equation}
	G_{\sigma_w}\left( z_i, z_j \right) = \exp{ \left(-\frac{(z_i - z_j)^2}{2\sigma_w^2} \right)}\enspace, 
\end{equation}
where $\sigma_w$ denotes the standard deviation. 
Joint filtering can be done in two modes: In \textit{static} guidance mode
\cite{Zhang2014,He2013}, spatially adaptive weights $\vec{w}$ are derived from an
external guidance image $\vec{z}$. In \textit{dynamic} guidance mode
\cite{Ham2015}, spatially adaptive weights $\vec{w}$ are internally derived in the domain
of the regularized image without considering additional guidance data. 
This can be done using either the fixed input $\vec{g}$ to obtain spatially adaptive weights (\ie, $\vec{z} = \vec{g}$) or using an indeterminate result $\vec{f}$ (\ie $\vec{z} = \vec{f}$) to gradually update the weights within iterative optimization as introduced in Sec.~\ref{sec:NumericalOptimizationWithAQuaSIPrior}.

Spatial adaptivity has a large impact on image structure preservation,
and as such greatly improves upon our earlier, spatially non-adaptive prior,
which we then called ``QuaSI''~\cite{Schirrmacher2018}. In order to show the
impact of spatial adaptivity, we compare \aquasi with \textit{static} and \textit{dynamic} guidance to the
non-adaptive QuaSI prior. The is done 
for the task of RGB and near-infrared (NIR) image restoration (see Sec.~\ref{sec:RGBAndNIRImageRestoration})~\cite{Yan2013},
where the filtering of noisy color images is guided by NIR data.
The parameters of the quantile filter
are $p = 0.5$ for a median filter on a $n = 5^2$ neighborhood and $\sigma_w = 0.1$ for the Gaussian kernel $G_{\sigma_w}$.

Figure~\ref{fig:bowlsComparison} shows the behavior of the different priors on the \emph{bowls} scene.
Fine details, such as the border of the flower
printing, are better preserved in the filtered color image using \aquasi with \textit{static} guidance obtained from NIR data: the edges in
Fig.~\ref{fig:bowlsAQuaSIGI} are sharper than the edges in
Fig.~\ref{fig:bowlsQuaSI}. Also between the spatially non-adaptive QuaSI and
\aquasi with \textit{dynamic} guidance, \aquasi performs better in preserving
more details of the flower printing while smoothing homogeneous regions.
We also observe superior denoising performance for \textit{dynamic} mode when using
intermediate results as guidance (Fig.~\ref{fig:bowlsAQuaSIII}) than when using
the noisy input image as guidance (Fig.~\ref{fig:bowlsAQuaSINI}). Also the flower printings are better preserved. 

Thus, the spatial adaptivity and the possibility of using additional guidance data broadens the
applicability and enhances the performance of the proposed \aquasi prior compared to the
QuaSI prior. 

\subsection{Pseudo-Linear Filtering}
\label{sec:PseudoLinearFiltering}

The proposed \aquasi prior is both non-convex and highly non-linear due to the
underlying weighted quantile filtering. However, this flexibility somewhat
complicates the numerical optimization when using this model for an image
restoration task. To mitigate this issue, a proper linearization can be used to
deploy our prior for the solution of inverse imaging problems. 
We propose to formulate $Q_{p,\vec{w}}(\vec{f})$ in the pseudo-linear form
\cite{Romano2016a}:

\begin{equation}
	\label{eqn:pseudoLinearForm}
	Q_{p,\vec{w}}(\vec{f}) = \vec{Q} \vec{f}\enspace,
\end{equation}
where $\vec{Q}$ denotes the weighted quantile filter in matrix
notation constructed from the image $\vec{f}$. This decomposes the filter into the construction of
its pseudo-linear form $\vec{Q}$ adapted to the image $\vec{f}$ and the actual filtering. While this formulation
applies in principle to many popular non-linear models, like non-local means or
bilateral filtering, the proposed weighted quantile filtering allows for the
efficient linearization
\begin{equation}
	Q_{ij} =
	\begin{cases}
1 & \mathrm{if} \  j=z \text{\ in the sense of \eqref{eqn:weightedQuantileDef}} \\ 
0 & \mathrm{otherwise}
\end{cases}\enspace,
\end{equation}
where $z$ denotes the position of the $p$-weighted quantile in the neighborhood $\mathcal{N}(i)$ centered at the $i$-th pixel.

The pseudo-linear form in \eqref{eqn:pseudoLinearForm} is accurate for a fixed image $\vec{f}$. In case of small perturbations in the image that do not affect the ordering of the pixel values in the neighborhood $\mathcal{N}(i)$, this form is still accurate and $\vec{Q}$ remains constant. 

\section{Numerical Optimization with AQuaSI Prior}
\label{sec:NumericalOptimizationWithAQuaSIPrior}

\aquasi can be deployed within several widely used iterative optimization
schemes for inverse imaging problems. These schemes follow the quite general
energy minimization framework of \eqref{eqn:invProbDef}.

\subsection{Gradient Descent}
\label{sec:GradientDescent}

Gradient descent is among the most frequently used optimization methods to
solve inverse imaging problems.
We study steepest descent as a conceptional simple, yet flexible gradient descent iteration scheme.
Given an estimate $\vec{f}^{t}$ for the solution of the inverse problem in
\eqref{eqn:invProbDef} with the problem-specific data fidelity term
$\mathcal{L}(\vec{f}, \vec{g})$ and our proposed regularization term $R_{\mathrm{AQuaSI}}(\vec{f})$, we refine $\vec{f}^{t}$ from iteration $t$ to
$t+1$ according to the update equation

\begin{equation}
	\label{eqn:gradDescUpdate}
	\begin{split}
	\vec{f}^{t+1} 
	= \enspace & \vec{f}^{t} - \mu \Big( \nabla_{\vec{f}} \mathcal{L}\big(\vec{f}^{t}, \vec{g}\big)
	+ \lambda \nabla_{\vec{f}}R_{\mathrm{AQuaSI}}\big(\vec{f}^{t}\big) \Big)\enspace,
	\end{split}
\end{equation}
where $\nabla_{\vec{f}}$ denotes the gradient calculated at $\vec{f} = \vec{f}^{t}$ and $\mu$ is the iteration step size parameter. To calculate the gradient of the \aquasi regularization term for this update rule, $Q_{p,\vec{w}}(\vec{f})$ is constructed with the
pseudo-linear form in \eqref{eqn:pseudoLinearForm} around the current estimate
$\vec{f}^t$ such that $Q_{p,\vec{w}}(\vec{f}^t) = \vec{Q} \vec{f}^t$. Based on this linearization, 
the gradient of the \aquasi term is
\begin{equation}
	\begin{split}
		\nabla_{\vec{f}} R_{\mathrm{AQuaSI}}(\vec{f}^{t})
		&= \nabla_{\vec{f}} \|\vec{f} - Q_{p,\vec{w}}(\vec{f})\|_1\big|_{\vec{f} = \vec{f}^{t}} \\
		&= (\vec{I} - \vec{Q})^\top \sign{ \vec{f} - \vec{Q} \vec{f}^{t} }\enspace.
	\end{split}
\end{equation}

To handle the non-differentiability of the $L_1$ norm at zero, we use an $\epsilon$-smoothed version with $\sign{u} \approx u ( \sqrt{u^2 + \epsilon} )^{-1}$ for small $\epsilon$. 

The above iteration scheme can be employed with fixed step size parameter $\mu$ or line searches to determine $\mu$ dynamically per iteration. Moreover, we can revise the selection of the search direction from the steepest descent approach to related techniques like conjugate gradient (CG) iterations.

\subsection{Variable Splitting}
\label{sec:AlternatingDirectionMethodOfMultipliersADMM}

\aquasi can also be used in variable splitting techniques for minimization of \eqref{eqn:invProbDef}.
As one of the most popular techniques within this class of optimizers, we study
the alternating direction method of multipliers (ADMM)
\cite{Goldstein2009}, where \eqref{eqn:invProbDef} is rewritten as the
constrained optimization problem
\begin{equation}
\begin{split}
	(\hat{\vec{f}}, \hat{\vec{u}}) 
	&= \argmin_{\vec{f}, \vec{u}} \left\{ \mathcal{L}(\vec{f}, \vec{g}) + \lambda \|\vec{u}\|_1 \right\}\\	
	&\textrm{s.t.}~ \vec{u} =  \vec{f} - Q_{p,\vec{w}}(\vec{f})
	\end{split}
\end{equation}
with auxiliary variable $\vec{u}$. This is equivalent to enforcing the
non-linear constraint $\vec{u} = \vec{f} - Q_{p,\vec{w}}(\vec{f})$ by
minimizing the Augmented Lagrangian
\begin{equation}
	\label{eqn:al}
	\begin{split}
	(\hat{\vec{f}}, \hat{\vec{u}}) 
	&= \argmin_{\vec{f}, \vec{u}}  \Big\{ \mathcal{L}(\vec{f}, \vec{g}) +  \lambda \|\vec{u}\|_1 \\ &
	+ \frac{\alpha}{2}\|\vec{u} - \vec{f} + Q_{p,\vec{w}}(\vec{f}) - \vec{b}_u \|_2^2 \Big\} \enspace,
	\end{split}
\end{equation}
where $\alpha$ and $\vec{b}_u$ denote the corresponding Lagrangian multiplier and a Bregman variable, respectively.

The minimization of \eqref{eqn:al} is accomplished in an alternating
manner
\begin{align}
	\label{eqn:admmMinF}
	\begin{split}
	\vec{f}^{t+1}&= \argmin_{\vec{f}} \Big\{ \mathcal{L}(\vec{f}, \vec{g}) \\
				 &\qquad \quad + \frac{\alpha}{2}\|\vec{u}^t - \vec{f} + Q_{p,\vec{w}}(\vec{f}) - 
						\vec{b}_u^t\|_2^2\Big\}\enspace, 
	\end{split}\\
	\begin{split}
	\vec{u}^{t+1} &= \argmin_{\vec{u}} \Big\{ \lambda \|\vec{u}\|_1 \\
				& \qquad \quad + \frac{\alpha}{2}\|\vec{u} - \vec{f}^{t+1} + Q_{p,\vec{w}}(\vec{f}^{t+1}) -
					 \vec{b}_u^t\|_2^2\Big\}\enspace,
	\end{split}\label{eqn:admmShrink}\\ 
	&= \mathrm{shrink} \left( \vec{f}^{t+1} + Q_{p,\vec{w}}(\vec{f}^{t+1}) + \vec{b}_u^t, \frac{\lambda}{\alpha} \right)\enspace, \\
	\vec{b}_u^{t+1} &= \vec{b}_u^t + \left( \vec{f}^{t+1} - Q_{p,\vec{w}}(\vec{f}^{t+1}) - \vec{u}^{t+1} \right)\enspace, 
	\label{eqn:admmBregman}
\end{align}
where $\mathrm{shrink}(u, \gamma) = \sign{u} \max(u - \gamma, 0)$ is the
shrinkage operator associated with the $L_1$ norm $\| \vec{u} \|_1$. To reduce the influence of the selection of the Lagrangian multiplier,
we adopt varying penalty parameters~\cite{Boyd2011}. The weighted quantile filter $Q_{p,\vec{w}}(\vec{f})$ is again approximated by the pseudo-linear form in \eqref{eqn:pseudoLinearForm}.

\subsection{Multi-Channel Inverse Problems}
\label{sec:MultiChannel}

The \aquasi prior can be easily generalized to handle multi-channel data like RGB color or multispectral images. Applying this model to each channel separately is straightforward but would ignore inter-channel correlations like structures present in all channels. Multi-channel images can be jointly optimized in order to allow for simultaneous processing of all channels. To this end, we couple all channels in the \aquasi prior. 

Let $\vec{f}_c \in \mathbb{R}^{N}$, $c = 1, \ldots, C$ be the individual channels of a multi-channel image denoted in vector notation. Based on \eqref{eqn:invProbDef}, we plug the \aquasi prior into a multi-channel inverse problem according to:
\begin{equation}
	\argmin_{\vec{f}_1, \ldots, \vec{f}_C} \left\{\sum_{c=1}^C \mathcal{L}(\vec{f}_c, \vec{g}_c)  \\ 
	+ \lambda  \left\Vert \bar{\vec{f}} - Q_{p,\vec{w}}( \bar{\vec{f}} )\right\Vert_1 \right\} \enspace,
\label{eqn:objective3D}
\end{equation}
where $\bar{\vec{f}}$ denotes the weighted average:
\begin{equation}
	\bar{\vec{f}} = \sum_{c=1}^C m_c \vec{f}_c\enspace.
	\label{eqn:meanImage}
\end{equation}

For instance, in case of RGB color images, using the weights $\vec{m} = (0.299, 0.587, 0.114)^\top$ is equivalent to a standard grayscale conversion. We replace the non-linear filter $Q_{p,\vec{w}}(\cdot)$ with its pseudo-linear form $\bar{\vec{Q}}$ computed from $\bar{\vec{f}}$ to exploit inter-channel correlations and to allow for better structure preservation. Then, we can resort the \aquasi prior on the average image:

\begin{equation}
	\begin{split}
		\Vert \bar{\vec{f}} - Q_{p,\vec{w}}( \bar{\vec{f}} )\Vert_1 
		&= \left\Vert \sum_{c=1}^C m_c \vec{f}_c - Q_{p,\vec{w}} \left( \sum_{c=1}^C m_c \vec{f}_c \right) \right\Vert_1 \\
		&\approx \left\Vert \sum_{c=1}^C m_c \vec{f}_c - \sum_{c = 1}^C m_c \bar{\vec{Q}} \vec{f}_c \right\Vert_1 \\
		&= \left\Vert \sum_{c=1}^C m_c \left(\vec{f}_c -  \bar{\vec{Q}}  \vec{f}_c \right) \right\Vert_1 \enspace.
	\end{split}
\end{equation}
Thus, we apply a single linearization $\bar{\vec{Q}}$ to the individual channels of the underlying inverse problem. All channels are stacked in the variables $\vec{f}$ and $\vec{g}$:
  \begin{align}
  	\begin{split}
    \vec{f} &= \begin{pmatrix} \vec{f}_1 ,\  \vec{f}_2 ,\    \cdots ,\   \vec{f}_C \end{pmatrix}^\top,\\ 
    \vec{g} & =\begin{pmatrix}
           \vec{g}_1 ,\
           \vec{g}_2,\
           \cdots ,\
           \vec{g}_C
         \end{pmatrix}^\top \enspace .
        \end{split}
	\label{eqn:variable}
  \end{align}
Then, the objective function in \eqref{eqn:objective3D} can be rewritten as
\begin{equation}
\begin{split}
\hat{\vec{f}} &= \argmin_{\vec{f}} \left\{ \mathcal{L}(\vec{f}, \vec{g})  
	 + \lambda \left\Vert  \vec{M} \left( \vec{f}- \vec{Q}\vec{f}\right) \right\Vert_1 \right\} \enspace ,
\end{split}
\label{eqn:objectiveMultiChannel}
\end{equation}
where $\vec{Q} = \text{diag}(\bar{\vec{Q}}, \ldots, \bar{\vec{Q}})$, $\vec{M} = \text{diag}(\vec{M}_1, \ldots, \vec{M}_C)$, and $\vec{M}_c = \text{diag}(m_c, \ldots, m_c)$ $\in \mathrm{R}^{CN \times CN}$. The optimization problem in \eqref{eqn:objectiveMultiChannel} can be solved using the algorithms described in Sec.~\ref{sec:AlternatingDirectionMethodOfMultipliersADMM}.

\begin{figure}[!tbp]
	\centering
	\subfloat[Ground truth]
	{\begin{tikzpicture}[spy using outlines={rectangle,orange,magnification=3, 
			height=1cm, width = 1cm, connect spies, every spy on node/.append style={thick}}]
		\node {\pgfimage[width = 0.232\textwidth]{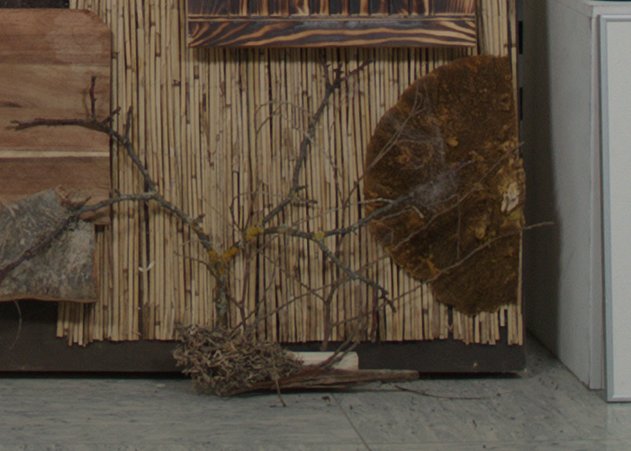}};
		\spy on (-0.5,0.3) in node [left] at (1.8, 0.8);   
		\end{tikzpicture}
		\label{fig:RGBNIR_detail_color}}
	\subfloat[Noisy image \newline ]
	{\begin{tikzpicture}[spy using outlines={rectangle,orange,magnification=3, 
			height=1cm, width = 1cm, connect spies, every spy on node/.append style={thick}}]
		\node {\pgfimage[width = 0.232\textwidth]{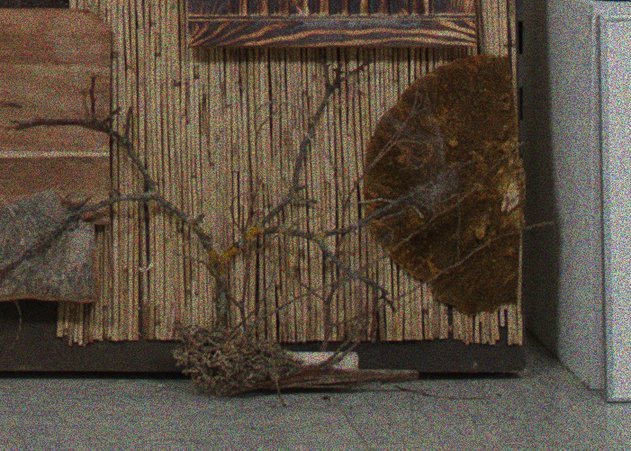}};
		\spy on (-0.5,0.3) in node [left] at (1.8, 0.8);
		\node[draw=none,overlay]  at (-0,-1.2){ \tiny{\color{white}(PSNR:~22.54.48 MS-SSIM:~0.86)}};   
		\end{tikzpicture}
		\label{fig:noisy_detail_color}}
	\subfloat[\aquasi channel-wise]
	{\begin{tikzpicture}[spy using outlines={rectangle,orange,magnification=3, 
			height=1cm, width = 1cm, connect spies, every spy on node/.append style={thick}}]
		\node {\pgfimage[width = 0.232\textwidth]{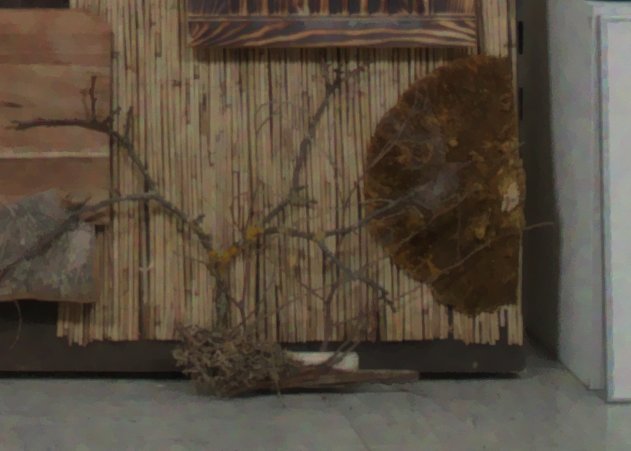}};
		\spy on (-0.5,0.3) in node [left] at (1.8, 0.8); 
			\node[draw=none,overlay]  at (-0,-1.2){ \tiny{\color{white}(PSNR:~31.08 MS-SSIM:~0.94)}};     
		\end{tikzpicture}
		\label{fig:ASCWG_detail_color}}
	\subfloat[\aquasi MC ]
	{\begin{tikzpicture}[spy using outlines={rectangle,orange,magnification=3, 
			height=1cm, width = 1cm, connect spies, every spy on node/.append style={thick}}]
		\node {\pgfimage[width = 0.232\textwidth]{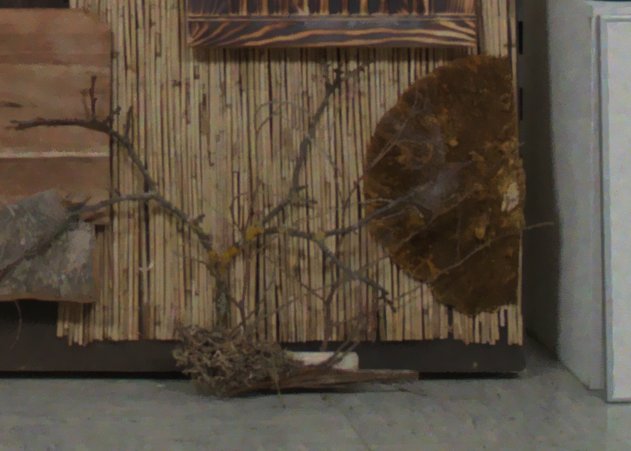}};
		\spy on (-0.5,0.3) in node [left] at (1.8, 0.8);  
			\node[draw=none,overlay]  at (-0,-1.2){ \tiny{\color{white}(PSNR:~33.05 MS-SSIM:~0.96)}};  
		\end{tikzpicture}	  
		\label{fig:AMCWG_detail_color}}
	\caption{RGB/NIR restoration on the ARRI dataset~\cite{Luethen2017}. \protect\subref{fig:RGBNIR_detail_color} Ground truth color image, \protect\subref{fig:noisy_detail_color} color image corrupted by speckle noise, \protect\subref{fig:ASCWG_detail_color} \aquasi channel-wise with \textit{static} guidance, and \protect\subref{fig:AMCWG_detail_color} \aquasi multi-channel with \textit{static} guidance.}
	\label{fig:CompareSCMC}
\end{figure}

We compare the channel-wise and multi-channel (MC) approach for RGB/NIR image restoration (see Sec.~\ref{sec:RGBAndNIRImageRestoration}) on an example of the ARRI dataset~\cite{Luethen2017} in Fig.~\ref{fig:CompareSCMC}. In general, the advantage of the multi-channel approach compared to a channel-wise optimization is two-fold: 1) \aquasi MC yields better structure preservation than \aquasi channel-wise. For example, the wooden sticks at the back of the image are less blurred and the gaps between the sticks are better visible. 2) \aquasi MC requires less computation time than simple channel-wise processing. This is explained by the fact that only a single linearization $\bar{\vec{Q}}$ is required per iteration. In contrast, channel-wise processing needs to construct one linearization per image channel. In Fig.~\ref{fig:CompareSCMC}, \aquasi MC converged after 388 seconds, whereas the channel-wise approach converged after 6678 seconds on an image of size $1575 \times 2800$ pixels.

\subsection{Implementation Remarks}
\label{sec:ImplementationRemarks}

The iterative algorithms with our \aquasi prior are easy to implement and can be seamlessly integrated into existing inverse imaging frameworks. There are two main steps at each iteration: 1) estimation of the image $\vec{f}^{t+1}$, and 2) the construction of the linearization $\vec{Q}$ for weighted quantile filtering $Q_{p,\vec{w}}(\vec{f})$. While the first step involves efficient matrix/vector operations, the computational complexity of an iteration is mainly related to the second step.

\begin{figure}[!tbp]
	
	\centering
	\scriptsize
	\subfloat{\includegraphics[width=0.5\linewidth]{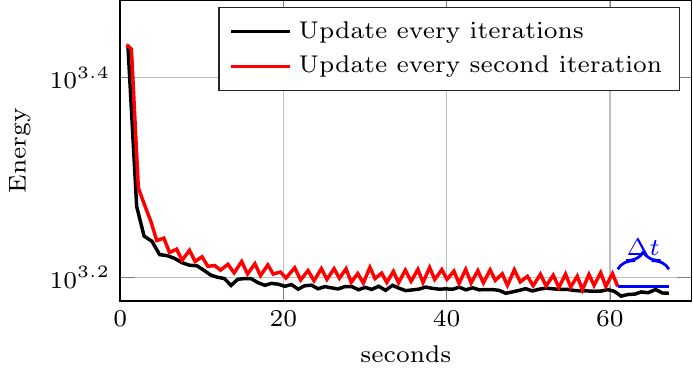}}
	\caption{Convergence analysis for iterative optimization with the \aquasi
		prior. We evaluated the overall energy function optimized by ADMM using an
		update of the \aquasi pseudo-linear form after every iteration and after
		every second iteration. The time gap $\Delta t$ indicates the
		performance improvement when updating every second iteration for images of size $256 \times 256$ pixels. This
		improvement increases with the image resolution.}
	\label{fig:RGBNIRConvergence}
\end{figure}

To decrease the computation time, we found that it is sufficient to re-use the same $\vec{Q}$ for several iterations instead of updating  $\vec{Q}$ in every iteration. In this case, the pseudo-linear form is an approximation of the actual filtering. We validate this approximation experimentally by examining the convergence of our iteration scheme. Figure~\ref{fig:RGBNIRConvergence} depicts the convergence of ADMM in terms of the optimized energy for RGB/NIR restoration (see Sec.~\ref{sec:RGBAndNIRImageRestoration}) versus the computation time of the algorithm. We evaluated the convergence on image patches of size $256 \times 256$ pixels.
Here, we compare an update of $\vec{Q}$ at every iteration against an update after every second iteration. It is worth noting that both variants convergences to comparable local minima, while the approximation is considerably faster in computation. Especially for large-scale images, significantly lower processing time are possible. 

\section{Experimental Evaluation and Applications}
\label{sec:Applications}

In order to emphasize the effectiveness of the \aquasi prior, we relate it to TV regularization and RED algorithms. Additionally, we evaluate the performance of the proposed method on two common inverse problems in computer vision, namely joint RGB/depth map upsampling and RGB/NIR image restoration. 

\begin{figure}[!t]
	\centering
	\scriptsize	
	\includegraphics[width=0.5\linewidth]{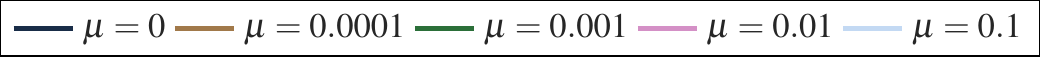}\\

	\subfloat[TV parameter sensitivity]
	{\includegraphics[width=0.35\linewidth]{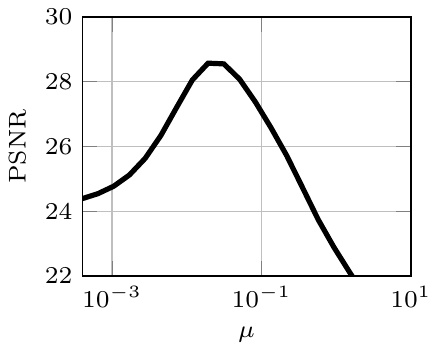}}~
	\subfloat[ \aquasi + TV parameter sensitivity]
	{\includegraphics[width=0.35\linewidth]{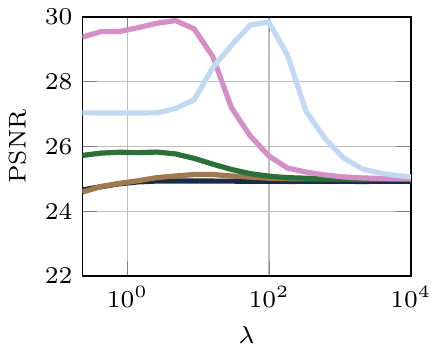}}
	\caption{Parameter sensitivity analysis for the TV prior with regularization weight $\mu$ (a) and the combination of \aquasi with regularization weight $\lambda$ and TV (b) on the RGB/NIR image restoration result.}
	\label{fig:TVvsAQuaSI}
\end{figure}

\begin{figure}[!t]
	\centering
	\subfloat[Input]
	{\includegraphics[width=0.229\textwidth]{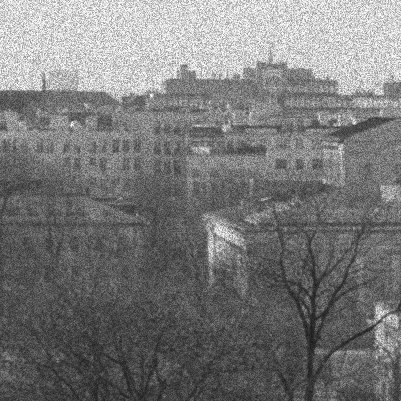}}\quad
	\subfloat[TV prior without \aquasi]
	{\includegraphics[width=0.229\textwidth]{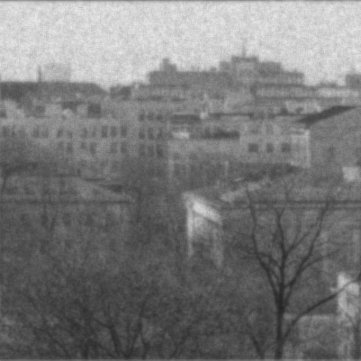}}\quad
	\subfloat[\aquasi prior without TV]
	{\includegraphics[width=0.229\textwidth]{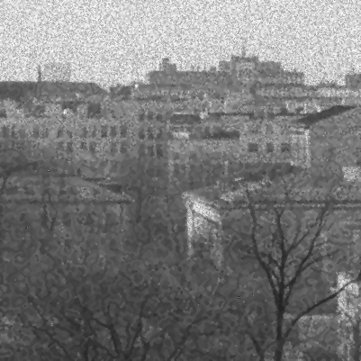}}\quad
	\subfloat[\aquasi + TV prior]
	{\includegraphics[width=0.229\textwidth]{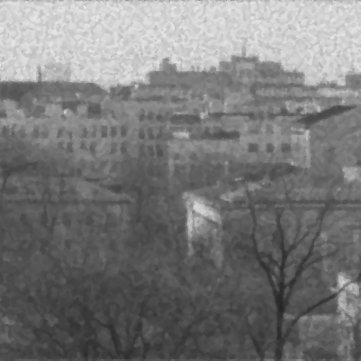}\label{fig:TVandAQuaSI}}
	\caption{Influence of the TV prior and the \aquasi prior to image denoising. We compare a noisy input (a) to denoised images using only the TV prior (b), only \aquasi (c), as well as the combination TV + \aquasi (d). For each configuration, we report denoised images with the highest PSNR.}
	\label{fig:TVvsAQuaSI_example}
\end{figure}

\subsection{Comparison to Total Variation Regularization}
\label{sec:ComparisonToTotalVariationPrior}

	We investigate the influence of the \aquasi prior with \textit{dynamic} guidance for RGB/NIR image restoration (see Sec.~\ref{sec:RGBAndNIRImageRestoration}) and compare it to the influence of the TV prior. We varied the regularization weights of the TV and \aquasi prior in \eqref{eqn:crossFieldInvProblem} separately, while setting the other one to zero. We additionally combined \aquasi with TV for varying regularization weight of the TV prior. 
	
Figure~\ref{fig:TVvsAQuaSI} shows the peak
signal-to-noise ratio (PSNR) for TV and \aquasi with different regularization weights. 
A small regularization weight $\mu$ for the TV prior effectively removes the
term from the optimization. Increasing $\mu$ for the TV prior beyond the
optimum point drastically decreases the performance again, since TV starts to
oversmooth the results. Changing the weight for \aquasi from $\lambda = 0$ to a larger value
corresponds to activating it. The jump in the activation can be steered by the choice of $\sigma_w$, the standard deviation of the Gaussian weighting function. \aquasi smooths the image with large $\sigma_w$ and allows for better structure preservation with smaller $\sigma_w$. For this experiment, we choose $n=3^2$ and $\sigma_w = 0.1$, which is rather low. Thus, compared to TV, \aquasi does not achieve a competitive denoising performance. The advantage of this configuration of \aquasi can be seen in combination with the TV prior in Fig.~\ref{fig:TVvsAQuaSI_example}. The smoothing property of TV is retained and \aquasi helps to preserve the original image content. This property of \aquasi is underlined in the following sections, where \aquasi drastically improves the performance of common image priors. We provide additional qualitative results for comparison to TV regularization in the supplementary material.

\subsection{Comparison to Regularization by Denoising (RED)}
\label{sec:ComparisonToRED}

\aquasi is similar in spirit to the RED framework proposed by Romano \etal~\cite{Romano2016a}. Using a weighted median filter $Q_{p, \vec{w}}(\cdot)$, $p = 0.5$ as denoising engine in the same variant as in the proposed prior, RED is modeled by the regularization term
\begin{equation}
	R_{\text{RED}}(\vec{f}) = \vec{f}^{\top} ( \vec{f} - Q_{p, \vec{w}}(\vec{f}) )\enspace.
\end{equation}

That is, instead of regularizing with the $L_1$ norm, RED uses the square of the image $\vec{f}$. For a comparison between these models, we apply both on denoising and deblurring tasks.

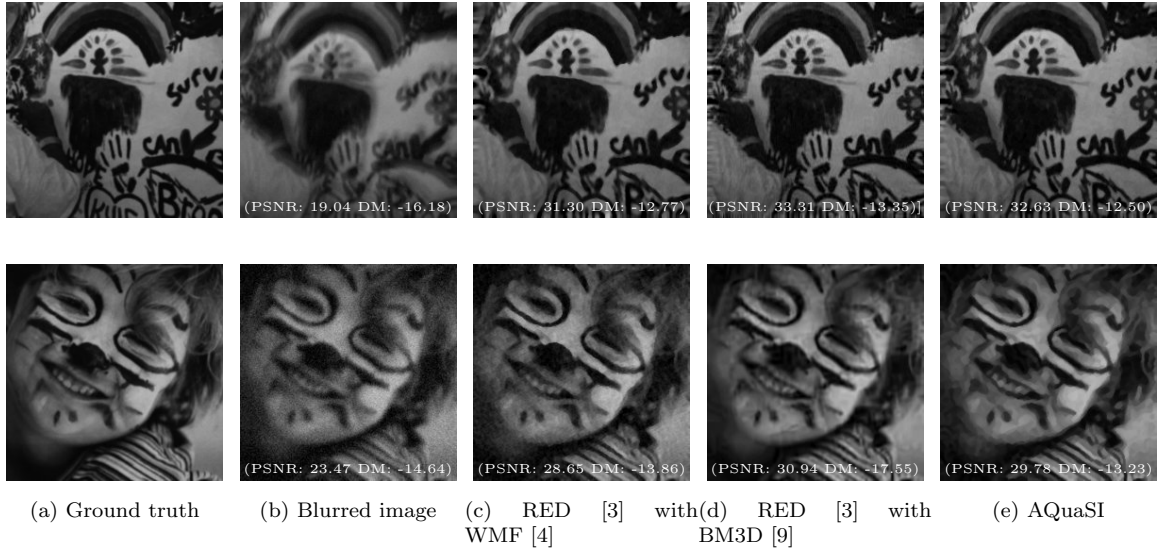
\begin{figure}	
	\centering
	\subfloat{\begin{tikzpicture}
		\node {\pgfimage[width=0.188\linewidth]{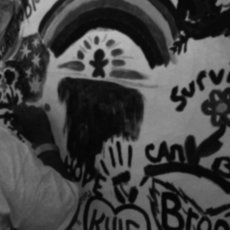}};
		\end{tikzpicture}}	
	\subfloat{\begin{tikzpicture}
		\node {\pgfimage[width=0.188\linewidth]{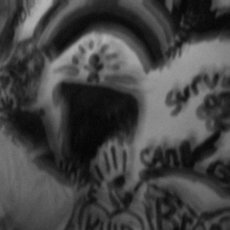}};		
		\node[draw=none,overlay]  at (-0,-1.3){\tiny {\color{white} (PSNR:~19.04 DM:~-16.18)}};    
		\end{tikzpicture}}	
	\subfloat
	{\begin{tikzpicture}
		\node {\pgfimage[width=0.188\linewidth]{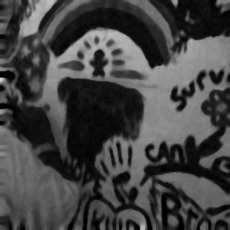}};		
		\node[draw=none,overlay]  at (-0,-1.3){\tiny {\color{white} (PSNR:~31.30 DM:~-12.77)}};    
		\end{tikzpicture}}	
	\subfloat{\begin{tikzpicture}
		\node {\pgfimage[width=0.188\linewidth]{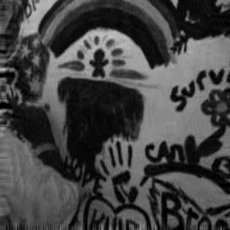}};		
		\node[draw=none,overlay]  at (-0,-1.3){\tiny {\color{white} (PSNR:~33.31 DM:~-13.35)]}};    
		\end{tikzpicture}}	
	\subfloat
	{\begin{tikzpicture}
		\node {\pgfimage[width=0.188\linewidth]{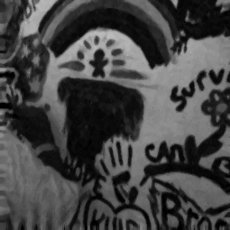}};		
		\node[draw=none,overlay]  at (-0,-1.3){\tiny {\color{white} (PSNR:~32.63 DM:~-12.50)}};    
		\end{tikzpicture}}\\
	\setcounter{subfigure}{0}	
	\subfloat[Ground truth]{\begin{tikzpicture}
		\node {\pgfimage[width=0.188\linewidth]{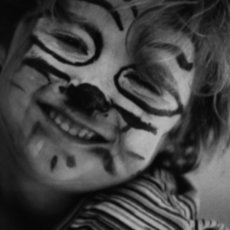}};
		\end{tikzpicture}}
	\subfloat[Blurred image]{\begin{tikzpicture}
		\node {\pgfimage[width=0.188\linewidth]{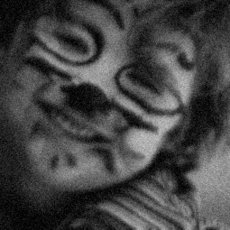}};		
		\node[draw=none,overlay]  at (-0,-1.3){\tiny {\color{white}(PSNR:~23.47 DM:~-14.64)}};    
		\end{tikzpicture}}
	\subfloat[RED~\cite{Romano2016a} with WMF~\cite{Zhang2014}]{\begin{tikzpicture}
		\node {\pgfimage[width=0.188\linewidth]{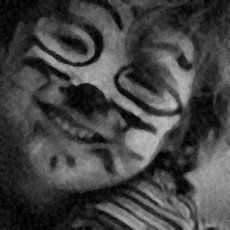}};		
		\node[draw=none,overlay]  at (-0,-1.3){\tiny {\color{white}(PSNR:~28.65 DM:~-13.86)}};    
		\end{tikzpicture}}
	\subfloat[RED~\cite{Romano2016a} with BM3D~\cite{Dabov2007a}]{\begin{tikzpicture}
		\node {\pgfimage[width=0.188\linewidth]{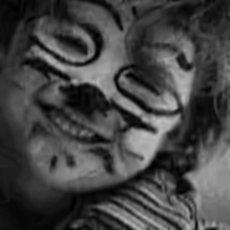}};		
		\node[draw=none,overlay]  at (-0,-1.3){\tiny {\color{white}(PSNR:~30.94 DM:~-17.55)}};  
		\end{tikzpicture}}
	\subfloat[\aquasi]{\begin{tikzpicture}
		\node {\pgfimage[width=0.188\linewidth]{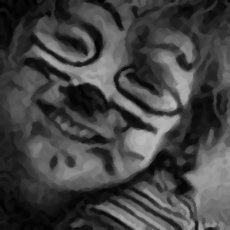}};		
		\node[draw=none,overlay]  at (-0,-1.3){\tiny {\color{white} (PSNR:~29.78 DM:~-13.23)}};    
		\end{tikzpicture}}
	\caption{Qualitative deblurring results for two images of the Levin dataset \cite{Levin2009}. The top row shows ground truth data, blurry input data corrupted by additive white Gaussian noise with $\sigma_{\text{noise}} = 0.0001$ and deblurring results of different algorithms. The bottom row depicts qualitative results for $\sigma_{\text{noise}} = 0.025$. }
	\label{fig:Deblurring_qualitative}
\end{figure}

We evaluated \aquasi and RED for non-blind deblurring using the dataset of Levin \etal \cite{Levin2009} comprising eight blur kernels and four ground truth grayscale images. We degraded the ground truths by convolutions with the known blur kernels. In addition, each blurred image is corrupted by additive white Gaussian noise and speckle noise ($\sigma_{\text{noise}} \in \{0.0001, 0.009, 0.0025\}$ for each model and images in $[0, 1]$). For deblurring, we applied \aquasi with \textit{dynamic} guidance and RED with the weighted median filter ($n = 5^2$, $\sigma_w = 0.6$) and BM3D in the framework provided by Romano \etal~\cite{Romano2016a}. The deblurring quality was assessed by the PSNR and the no-reference motion deblurring metric (DM) \cite{Liu2013}. While PSNR measures distortions relative to the ground truth, DM is specifically designed to assess the perceptual quality of the deblurring methods and lower values express better performance. Quantitative results for both noise models are reported in Tab.~\ref{tab:Deblurring_quantitative}. RED with BM3D outperforms \aquasi \wrt PSNR. However \aquasi excels in the deblurring metric DM. Here, \aquasi overall outperforms RED with WMF and RED with BM3D especially under heavy speckle noise that is difficult to handle by of-the-shelf denoisers. Analogously to the other image restoration problems (cf. Sec.~\ref{sec:RGBAndNIRImageRestoration}), there is an inherent tradeoff between signal distortion and perceptual quality \cite{Blau2018}. In our experiments, \aquasi performs reasonably in terms of distortions (second best in PSNR) while providing excellent perceptual quality (best in DM) leading to good perception-distortion tradeoffs. This is also confirmed in the qualitative results in Fig.~\ref{fig:Deblurring_qualitative}.

\begin{table}[!tbp]
	\centering
	\caption{Averaged peak signal to noise ratio (PSNR) and no-reference deblurring metric (DM) \cite{Liu2013} for non-blind deblurring using RED~\cite{Romano2016a} with WMF \cite{Zhang2014} or BM3D denoising \cite{Dabov2007a}, and \aquasi on the Levin dataset \cite{Levin2009}. Results are reported for different noise levels of additive white Gaussian noise (AWGN) and speckle noise (SN). }
	\begin{tabular}{c p{0.9cm} p{0.6cm}c p{0.6cm}c p{0.6cm}c p{0.6cm}c p{0.6cm}c p{0.6cm}c}
		\toprule
		& Level & \multicolumn{2}{c}{RED with WMF} & \multicolumn{2}{c}{RED with BM3D} & \multicolumn{2}{c}{\aquasi}	\\		
		& $\sigma_{\text{noise}}$	& PSNR & DM & PSNR & DM	& PSNR & DM \\
		\midrule
		\multirow{3}{*}{\rotatebox[origin=c]{90}{AWGN}} 
		& $0.0001$			& 29.40		& -13.96 					& \textbf{30.62}		& -14.11			&  30.47 					& \textbf{-12.94}\\
		& $0.0009$			& 27.12  	& -14.02  					& \textbf{28.41}		& -14.60	 		& 27.51 					& \textbf{-12.77}\\
		& $0.0025$			& 25.87 	&\textbf{-13.10}   & \textbf{27.32}		& -15.46			&	26.34						& -13.36 \\				
		\midrule
		\multirow{3}{*}{\rotatebox[origin=c]{90}{SN}}
		& $0.0001$			& 33.69		& -13.13 					& \textbf{35.04}		& -13.48			& 33.92 					& \textbf{-12.47}\\
		& $0.0009$			& 30.04  	& -13.05  					& \textbf{30.87}		& -14.23			& 30.25 					& \textbf{-12.30}\\
		& $0.0025$			& 28.41 	& -12.89   				& 28.58							& -14.93			&	\textbf{28.76}	& \textbf{-12.10} \\		
		\bottomrule
	\end{tabular}
	\label{tab:Deblurring_quantitative}
\end{table}

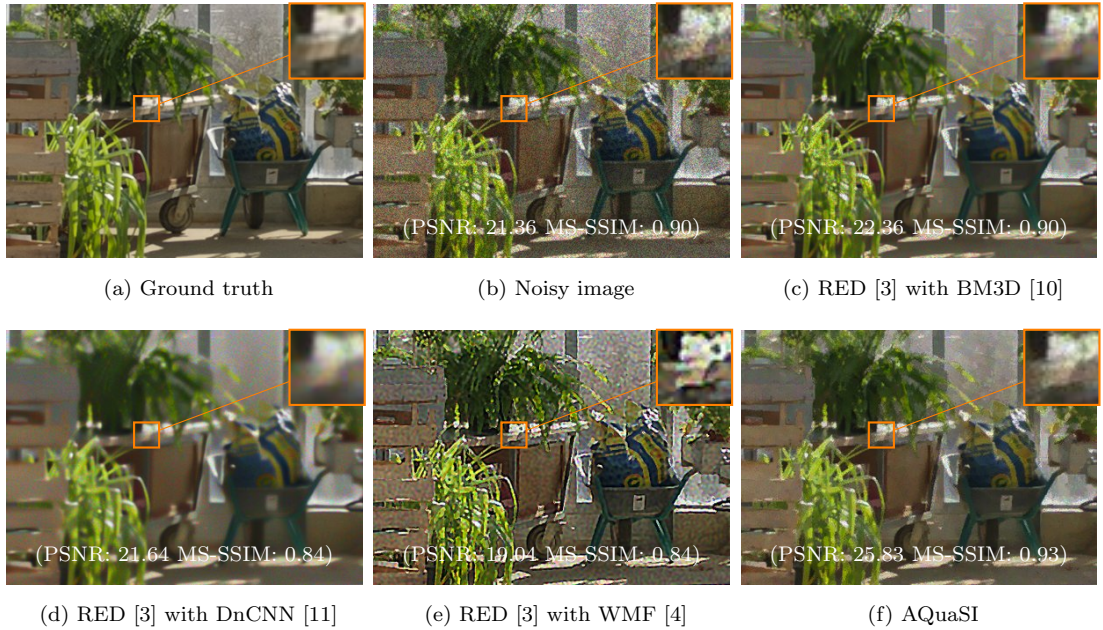
\begin{figure}[!t]
	\centering
	\subfloat[Ground truth ]
	{\begin{tikzpicture}[spy using outlines={rectangle,orange,magnification=3, 
			height=1cm, width = 1cm, connect spies, every spy on node/.append style={thick}}]
		\node {\pgfimage[width = 0.31\textwidth]{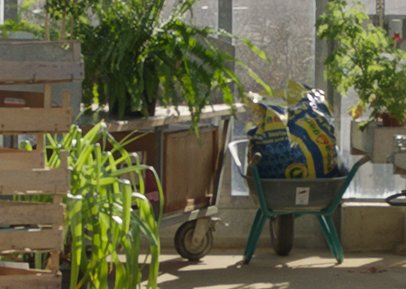}};
		\spy on (-0.5,0.3) in node [left] at (2.4, 1.2);   
		\end{tikzpicture}
		\label{fig:RGBNIR_flower}}\hspace{-0.9em}
	\subfloat[Noisy image]
	{\begin{tikzpicture}[spy using outlines={rectangle,orange,magnification=3, 
			height=1cm, width = 1cm, connect spies, every spy on node/.append style={thick}}]
		\node {\pgfimage[width = 0.31\textwidth]{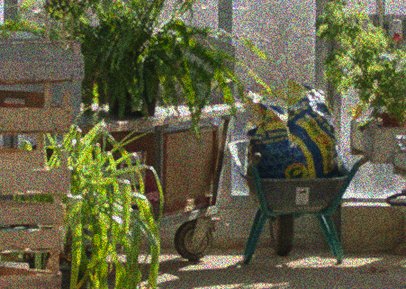}};
		\spy on (-0.5,0.3) in node [left] at (2.4, 1.2);
		\node[draw=none,overlay]  at (-0,-1.3){ \footnotesize {\color{white} (PSNR:~21.36 MS-SSIM:~0.90)}}; 
		\end{tikzpicture}
		\label{fig:noisy_flower}}\hspace{-0.9em}
	\subfloat[RED~\cite{Romano2016a} with BM3D \cite{Dabov2007} ]
	{\begin{tikzpicture}[spy using outlines={rectangle,orange,magnification=3, 
			height=1cm, width = 1cm, connect spies, every spy on node/.append style={thick}}]
		\node {\pgfimage[width = 0.31\textwidth]{RED_BM3D_flower}};
		\spy on (-0.5,0.3) in node [left] at (2.4, 1.2);   
		\node[draw=none,overlay]  at (-0,-1.3){ \footnotesize {\color{white}  (PSNR:~22.36 MS-SSIM:~0.90)}}; 
		\end{tikzpicture}
		\label{fig:RED_BM3D_flower}}\\[-0.8ex]
	\subfloat[RED~\cite{Romano2016a} with DnCNN~\cite{Zhang2017beyond} ]
	{\begin{tikzpicture}[spy using outlines={rectangle,orange,magnification=3, 
			height=1cm, width = 1cm, connect spies, every spy on node/.append style={thick}}]
		\node {\pgfimage[width = 0.31\textwidth]{RED_DnCNN_flower}};
		\spy on (-0.5,0.3) in node [left] at (2.4, 1.2); 
		\node[draw=none,overlay]  at (-0,-1.3){ \footnotesize {\color{white} (PSNR:~21.64 MS-SSIM:~0.84)}}; 
		\end{tikzpicture}
		\label{fig:RED_DnCNN_flower}}\hspace{-0.9em}
	\subfloat[RED~\cite{Romano2016a} with WMF~\cite{Zhang2014} ]
	{\begin{tikzpicture}[spy using outlines={rectangle,orange,magnification=3, 
			height=1cm, width = 1cm, connect spies, every spy on node/.append style={thick}}]
		\node {\pgfimage[width = 0.31\textwidth]{RED_flower}};
		\spy on (-0.5,0.3) in node [left] at (2.4, 1.2); 
		\node[draw=none,overlay]  at (-0,-1.3){ \footnotesize {\color{white}  (PSNR:~19.04 MS-SSIM:~0.84)}};  
		\end{tikzpicture}
		\label{fig:RED_flower}}\hspace{-0.9em}
	\subfloat[\aquasi]
	{\begin{tikzpicture}[spy using outlines={rectangle,orange,magnification=3, 
			height=1cm, width = 1cm, connect spies, every spy on node/.append style={thick}}]
		\node {\pgfimage[width = 0.31\textwidth]{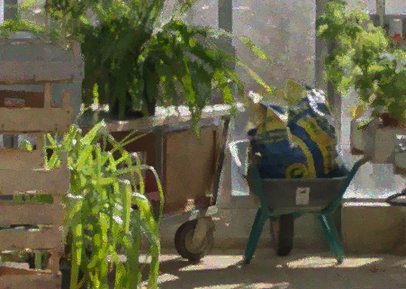}};
		\spy on (-0.5,0.3) in node [left] at (2.4, 1.2);
		\node[draw=none,overlay]  at (-0,-1.3){ \footnotesize {\color{white}(PSNR:~25.83 MS-SSIM:~0.93)}}; 
		\end{tikzpicture}
		\label{fig:AMCWG_flower}}
	\caption{RGB/NIR restoration using our \aquasi prior under non-Gaussian noise in comparison to regularization by denoising (RED)~\cite{Romano2016a}.\protect\subref{fig:RGBNIR_flower} Ground truth color image, \protect\subref{fig:noisy_flower} color image corrupted by speckle noise, \protect\subref{fig:RED_flower} RED with weighted median filter (WMF), and \protect\subref{fig:AMCWG_flower} \aquasi prior. The priors are steered by NIR data in a \textit{static} guidance mode. The reported PSNR and MS-SSIM values are calculated from the sections to be seen.}
	\label{fig:CompareAQuaSIRED}
\end{figure}

Figure~\ref{fig:CompareAQuaSIRED} depicts RED versus the proposed \aquasi prior for RGB/NIR image restoration (see Sec.~\ref{sec:RGBAndNIRImageRestoration}). Here, NIR data guides the restoration of noisy RGB images that are degraded by speckle noise ($\sigma_{\text{noise}} = 0.2$). For fair comparison we set $\mu = 0$, thus eliminate the influence of the TV prior. We couple both priors with an $L_2$ norm data term and use the weighted median filter ($n = 9^2$, $\sigma_w = 0.006$) in \textit{static} guidance mode within ADMM optimization. It is worth noting that \aquasi yields considerably better perceptual quality in terms of noise reduction as well as a higher PSNR. We explain this behavior by the different models underlying the priors: RED with its quadratic model has difficulties in handling non-Gaussian noise on its own, while \aquasi with the $L_1$ norm model is less sensitive in that case. This also confirms the properties of our sparsity-promoting model shown in Sec.~\ref{sec:InfluenceOfImageDegradationsToTheAQuaSIPrior}. We additionally report the results of RED without guidance data using BM3D~\cite{Dabov2007}, DnCNN~\cite{Zhang2017beyond} as denoiser. While DnCNN tends to oversmooth the result, BM3D provides a better tradeoff between noise reduction and structure preservation. 

\subsection{Joint Upsampling}
\label{sec:JointUpsampling}

\begin{figure}[!tbp]
	\centering
	\footnotesize
	\subfloat{\includegraphics[width=0.95\linewidth]{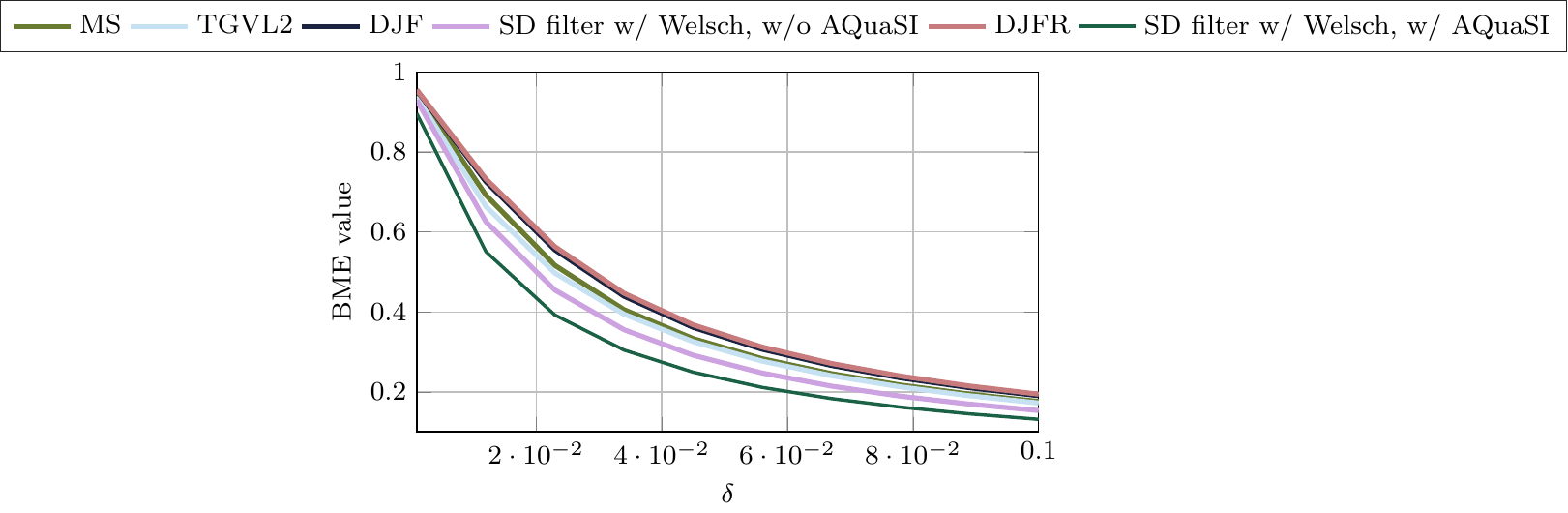}}
	\caption{Bad matching error (BME) of various joint RGB/depth map upsampling methods at different error thresholds $\delta$ on the Middlebury data.}
	\label{fig:BME}
\end{figure}

\begin{figure}[!tbp]
	\centering
	\subfloat[Ground truth]
	{\begin{tikzpicture}[spy using outlines={rectangle,orange,magnification=3, 
			height=1cm, width = 1cm, connect spies, every spy on node/.append style={thick}}]
		\node {\pgfimage[width = 0.23\textwidth]{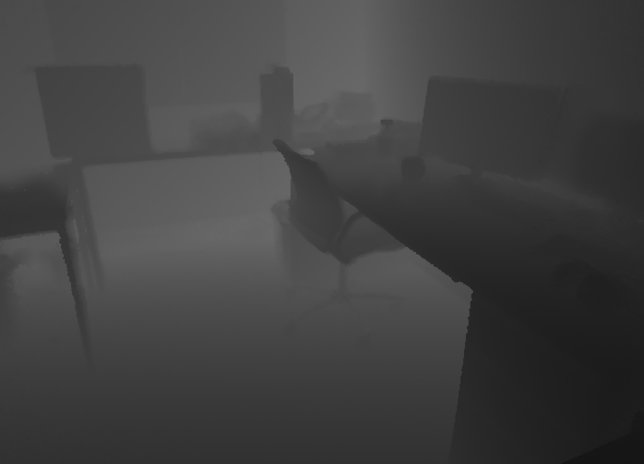}};
		\spy on (0.2,0.6) in node [left] at (-0.8, 0.8);		  
		\end{tikzpicture}\label{fig:jointUpsamplingGt}}
	\subfloat[Degraded depth map]
	{\begin{tikzpicture}[spy using outlines={rectangle,orange,magnification=3, 
			height=1cm, width = 1cm, connect spies, every spy on node/.append style={thick}}]
		\node {\pgfimage[width = 0.23\textwidth]{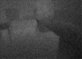}};
	\spy on (0.2,0.6) in node [left] at (-0.8, 0.8);			  
		\end{tikzpicture}\label{fig:jointUpsamplingNoisy}}
	\subfloat[Color image]	
	{\begin{tikzpicture}[spy using outlines={rectangle,orange,magnification=3, 
			height=1cm, width = 1cm, connect spies, every spy on node/.append style={thick}}]
		\node {\pgfimage[width = 0.23\textwidth]{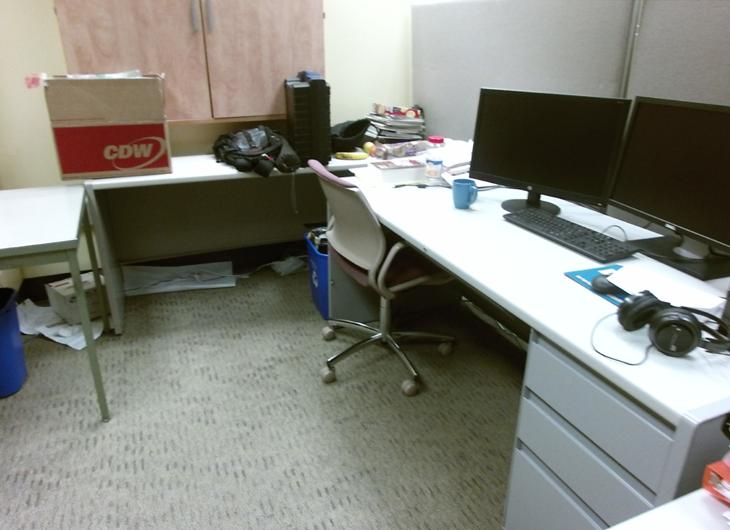}};
		\spy on (0.2,0.6) in node [left] at (-0.8, 0.8);
		\end{tikzpicture}\label{fig:jointUpsamplingRGB}}	
	\subfloat[MS~\cite{Shen2015} ]
	{\begin{tikzpicture}[spy using outlines={rectangle,orange,magnification=3, 
			height=1cm, width = 1cm, connect spies, every spy on node/.append style={thick}}]
		\node {\pgfimage[width = 0.23\textwidth]{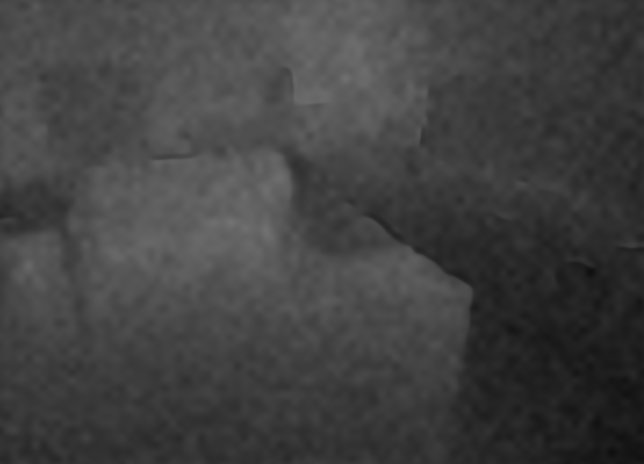}};
		\spy on (0.2,0.6) in node [left] at (-0.8, 0.8);	
		\node[draw=none,overlay]  at (-0,-1.1){\footnotesize {\color{white}(RMSE:~0.0152 BME: 0.50)}};    	  
		\end{tikzpicture}\label{fig:jointUpsamplingMS}}\\[-0.8ex]
	\subfloat[TGVL2~\cite{Ferstl2013} ]
	{\begin{tikzpicture}[spy using outlines={rectangle,orange,magnification=3, 
			height=1cm, width = 1cm, connect spies, every spy on node/.append style={thick}}]
		\node {\pgfimage[width = 0.23\textwidth]{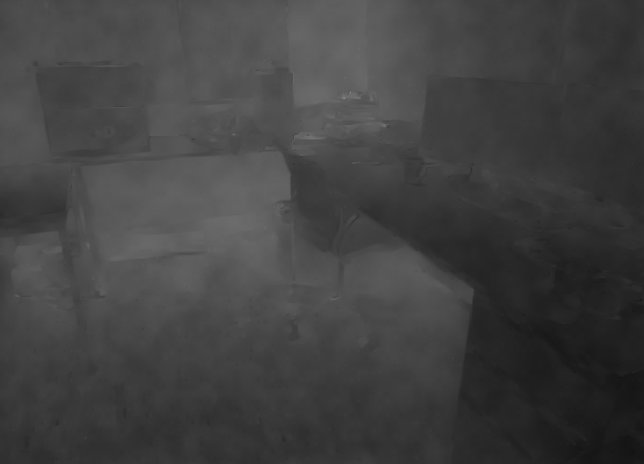}};
		\spy on (0.2,0.6) in node [left] at (-0.8, 0.8);
		\node[draw=none,overlay]  at (-0,-1.1){\footnotesize {\color{white}(RMSE:~0.0133 BME: 0.36)}}; 	  
		\end{tikzpicture}\label{fig:jointUpsamplingTGVL2}}
	\subfloat[DJF~\cite{Li2016} ]
	{\begin{tikzpicture}[spy using outlines={rectangle,orange,magnification=3, 
			height=1cm, width = 1cm, connect spies, every spy on node/.append style={thick}}]
		\node {\pgfimage[width = 0.23\textwidth]{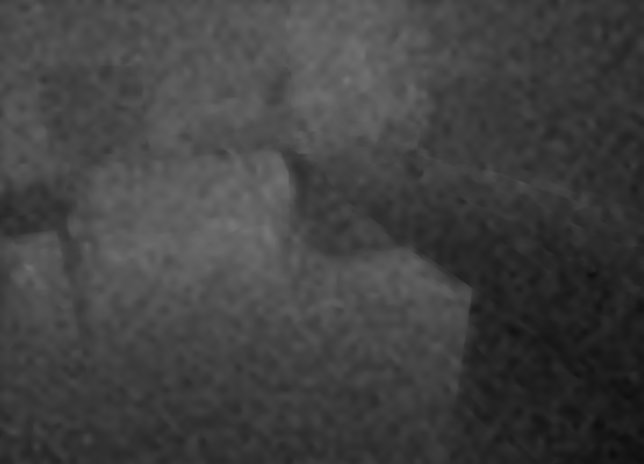}};
		\spy on (0.2,0.6) in node [left] at (-0.8, 0.8);
		\node[draw=none,overlay]  at (-0,-1.1){\footnotesize {\color{white}(RMSE:~0.0178 BME: 0.57)}}; 			  
		\end{tikzpicture}\label{fig:jointUpsamplingDJF}}
	\subfloat[DJFR~\cite{Li2019} ]
	{\begin{tikzpicture}[spy using outlines={rectangle,orange,magnification=3, 
			height=1cm, width = 1cm, connect spies, every spy on node/.append style={thick}}]
		\node {\pgfimage[width = 0.23\textwidth]{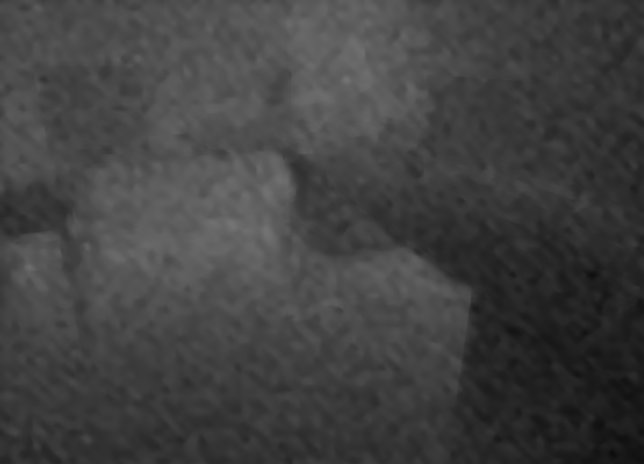}};
		\spy on (0.2,0.6) in node [left] at (-0.8, 0.8);
		\node[draw=none,overlay]  at (-0,-1.1){\footnotesize {\color{white}(RMSE:~0.0178 BME: 0.57)}}; 			  
		\end{tikzpicture}\label{fig:jointUpsamplingDJFR}}	
	\subfloat[SD filter~\cite{Ham2015} w/ Welsch, w/ \aquasi][\centering SD filter~\cite{Ham2015} \linebreak w/ Welsch, w/ \aquasi ]	
{\begin{tikzpicture}[spy using outlines={rectangle,orange,magnification=3, 
		height=1cm, width = 1cm, connect spies, every spy on node/.append style={thick}}]
	\node {\pgfimage[width = 0.23\textwidth]{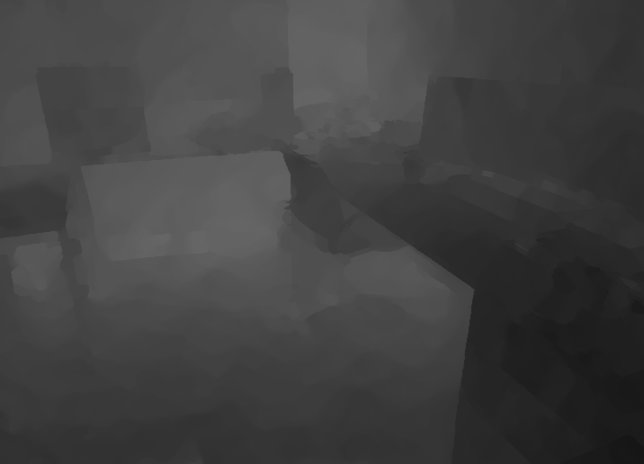}};
\spy on (0.2,0.7) in node [left] at (-0.8, 0.8);
\node[draw=none,overlay]  at (-0,-1.1){\footnotesize {\color{white}(RMSE:~0.0113 BME: 0.25)}};   
	\end{tikzpicture}\label{fig:jointUpsamplingAQuaSI}}
	
	\caption{Joint RGB/depth map upsampling on a SUN image pair. \protect\subref{fig:jointUpsamplingGt} Ground truth depth map, \protect\subref{fig:jointUpsamplingNoisy} degraded depth map, \protect\subref{fig:jointUpsamplingRGB} color image, \protect\subref{fig:jointUpsamplingMS} - \protect\subref{fig:jointUpsamplingAQuaSI} results of mutual structure (MS) filter, TGVL2 upsampling, deep joint filter (DJF), residual-based deep joint filter (DJFR), and SD filter w/ Welsch, w/ \aquasi. RMSE and BME, with $\delta = 0.01$, are reported for each result.} 
	\label{fig:jointUpsampling}
\end{figure}

\begin{figure}[!tbp]
	\centering
	\subfloat[SD filter~\cite{Ham2015} w/ Welsch, w/o \aquasi][\centering SD filter~\cite{Ham2015} \linebreak w/ Welsch, w/o \aquasi]
{\begin{tikzpicture}[spy using outlines={rectangle,orange,magnification=3, 
		height=1cm, width = 1cm, connect spies, every spy on node/.append style={thick}}]
	\node {\pgfimage[width = 0.23\textwidth]{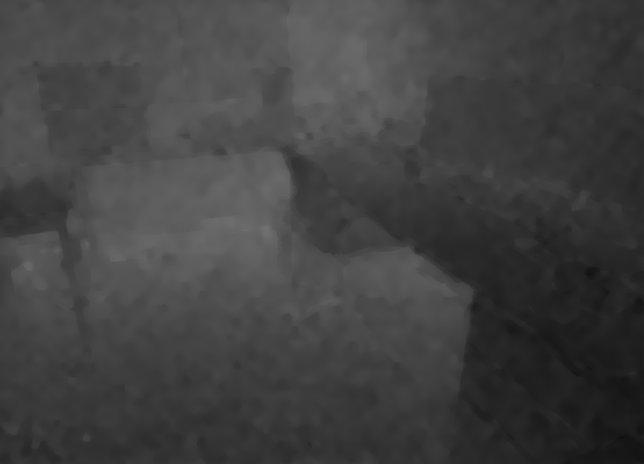}};
	\spy on (0.95,-0.37) in node [left] at (1.8, 0.8);
	\node[draw=none,overlay]  at (-0,-1.15){\footnotesize {\color{white}(RMSE:~0.0128 BME: 0.38)}}; 		  
	\end{tikzpicture}\label{fig:jointUpsamplingSD}}	
\subfloat[SD filter~\cite{Ham2015} w/o Welsch, w/ \aquasi][\centering SD filter~\cite{Ham2015} \linebreak w/o Welsch, w/ \aquasi]		
{\begin{tikzpicture}[spy using outlines={rectangle,orange,magnification=3, 
		height=1cm, width = 1cm, connect spies, every spy on node/.append style={thick}}]
	\node {\pgfimage[width = 0.23\textwidth]{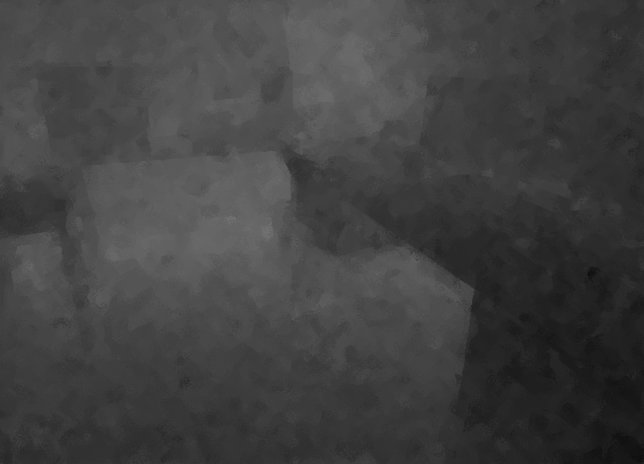}};
	\spy on (0.95,-0.37) in node [left] at (1.8, 0.8);
	\node[draw=none,overlay]  at (-0,-1.15){\footnotesize {\color{white} (RMSE:~0.0161 BME: 0.50)}}; 		  
	\end{tikzpicture}\label{fig:jointUpsamplingAQuaSIonly}}
\subfloat[SD filter~\cite{Ham2015} w/ Welsch, w/ QuaSI][\centering SD filter~\cite{Ham2015} \linebreak w/ Welsch, w/ QuaSI]		
{\begin{tikzpicture}[spy using outlines={rectangle,orange,magnification=3, 
		height=1cm, width = 1cm, connect spies, every spy on node/.append style={thick}}]
	\node {\pgfimage[width = 0.23\textwidth]{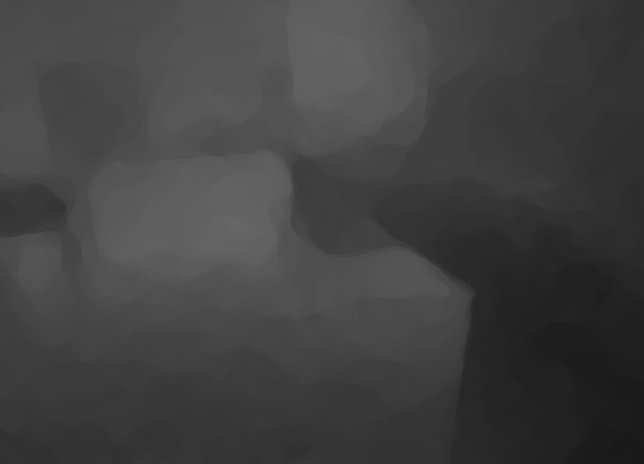}};
	\spy on (0.95,-0.37) in node [left] at (1.8, 0.8);
	\node[draw=none,overlay]  at (-0,-1.15){\footnotesize {\color{white}(RMSE:~0.0114 BME: 0.27)}}; 		  
	\end{tikzpicture}\label{fig:jointUpsamplingQuaSI}}
\subfloat[SD filter~\cite{Ham2015} w/ Welsch, w/ \aquasi][\centering SD filter~\cite{Ham2015} \linebreak w/ Welsch, w/ \aquasi]		
{\begin{tikzpicture}[spy using outlines={rectangle,orange,magnification=3, 
		height=1cm, width = 1cm, connect spies, every spy on node/.append style={thick}}]
	\node {\pgfimage[width = 0.23\textwidth]{Sun_aquasi.jpg}};
	\spy on (0.95,-0.37) in node [left] at (1.8, 0.8);
	\node[draw=none,overlay]  at (-0,-1.15){\footnotesize {\color{white}(RMSE:~0.0113 BME: 0.25)}}; 		  
	\end{tikzpicture}\label{fig:jointUpsamplingAQuaSI2}}
	\caption{Comparison of the different variants of the SD filter~\cite{Ham2015} on a SUN image pair. \protect\subref{fig:jointUpsamplingSD} SD w/ Welsch, w/o \aquasi, \protect\subref{fig:jointUpsamplingAQuaSIonly} SD filter w/o Welsch, w/ \aquasi, \protect\subref{fig:jointUpsamplingQuaSI} SD filter w/ Welsch, w/ QuaSI, \protect\subref{fig:jointUpsamplingAQuaSI2} SD filter w/ Welsch, w/ \aquasi. RMSE and BME with $\delta = 0.01$ are reported for each result.} 
\label{fig:jointUpsamplingSDvariants}
\end{figure}

In joint upsampling, we aim at upsampling a noisy depth map under the guidance of an RGB image. To demonstrate the applicability of the proposed method for this task, we augment the popular joint static and dynamic (SD) filter~\cite{Ham2015} by our \aquasi prior. Given a noisy depth map $\vec{g}$ and a complementary RGB color image $\vec{z}$, we obtain an upsampled depth map $\vec{f}$ according to the inverse problem
\begin{equation}
	\label{eqn:jointUpsamplingInvProblem}
	\hat{\vec{f}} = \argmin_{\vec{f}}
	\mathcal{L}(\vec{f}, \vec{g}) 
	+ \underbrace{\mu R(\vec{f}, \vec{z}) }_{\text{SD prior}}
	+ \underbrace{\lambda \|\vec{f} - Q_{p,\vec{w}}(\vec{f})\|_1}_{\text{\aquasi prior}}
	\enspace,
\end{equation}
where $R(\vec{f}, \vec{z})$ denotes the Welsch function as proposed in~\cite{Ham2015} with regularization weight $\mu \geq 0$. The data fidelity term is defined analogously to~\cite{Ham2015} as $\mathcal{L}(\vec{f}, \vec{g}) = \| \vec{c} \odot (\vec{f} - \vec{g}) \|_2^2$,
where $\vec{c}$ denotes a confidence map to mask out outliers, \eg, invalid depth measurements, and $\odot$ is the Hadamard (element-wise) product. In order to solve \eqref{eqn:jointUpsamplingInvProblem}, we perform iteratively reweighted least squares as in~\cite{Ham2015}, which essentially is a variant of gradient descent as introduced in Sec.~\ref{sec:GradientDescent}.

\begin{table}[!tbp]
	\centering
	\caption{Averaged root mean square error (RMSE) for RGB/depth map upsampling using different general-purpose (MS~\cite{Shen2015}, DJF~\cite{Li2016}, and DJFR~\cite{Li2019}) and customized methods for depth maps (TGVL2~\cite{Ferstl2013}) on the Middlebury, NYU, and SUN benchmark datasets.}
	\begin{tabular}{p{4.3cm} p{1.1cm}c p{1.0cm}c p{1.0cm}c}
		\toprule
		Method												& RMSE  			& RMSE 			& RMSE	 \\
		& Middlebury	& NYU			 & SUN \\
		\midrule
		MS~\cite{Shen2015}					& 0.0245 & 0.0167 & 0.0190 \\
		TGVL2~\cite{Ferstl2013}				& 0.0245 & 0.0169 & 0.0213\\
		DJF	\cite{Li2016}					& 0.0266 & - & 0.0218 \\
		DJFR~\cite{Li2019}				 	& 0.0273 & - & 0.0218 \\
		SD filter~\cite{Ham2015} w/ Welsch, w/o \aquasi			& 0.0220 & 0.0143 & 0.0178 \\
		SD filter~\cite{Ham2015} w/o Welsch, w/ \aquasi 							& 0.0253 & 0.0176 & 0.0205\\
		SD filter~\cite{Ham2015} w/ Welsch, w/ QuaSI	& 0.0211 & 0.0136 & 0.0177  \\
		SD filter~\cite{Ham2015} w/ Welsch, w/ \aquasi	& \textbf{0.0197}& \textbf{0.0124}& \textbf{0.0165} \\
		
		\bottomrule
	\end{tabular}
	\label{tab:RMSE}
\end{table}

We study joint upsampling on the Middlebury dataset (using the \textit{Art}, \textit{Books}, \textit{Dolls}, \textit{Laundry}, \textit{Moebius} and \textit{Reindeer} depth maps), the NYU dataset~\cite{Silberman2012}, and the SUN dataset~\cite{Song2015, Janoch2011, Xiao2013}. We generated synthetic depth maps normalized in $[0, 1]$ from their ground truth by Gaussian blurring with standard deviation $\sigma_{\text{blur}} = 4$ followed by nearest neighbor downsampling with a factor of $8$ and by adding Gaussian noise with standard deviation $\sigma_{\text{noise}} = 0.0005$. Joint upsampling is assessed quantitatively with the Root Mean Square Error (RMSE) between the ground truth depth map $\tilde{\vec{f}}$ and a recovered depth map $\vec{f}$. Additionally, we report the Bad Matching Error (BME)~\cite{Scharstein01}
\begin{equation}
	E_{\mathrm{BME}} = \dfrac{1}{N}\sum (\vert f_i - \tilde{f}_i \vert > \delta)\enspace,
\end{equation}
where $N$ are the number of pixels in a depth map and $\delta$ denotes a scalar that can be set accordingly. 

The SD filter with the \aquasi prior according to \eqref{eqn:jointUpsamplingInvProblem} (SD filter, w/ \aquasi) is compared to TGVL2 upsampling~\cite{Ferstl2013}, the mutual structure filter (MS)~\cite{Shen2015}, the deep joint filter (DJF)~\cite{Li2016},  and the residual-based deep joint filter (DJFR)~\cite{Li2019}. We also conducted an ablation study using \eqref{eqn:jointUpsamplingInvProblem} and evaluated different variants of the proposed joint upsampling by omitting individual regularization terms (SD filter w/ Welsch and w/o \aquasi, SD filter w/o Welsch and w/ \aquasi) and by replacing \aquasi with its non-adaptive counterpart (SD filter w/ Welsch and w/ QuaSI). The parameters of \aquasi were set to $p = 0.5$, $n = 9^2$, $\sigma_w = 0.1$, and $\lambda = 0.1$ and the parameters of the competing methods were chosen according to suggestions of the authors. 

\begin{figure}[!tbp]
	\subfloat[Ground truth color image]{\includegraphics[width=0.45\textwidth]{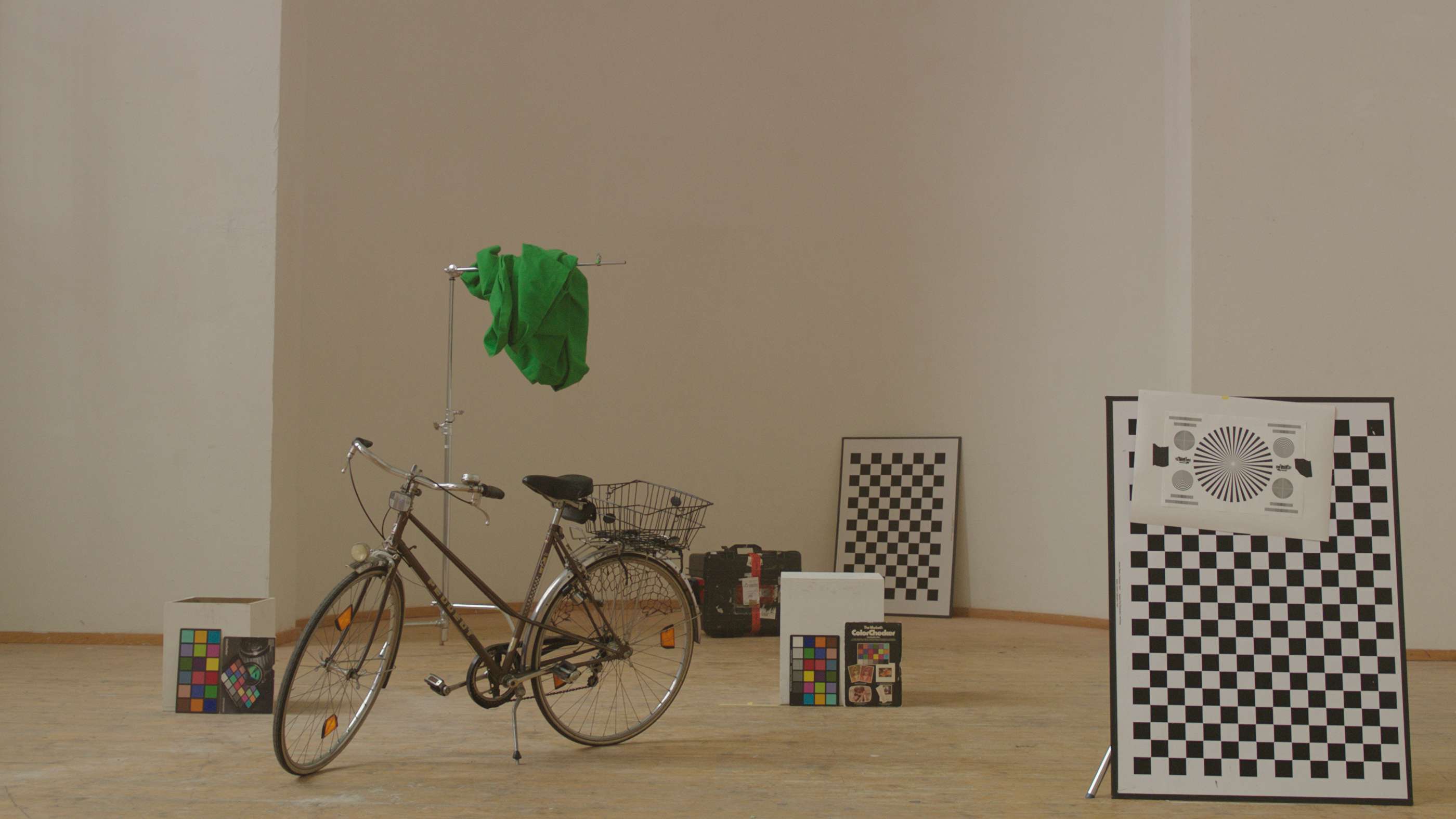}}~
	\subfloat[Near-infrared image]{\includegraphics[width=0.45\textwidth]{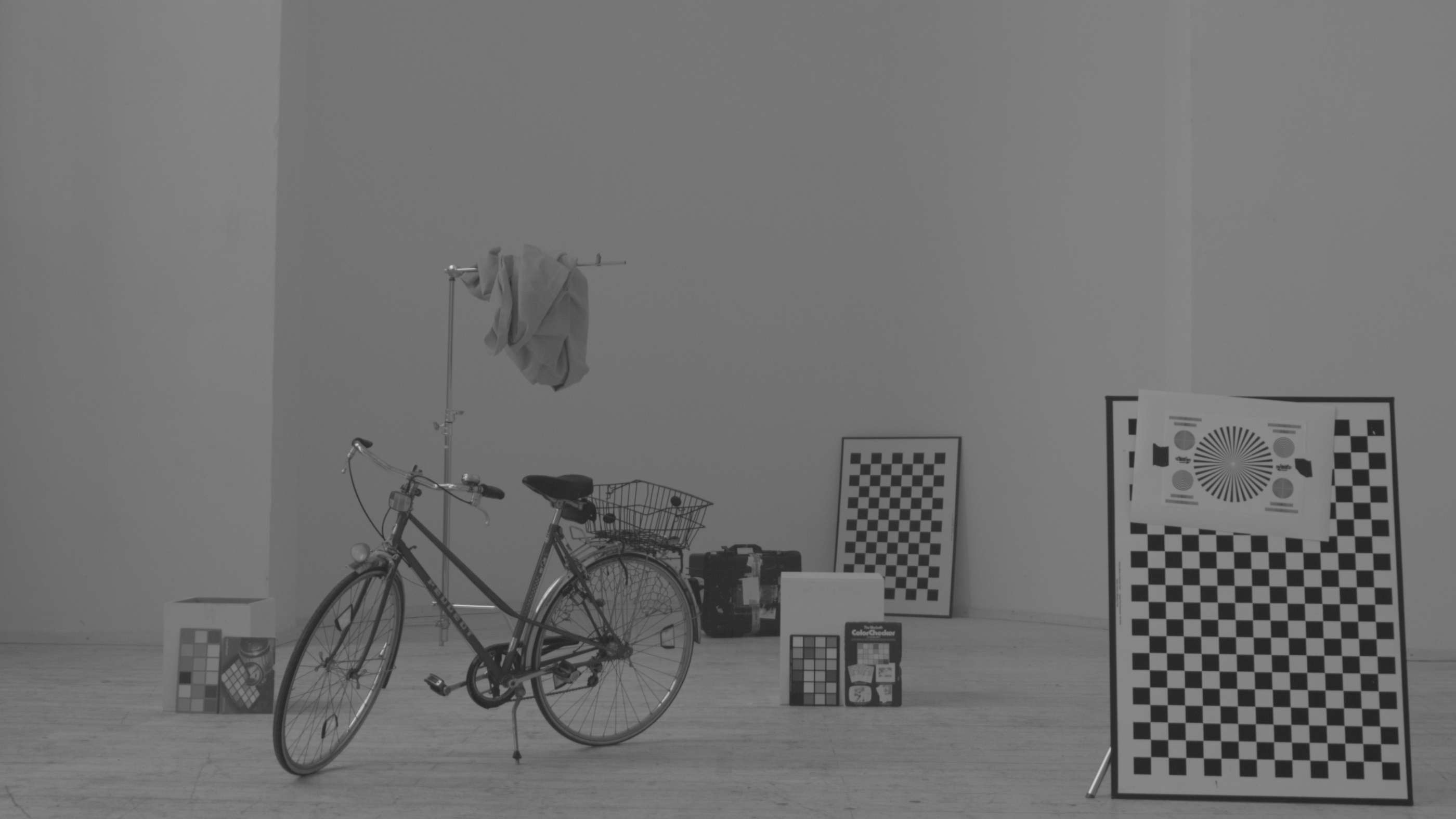}}
	\caption{Example color and near-infrared image from the ARRI dataset~\cite{Luethen2017}.}
	\label{fig:RGBNIR_example}
\end{figure}

\begin{figure}[!tbp]
	\footnotesize
	\centering	
	\subfloat{\includegraphics[width=0.24\linewidth]{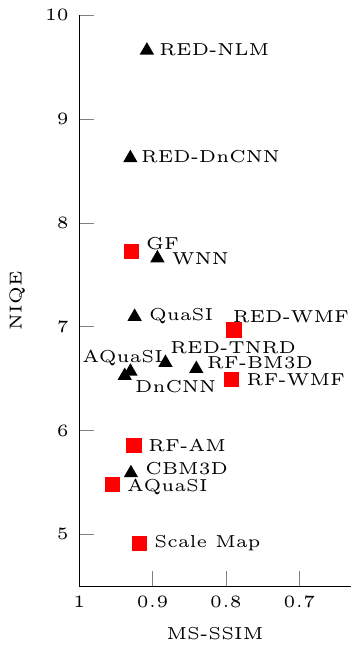}}~	
	\subfloat{\includegraphics[width=0.24\linewidth]{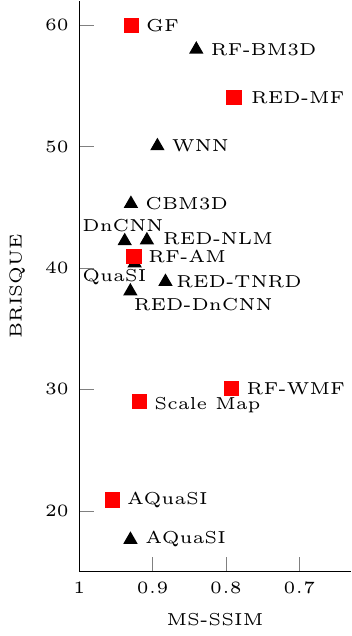}}~
	\subfloat{\includegraphics[width=0.24\linewidth]{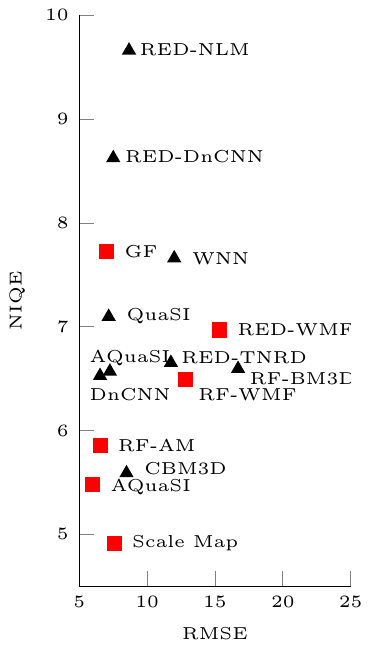}}~	
	\subfloat{\includegraphics[width=0.24\linewidth]{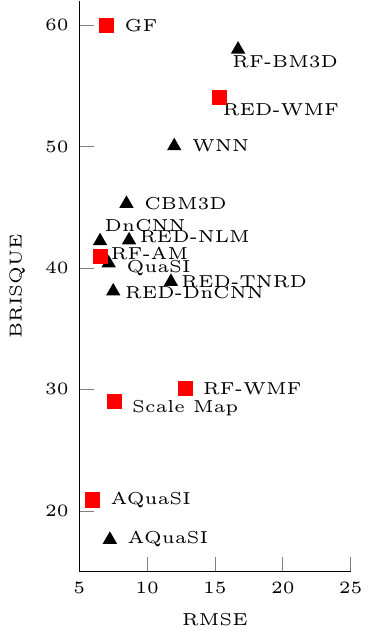}}		
	\caption{Quantitative results of all competing methods on the ARRI dataset using the perception-distortion plane to visualize BRISQUE and NIQE over RMSE and MS-SSIM. The red squares indicate methods with near-infrared guidance and the black triangles indicate methods without near-infrared guidance. All variants of QuaSI and \aquasi use the multi-channel mode.}
	\label{fig:RGBNIRPDPlane}
\end{figure}

Table~\ref{tab:RMSE} shows the average RMSEs on the three datasets achieved by the different methods\footnote{DJF and DJFR are not evaluated on the NYU dataset, since they used the NYU dataset for training.}. Notice that the combination of the SD filter with the \aquasi prior yields the lowest errors in all benchmarks. In comparison to the baseline SD filter (w/ Welsch and w/o \aquasi), SD w/ Welsch, w/ \aquasi reduces the RMSE by $10.5\%$, $13.3\%$, and $7.3\%$ in the Middlebury, NYU, and SUN benchmark, respectively. Figure~\ref{fig:BME} shows the mean BME relative to the ground truth depth maps for different $\delta$ values on the Middlebury data. The \aquasi prior further improves the results of the SD filter and outperforms all competing methods.

Figure~\ref{fig:jointUpsampling} shows qualitative results of the competing upsampling approaches. The MS and DJFR filters were sensitive to noise. In contrast, TGVL2 erroneously transfers textures from the color image to the depth map. An example can be seen in Fig.~\ref{fig:jointUpsamplingTGVL2}, where the layers of the books cause artifacts in the depth map. The SD filter w/ Welsch and w/ \aquasi achieves the best upsampling performance, which is consistent with the quantitative results. The SD filter w/ Welsch and w/ QuaSI leads to a good noise reduction but causes to oversmoothing as shown in Fig.~\ref{fig:jointUpsamplingQuaSI}. While the SD filter w/o Welsch and w/ \aquasi does not remove noise as effectively as SD filter w/ Welsch and w/o \aquasi, edge preservation is greatly improved. This observation underlines the reported structure preservation property of the \aquasi prior.

\subsection{RGB/NIR Cross Field Image Restoration}
\label{sec:RGBAndNIRImageRestoration}

Another popular application is color (RGB) and near-infrared (NIR) image restoration. The scene is captured using a RGB and a NIR sensor. While NIR images in such setups are oftentimes flashed and feature high signal-to-noise ratios, the RGB data can be noisy. Here, we aim at denoising RGB images. \aquasi is plugged into classical TV denoising~\cite{Goldstein2009}. That is, given a noisy RGB image $\vec{g}$, we recover its clean counterpart $\vec{f}$ according to
\begin{equation}
	\label{eqn:crossFieldInvProblem}
	\begin{split}
	\hat{\vec{f}} &= \argmin_{\vec{f}} 
	\mathcal{L}(\vec{f}, \vec{g}) 
	+ \underbrace{\mu \left( \|\nabla_x \vec{f}\|_1 + \|\nabla_y \vec{f}\|_1 \right)}_{\text{TV prior}} \\
	&\qquad\qquad\qquad\qquad + \underbrace{\lambda \|\vec{f} - Q_{p,\vec{w}}(\vec{f})\|_1}_{\text{\aquasi prior}}
	\enspace,
\end{split}
\end{equation}
where $\nabla_x \vec{f}$ and $\nabla_y \vec{f}$ denote the derivatives of $\vec{f}$ in $x$- and $y$-direction to define the TV prior with regularization weight $\mu \geq 0$ and $\mathcal{L}(\vec{f}, \vec{g}) = \|\vec{f} - \vec{g}\|_2^2$ models the data fidelity. The optimization of \eqref{eqn:crossFieldInvProblem} is performed via ADMM (see Sec.~\ref{sec:AlternatingDirectionMethodOfMultipliersADMM}). We apply \aquasi with $p = 0.5$, $n = 9^2$, and $\sigma_w = 0.006$ and the TV prior with $\mu = 0.05$ and $\beta = 7$ for the corresponding auxiliary variable. We compare the channel-wise approach of \aquasi with $\lambda = 13$ and $\alpha = 1100$ to the multi-channel (MC) mode to simultaneously denoise three color channels with $\lambda = 13 \cdot 3$, $\alpha = 1100 \cdot 3$.

\begin{figure}[!tbp]
	\footnotesize
	\setlength{\tabcolsep}{1.5pt}
	\begin{tabular}{>{\centering\arraybackslash}p{0.24\linewidth} p{0.24\linewidth} p{0.24\linewidth}}
		(a) Near-infrared image & 
		\includegraphics[width=\linewidth,align=c]{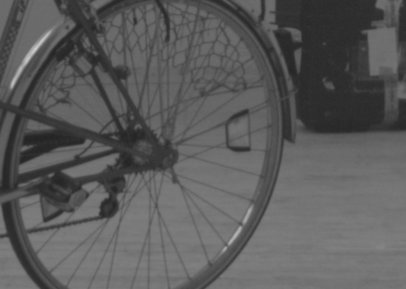} &
		\includegraphics[width=\linewidth,align=c]{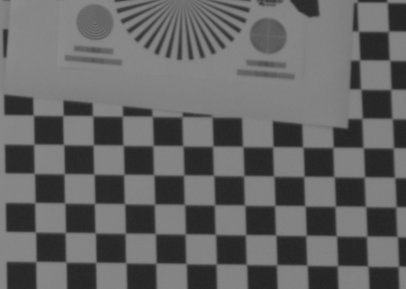} \vspace*{0.01mm}  \\	
		(b) Guided filter~\cite{He2013}& 
		\includegraphics[width=\linewidth,align=c]{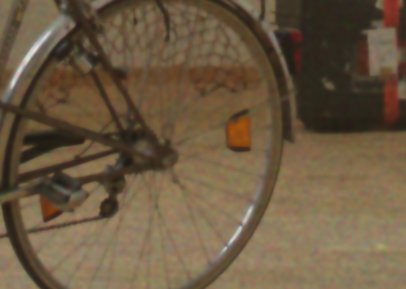} &
		\includegraphics[width=\linewidth,align=c]{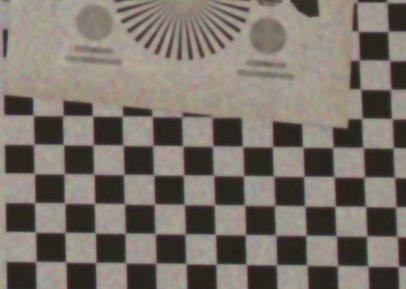} \vspace*{0.01mm}  \\
		(c) Scale map~\cite{Yan2013}& 
		\includegraphics[width=\linewidth,align=c]{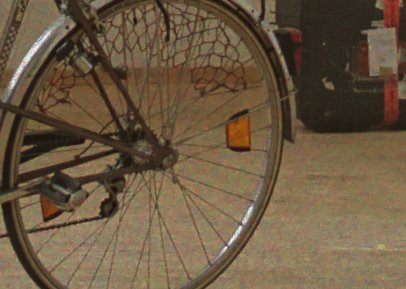} &
		\includegraphics[width=\linewidth,align=c]{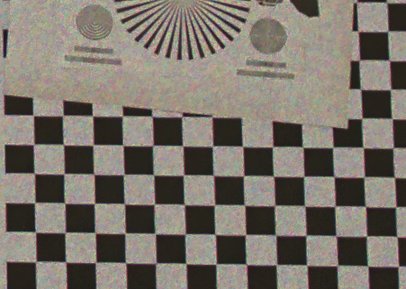} \vspace*{0.01mm}  \\
		(d) RED~\cite{Romano2016a} with WMF~\cite{Zhang2014} & 
		\includegraphics[width=\linewidth,align=c]{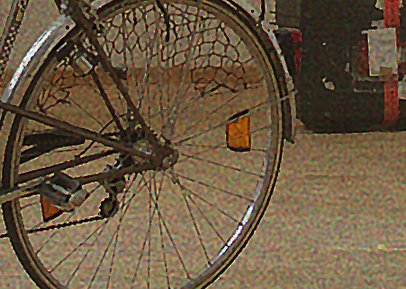} &
		\includegraphics[width=\linewidth,align=c]{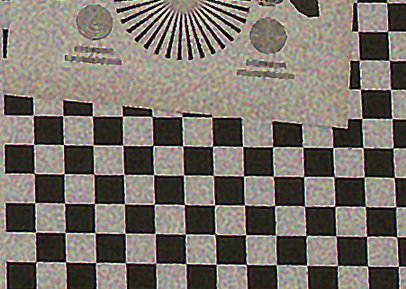} \vspace*{0.01mm}  \\	
		(e) RF~\cite{Tao2017} with AM~\cite{Gastal2012} & 
		\includegraphics[width=\linewidth,align=c]{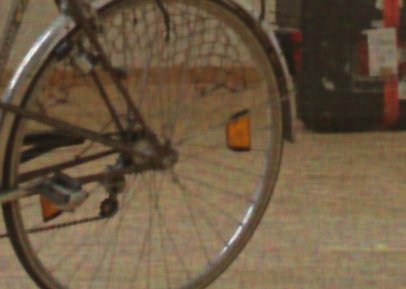} &
		\includegraphics[width=\linewidth,align=c]{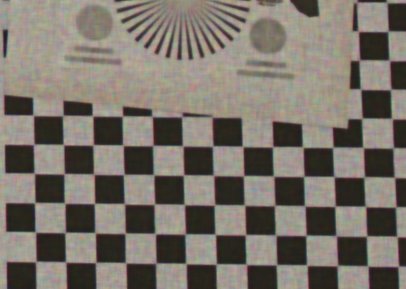} \vspace*{0.01mm} \\
		(f) RF~\cite{Tao2017} with WMF~\cite{Zhang2014}  & 
		\includegraphics[width=\linewidth,align=c]{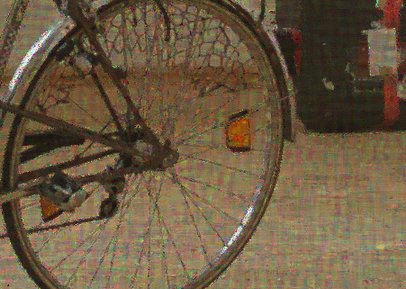} &
		\includegraphics[width=\linewidth,align=c]{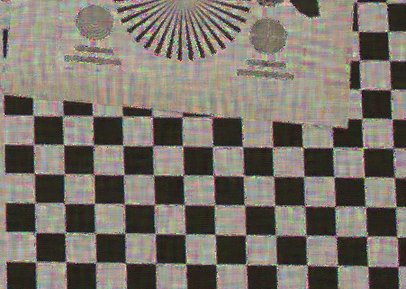} \vspace*{0.01mm} \\ 
		(g) \aquasi\ - MC (\textit{static} guidance)& 
		\includegraphics[width=\linewidth,align=c]{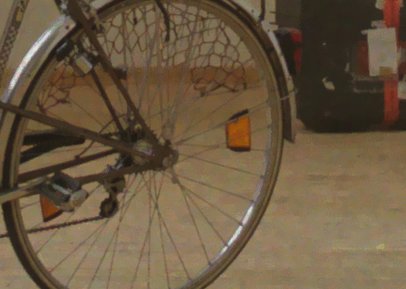} &
		\includegraphics[width=\linewidth,align=c]{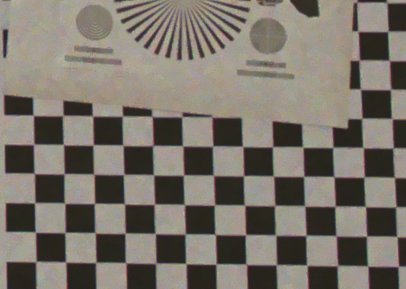} \vspace*{0.01mm} \\
	\end{tabular}
	\caption{RGB/NIR image restoration on the ARRI dataset~\cite{Luethen2017}. All competing methods enable guided filtering to denoise the degraded color image using the near-infrared image for guidance.}
	\label{tab:RGBNIR_guidance}
\end{figure}

\subsubsection{With Near-Infrared Guidance}
\label{sec:WithNIRGuidance}

First, we study the case, where NIR images are used to restore clean RGB images using the \textit{static} guidance mode of our prior. The evaluation is performed on eleven registered ground truth RGB/NIR image pairs taken from the publicly available ARRI dataset~\cite{Luethen2017} as shown in Fig.~\ref{fig:RGBNIR_example}. We simulate noisy color images by adding speckle noise (variance $\sigma_{\text{noise}} = 0.2$) to the respective ground truth images. We compare \aquasi against two state-of-the-art static guidance based methods, namely guided filtering~\cite{He2013} and cross-field restoration via scale map~\cite{Yan2013}. In addition, we adopt regularization by denoising (RED)~\cite{Romano2016a} and zero-order reverse filtering (RF)~\cite{Tao2017}. Specifically, we equipped RED with the weighted median filter~\cite{Zhang2014} (RED with WMF) and the RF framework with the weighted median filter (RF with WMF) as well as adaptive manifold filtering~\cite{Gastal2012} (RF with AM).

For a quantitative comparison, the methods are plotted on a
perception-distortion plane using the evaluation framework by Blau and
Michaeli~\cite{Blau2018}. The distortion is quantified by the
RMSE and multiscale structural similarity (MS-SSIM)~\cite{Wang2003}. The
perceptual quality is quantified by the natural image quality evaluator
(NIQE)~\cite{Mittal2013} and the blind/referenceless image spatial quality
evaluator (BRISQUE)~\cite{Mittal2012}. These metrics are arranged in the
perception-distortion plane in a way that better methods are closer to the
bottom left of the plane. For BRISQUE and NIQE, low values imply high image quality in terms of observed naturalness. The closer MS-SSIM is to the value 1 the higher is the structural similarity between the restored image and the ground truth. 

Figure~\ref{fig:RGBNIRPDPlane} shows the averaged perception-distortion planes. The distortion measure MS-SSIM
is used in the first two plots, where \aquasi performs best. For the associated
perceptual metrics NIQE and BRISQUE, the best performer is in one case Scale
Map and in the other case \aquasi. The third and fourth plots evaluate RMSE as a distortion measure and draw a similar picture.
Again, \aquasi performs best \wrt RMSE, again only challenged by
scale map for the perceptual metric NIQE. Under BRISQUE, \aquasi performs best.
Overall, \aquasi shows a highly competitive performance, and is in all cases
closely located to the bottom left corner.

Figure \ref{tab:RGBNIR_guidance} depicts qualitative results for several methods. The guided filter slightly oversmooths the image. For example, the contours on the
bike are barley visible. Conversely, scale map produces a crisp
output at the expense of noise and color artifacts. Especially homogeneous areas, such as the
checkerboard fields, remain noisy. RED with WMF removes noisy color pixels but cause blob-like artifacts. RF with AM produces color block artifacts and blurs small structures. In contrast, RF with WMF causes grid-like artifacts. Overall, \aquasi with \textit{static} guidance yields high perceptual image quality conforming the perception-distortion tradeoff analysis.

\begin{figure}[!t]
	\footnotesize
	\setlength{\tabcolsep}{1.5pt}
	\begin{tabular}{>{\centering\arraybackslash}p{0.24\linewidth} p{0.24\linewidth} p{0.24\linewidth}}
		(a) Ground truth & 
		\includegraphics[width=\linewidth,align=c]{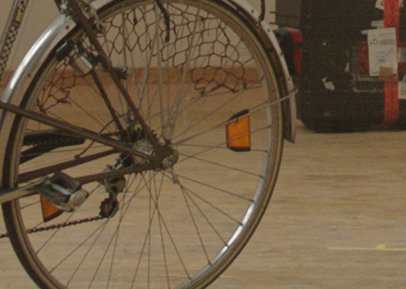}&
		\includegraphics[width=\linewidth,align=c]{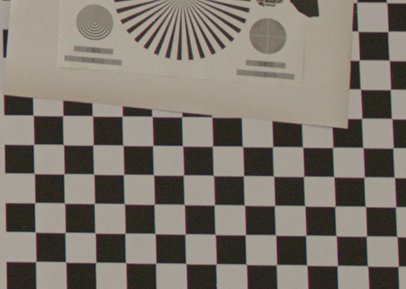} \vspace*{0.01mm}  \\
		(b) Noisy RGB image & 
		\includegraphics[width=\linewidth,align=c]{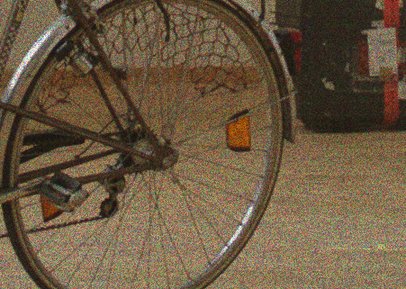}&
		\includegraphics[width=\linewidth,align=c]{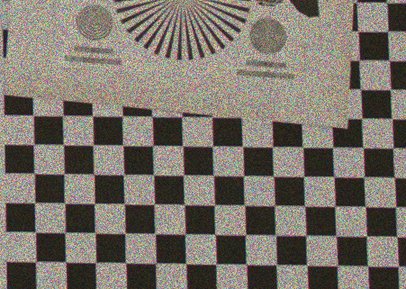} \vspace*{0.01mm}  \\
		(c) WNN~\cite{Gu2014} & 
		\includegraphics[width=\linewidth,align=c]{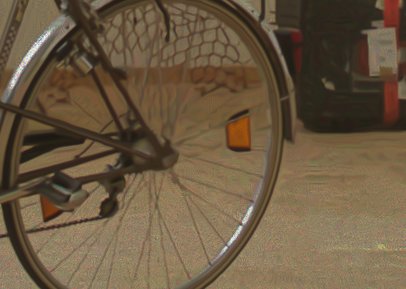}&
		\includegraphics[width=\linewidth,align=c]{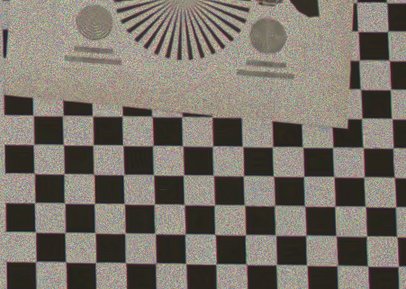} \vspace*{0.01mm}  \\
		(d) CBM3D~\cite{Dabov2007a} & 
		\includegraphics[width=\linewidth,align=c]{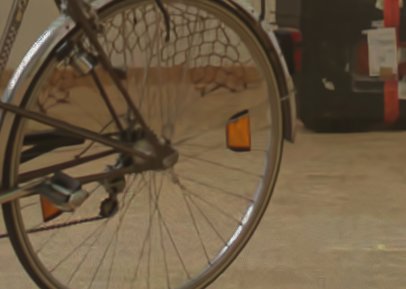}&
		\includegraphics[width=\linewidth,align=c]{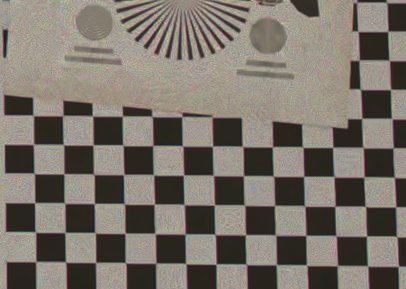} \vspace*{0.01mm}  \\
		(e) DnCNN \cite{Zhang2017beyond}& 
		\includegraphics[width=\linewidth,align=c]{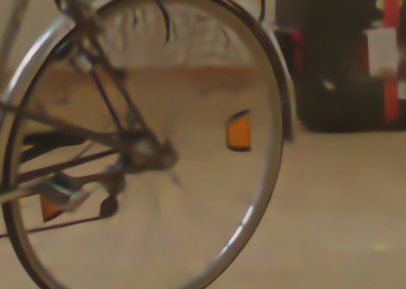}&
		\includegraphics[width=\linewidth,align=c]{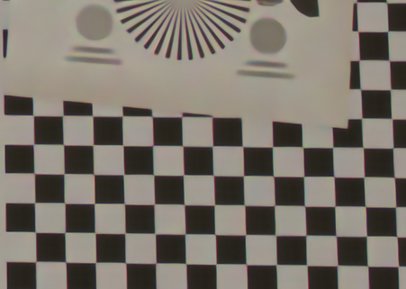} \vspace*{0.01mm} \\
		(f) QuaSI - MC~\cite{Schirrmacher2018} & 
		\includegraphics[width=\linewidth,align=c]{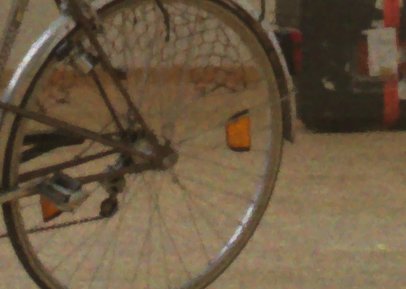}&
		\includegraphics[width=\linewidth,align=c]{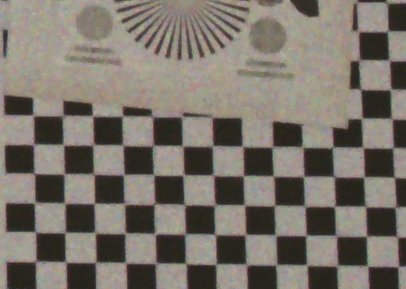} \vspace*{0.01mm}  \\
		(g) \aquasi\ - MC (\textit{dynamic} guidance) & 
		\includegraphics[width=\linewidth,align=c]{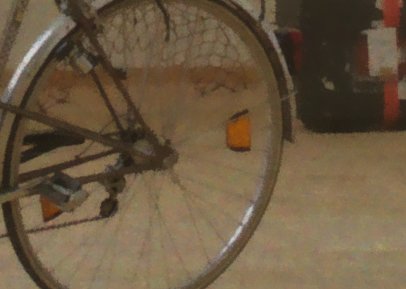}&
		\includegraphics[width=\linewidth,align=c]{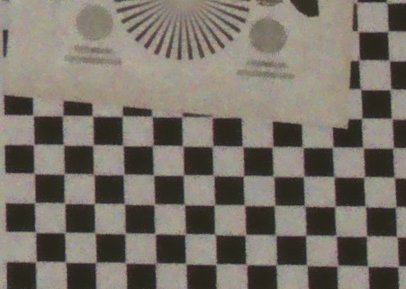} \vspace*{0.01mm} \\
	\end{tabular}
	\caption{RGB/NIR image restoration on the ARRI dataset~\cite{Luethen2017}. All competing methods perform denoising on the degraded color image only.}
	\label{tab:RGBNIR_Noguidance}
\end{figure}

\subsubsection{Without Near-Infrared Guidance}
\label{sec:WithoutNearInfraredGuidance}

We also examine the more challenging problem of restoring a clean RGB image without guidance by NIR data. For that use case, we compare \aquasi in its \textit{dynamic} guidance mode to different well known stand-alone image denoising methods, namely weighted nuclear norm minimization (WNN)~\cite{Gu2014}, color block-matching and 3D filtering (CBM3D)~\cite{Dabov2007a}, and feed-forward denoising convolutional neural networks (DnCNN) \cite{Zhang2017beyond}. We also adopted the RF approach in combination with BM3D~\cite{Dabov2007} (RF with BM3D) as well as RED~\cite{Romano2016a} with non-local means (NLM)~\cite{Buades2005}, BM3D~\cite{Dabov2007}, DnCNN~\cite{Zhang2017beyond}, and trainable nonlinear reaction-diffusion (TNRD)~\cite{Chen2017}.

In Fig.~\ref{fig:RGBNIRPDPlane} the quantitative results are shown. In terms of distortion, \aquasi with \textit{dynamic} guidance and CBM3D yield similar results. In terms of perceptual quality, \aquasi with \textit{dynamic} guidance has the best BRISQUE value whereas CBM3D performs better \wrt NIQE.

Figure~\ref{tab:RGBNIR_Noguidance} shows the qualitative results of all competing methods that do not use additional guidance data for denoising. WNN produces a sharp output at the expense of colored noise, especially in homogeneous areas. Compared to WNN, CBM3D suppresses more noise, but produces small streaking artifacts in homogeneous areas. While CBM3D does contain more residual noise than DnCNN, small details such as the spokes of the wheel are preserved.  QuaSI - MC provides competitive denoising performance \wrt distortion, but produces slightly oversmoothed results. \aquasi with \textit{dynamic} guidance has a similar performance to QuaSI, but avoids color artifacts. To enable a fair comparison, AQuaSI's smoothing parameter $\sigma_w$ was optimized for \textit{static} guidance mode. Hence, the performance of the \textit{dynamic} mode can be further improved by tuning $\sigma_w$ either towards low distortions or good perceptual image quality in the perception-distortion plane. 

\subsubsection{Comparison of Different \aquasi Variants}
\label{sec:ComparisonOfDifferentAquasiVariants}

In Table \ref{tab:RGBNIR_AQUASI}, the average RMSE and MS-SSIM values of all four variants of \aquasi on the ARRI dataset are shown. 
In terms of distortion measures, the multi-channel approach slightly outperforms channel-wise processings in both guidance modes. We explain this behavior by the avoidance of color artifacts under a simultaneous processing of the RGB channels. As shown in Sec.~\ref{sec:MultiChannel}, another major advantage of the multi-channel over the channel-wise approach is the lower computation time. Thus, the multi-channel approach is preferable to simple channel-wise processing. Moreover, \aquasi with \textit{dynamic} guidance is outperformed by the \textit{static} guidance mode. Notice that the former achieves a better recovery of small structures, as shown in Fig.~\ref{tab:RGBNIR_guidance} (g). This reveals the benefit of exploiting NIR data for regularizing RGB denoising.

\begin{table}[!tb]
	\centering
	\caption{Averaged RMSE and MS-SSIM for RGB/NIR image restoration using all \aquasi variants on the ARRI dataset.}
	\begin{tabular}{p{6.0cm}cc}
	\toprule
	Method							& RMSE    & MS-SSIM	 \\	
	\midrule
	\aquasi\ - channel-wise, \textit{dynamic} guidance	& 14.89	 & 0.84 \\
	\aquasi\ - channel-wise, \textit{static} guidance	& 6.62	  & \textbf{0.95}\\
	\aquasi\ - multi-channel, \textit{dynamic} guidance	& 7.26	  & 0.93 \\
	\aquasi\ - multi-channel, \textit{static} guidance	& \textbf{5.99}	 & \textbf{0.95} \\
	\bottomrule
\end{tabular}
\label{tab:RGBNIR_AQUASI}
\end{table}

\section{Conclusions}
\label{sec:Conclusion}

We propose \aquasi, a novel prior for solving inverse problems. The key assumption of
this prior is that good natural images are fixed points under a quantile filter 
with a spatially adaptive weighting scheme. The
flexibility of \aquasi can be attributed to the optional use of a guidance
image, which can be either used in the direction of joint filtering based on a
\textit{static} guidance, or in \textit{dynamic} guidance, in which case it is similar in spirit
to filtering.
We present \aquasi within a full mathematical
framework to seamlessly integrate it into
popular optimization schemes such as gradient descent and ADMM. 

In our experimental results, we demonstrate the strength of \aquasi for the
application of RGB/depth map upsampling, RGB/NIR image restoration, and image deblurring. 
Our prior is universally applicable and compares favorably against the state-of-the-art in the 
investigated applications. Its importance over related RED or PnP approaches is especially evident for inverse problems with unknown or non-Gaussian noise, \eg, RGB/NIR restoration. In analyzing perception-distortion planes, we found that \aquasi shows decent tradeoffs between low signal distortions and high perceptional quality. In the simpler cases of noise-free problems or degradations by moderate levels of Gaussian noise, RED algorithms perform equally well as shown for non-blind deblurring. We also found that for certain problems like RGB/depth map upsampling, task-specific priors can outperform the solely use of \aquasi due to their superior denoising capabilities. However, thanks to its structure preservation, state-of-the-art performance is achieved when combining \aquasi with task-specific regularization.

Due to the ever-increasing number of learning-based image restoration techniques, we believe that transferring our universal prior to the field of deep learning, \eg, as operator in neural networks, is an interesting area for future research and can help to boost the performance of these methods.    

\section*{Acknowledgment}
This work was supported in part by the Deutsche Forschungsgemeinschaft (DFG, German Research Foundation) – Project number 146371743 – TRR 89 Invasive Computing

\bibliographystyle{IEEEtran}

\clearpage

\begin{appendices}
	
\section{Supplemental Material }
	
%
%
%

	\begin{figure}[!bp]
		\footnotesize
		\setlength{\tabcolsep}{1.5pt}
		\begin{tabular}{>{\centering\arraybackslash}p{0.24\linewidth} p{0.22\linewidth} p{0.22\linewidth}}
			(a) Near-infrared image & 
			\includegraphics[width=\linewidth,align=c]{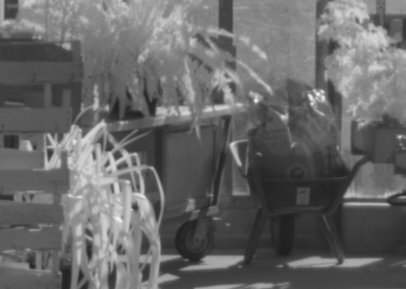} &
			\includegraphics[width=\linewidth,align=c]{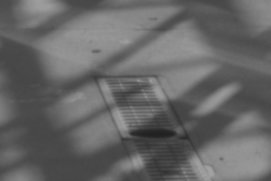} \vspace*{0.01mm}  \\	
			(b) Guided filter \cite{He2013} & 
			\includegraphics[width=\linewidth,align=c]{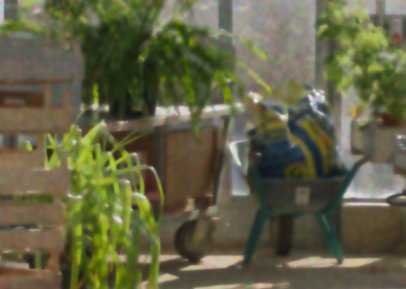} &
			\includegraphics[width=\linewidth,align=c]{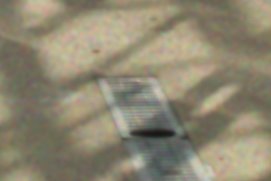} \vspace*{0.01mm}  \\
			(c) Scale map \cite{Yan2013} & 
			\includegraphics[width=\linewidth,align=c]{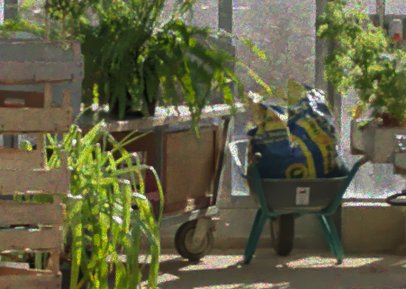} &
			\includegraphics[width=\linewidth,align=c]{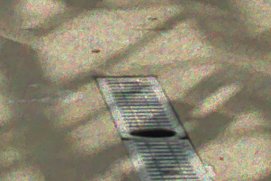} \vspace*{0.01mm}  \\
			(d) RED \cite{Romano2016a} with WMF \cite{Zhang2014} & 
			\includegraphics[width=\linewidth,align=c]{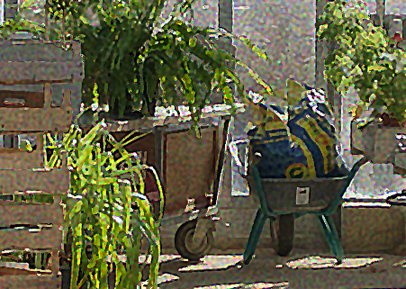} &
			\includegraphics[width=\linewidth,align=c]{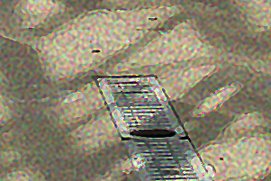} \vspace*{0.01mm}  \\	
			(e) RF \cite{Tao2017} with AM \cite{Gastal2012} & 
			\includegraphics[width=\linewidth,align=c]{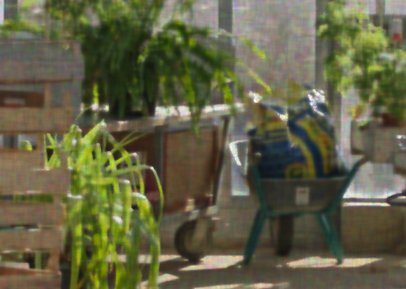} &
			\includegraphics[width=\linewidth,align=c]{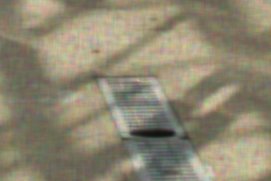} \vspace*{0.01mm} \\
			(f) RF \cite{Tao2017} with WMF \cite{Zhang2014} & 
			\includegraphics[width=\linewidth,align=c]{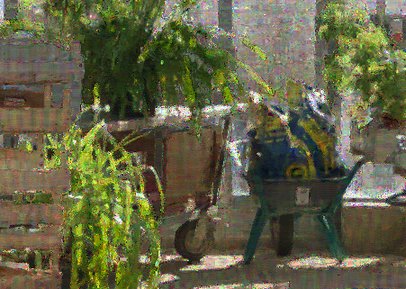} &
			\includegraphics[width=\linewidth,align=c]{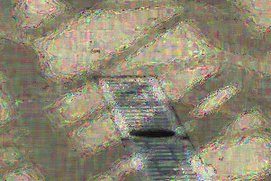} \vspace*{0.01mm} \\ 
			(g) \aquasi\ - MC (static guidance) & 
			\includegraphics[width=\linewidth,align=c]{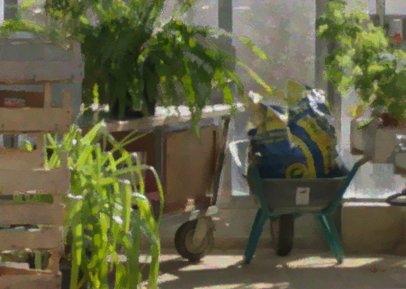} &
			\includegraphics[width=\linewidth,align=c]{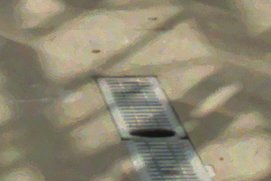} \vspace*{0.01mm} \\
		\end{tabular}
		\caption{RGB/NIR image restoration on the ARRI dataset \cite{Luethen2017}. All competing methods enable guided filtering to denoise the degraded color image using the near-infrared image for guidance.}
		
	\end{figure}

	\begin{figure}[!tbp]
		\footnotesize
		\setlength{\tabcolsep}{1.5pt}
		\begin{tabular}{>{\centering\arraybackslash}p{0.24\linewidth} p{0.22\linewidth} p{0.22\linewidth}}
			(a) Ground truth & 
			\includegraphics[width=\linewidth,align=c]{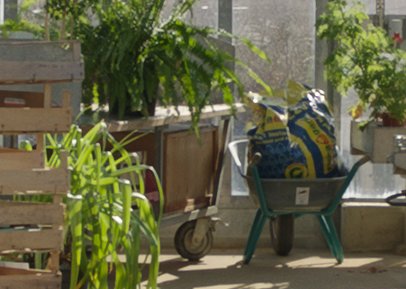}&
			\includegraphics[width=\linewidth,align=c]{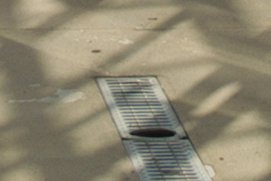} \vspace*{0.01mm}  \\
			(b) Noisy RGB image & 
			\includegraphics[width=\linewidth,align=c]{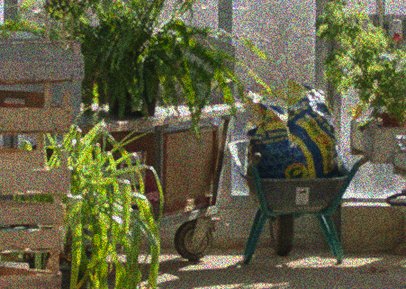}&
			\includegraphics[width=\linewidth,align=c]{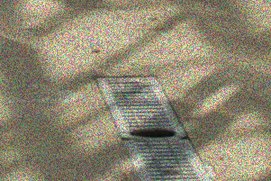} \vspace*{0.01mm}  \\
			(c) WNN \cite{Gu2014} & 
			\includegraphics[width=\linewidth,align=c]{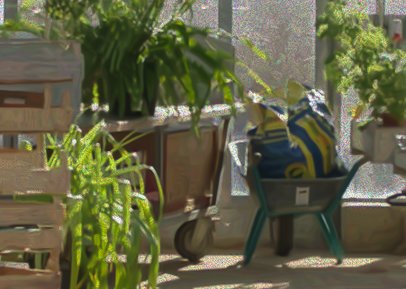}&
			\includegraphics[width=\linewidth,align=c]{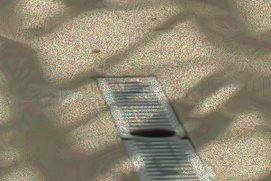} \vspace*{0.01mm}  \\
			(d) CBM3D \cite{Dabov2007a} & 
			\includegraphics[width=\linewidth,align=c]{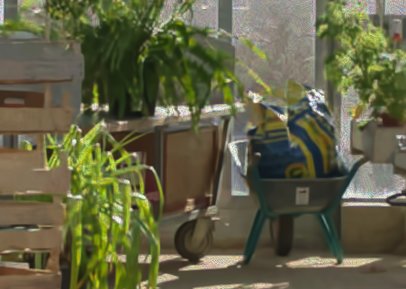}&
			\includegraphics[width=\linewidth,align=c]{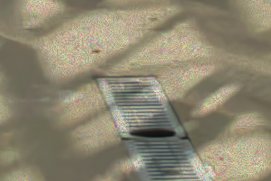} \vspace*{0.01mm}  \\
			(e) RF \cite{Tao2017} with BM3D \cite{Dabov2007} & 
			\includegraphics[width=\linewidth,align=c]{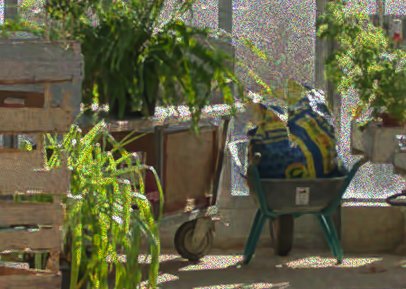}&
			\includegraphics[width=\linewidth,align=c]{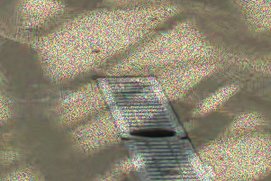} \vspace*{0.01mm} \\
			(d)  DnCNN  ~\cite{Zhang2017beyond} & 
			\includegraphics[width=\linewidth,align=c]{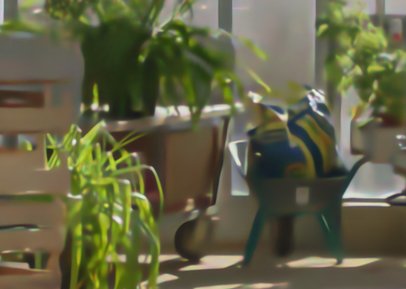}&
			\includegraphics[width=\linewidth,align=c]{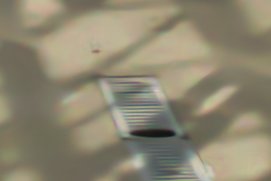} \vspace*{0.01mm}  \\
			(f) QuaSI - MC \cite{Schirrmacher2018} & 
			\includegraphics[width=\linewidth,align=c]{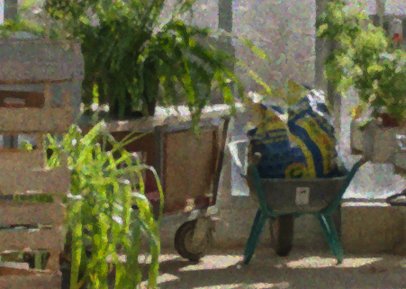}&
			\includegraphics[width=\linewidth,align=c]{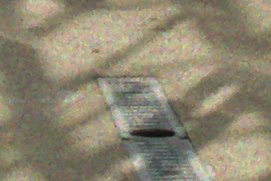} \vspace*{0.01mm}  \\
			(g) \aquasi\ - MC (dynamic guidance) & 
			\includegraphics[width=\linewidth,align=c]{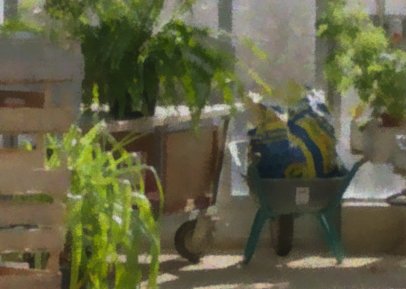}&
			\includegraphics[width=\linewidth,align=c]{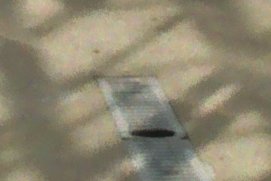} \vspace*{0.01mm} \\
		\end{tabular}
		\caption{RGB/NIR image restoration on the ARRI dataset \cite{Luethen2017}. All competing methods perform denoising on the degraded color image only.}
		
	\end{figure}

	\begin{figure}[!tbp]
		\footnotesize
		\setlength{\tabcolsep}{1.5pt}
		\begin{tabular}{>{\centering\arraybackslash}p{0.24\linewidth} p{0.22\linewidth} p{0.22\linewidth}}
			
			(a) Near-infrared image & 
			\includegraphics[width=\linewidth,align=c]{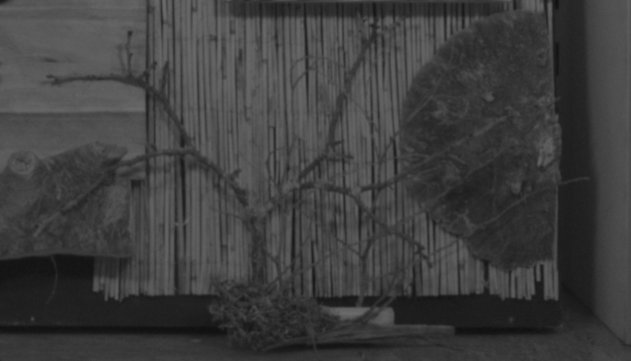} &
			\includegraphics[width=\linewidth,align=c]{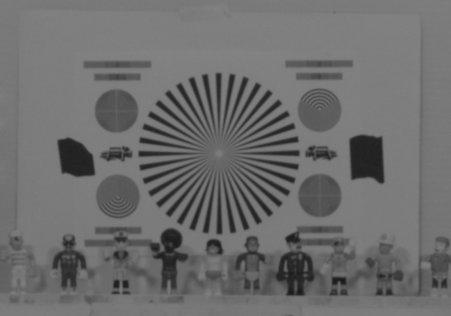} \vspace*{0.01mm}  \\	
			(b) Guided filter \cite{He2013} & 
			\includegraphics[width=\linewidth,align=c]{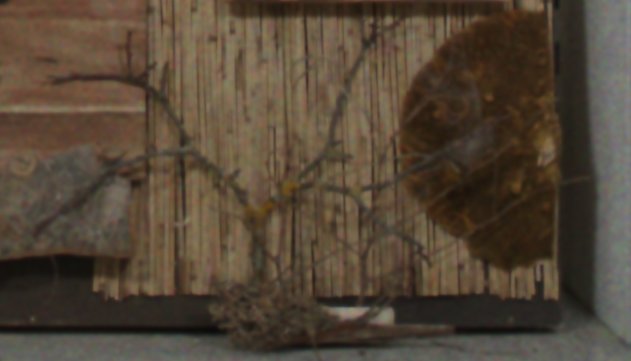} &
			\includegraphics[width=\linewidth,align=c]{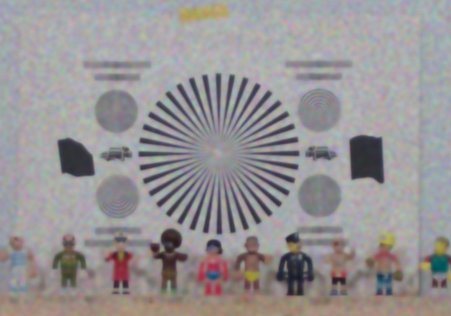} \vspace*{0.01mm}  \\
			(c) Scale map \cite{Yan2013} & 
			\includegraphics[width=\linewidth,align=c]{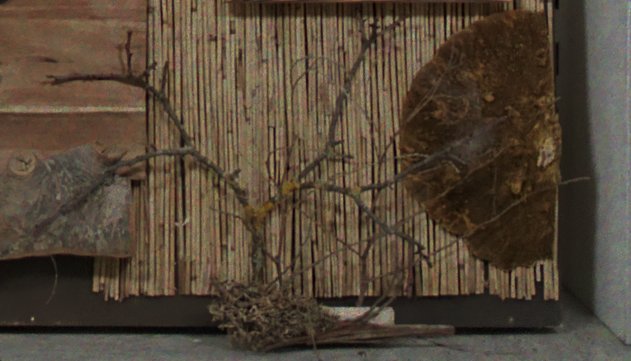} &
			\includegraphics[width=\linewidth,align=c]{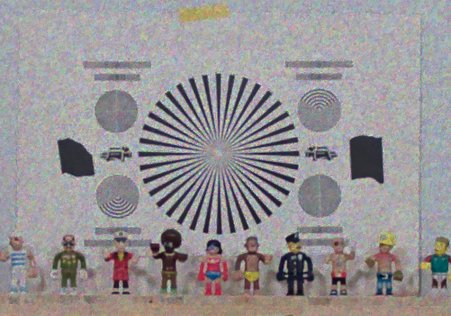} \vspace*{0.01mm}  \\
			(d) RED \cite{Romano2016a} with WMF \cite{Zhang2014} & 
			\includegraphics[width=\linewidth,align=c]{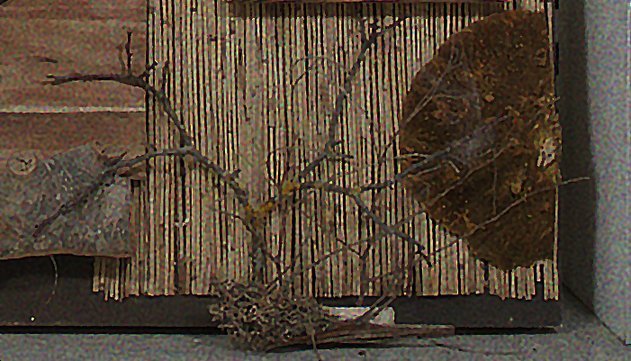} &
			\includegraphics[width=\linewidth,align=c]{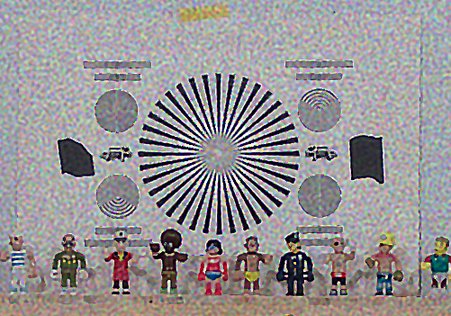} \vspace*{0.01mm}  \\	
			(e) RF \cite{Tao2017} with AM \cite{Gastal2012} & 
			\includegraphics[width=\linewidth,align=c]{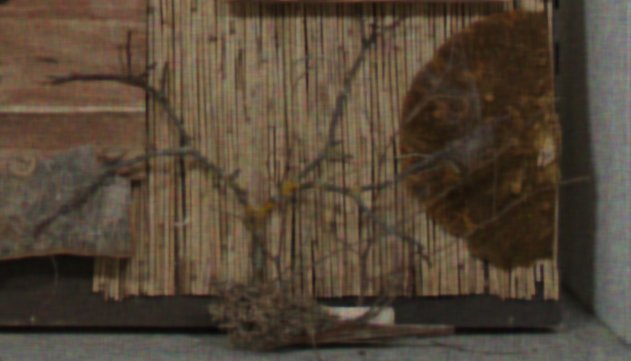} &
			\includegraphics[width=\linewidth,align=c]{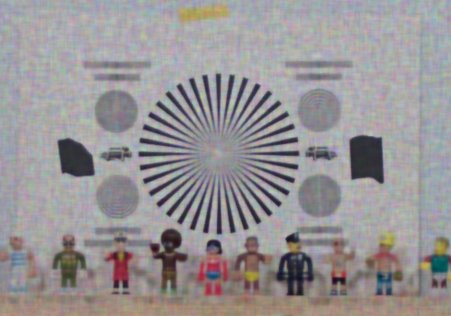} \vspace*{0.01mm} \\
			(f) RF \cite{Tao2017} with WMF \cite{Zhang2014} & 
			\includegraphics[width=\linewidth,align=c]{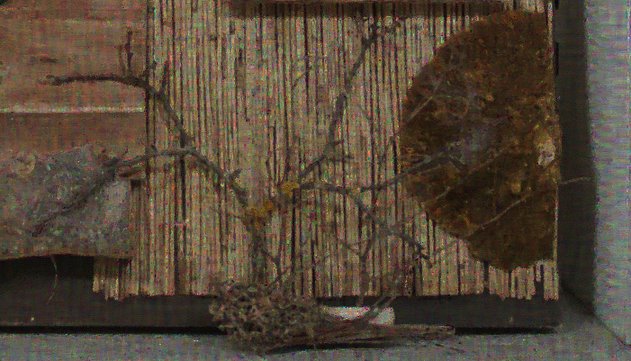} &
			\includegraphics[width=\linewidth,align=c]{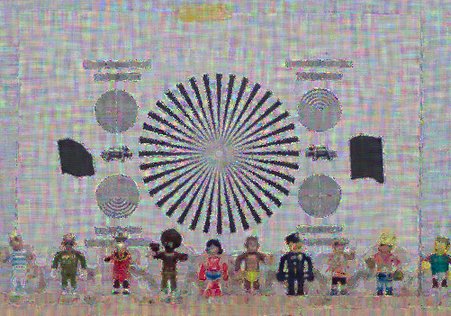} \vspace*{0.01mm} \\ 
			(g) \aquasi\ - MC (static guidance) & 
			\includegraphics[width=\linewidth,align=c]{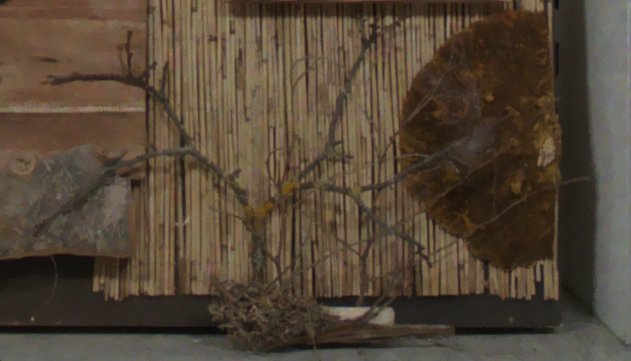} &
			\includegraphics[width=\linewidth,align=c]{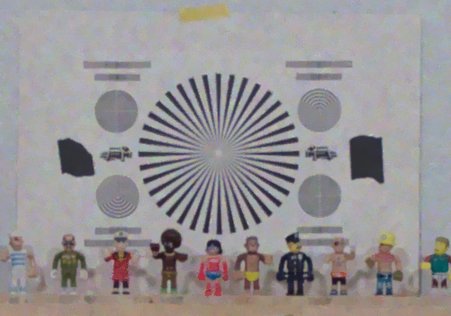} \vspace*{0.01mm} \\
		\end{tabular}
		\caption{RGB/NIR image restoration on image 11 of the ARRI dataset \cite{Luethen2017}. All competing methods enable guided filtering to denoise the degraded color image using the near-infrared image for guidance.}
		
	\end{figure}

	\begin{figure}[!tbp]
		\footnotesize
		\setlength{\tabcolsep}{1.5pt}
		\begin{tabular}{>{\centering\arraybackslash}p{0.24\linewidth} p{0.22\linewidth} p{0.22\linewidth}}
			(a) Ground truth & 
			\includegraphics[width=\linewidth,align=c]{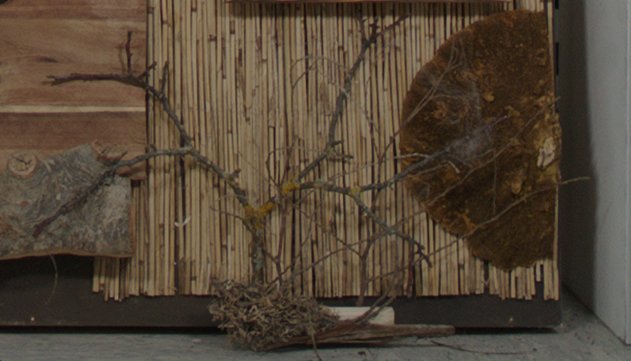}&
			\includegraphics[width=\linewidth,align=c]{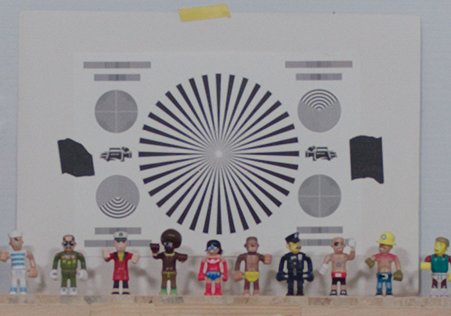} \vspace*{0.01mm}  \\
			(b) Noisy RGB image & 
			\includegraphics[width=\linewidth,align=c]{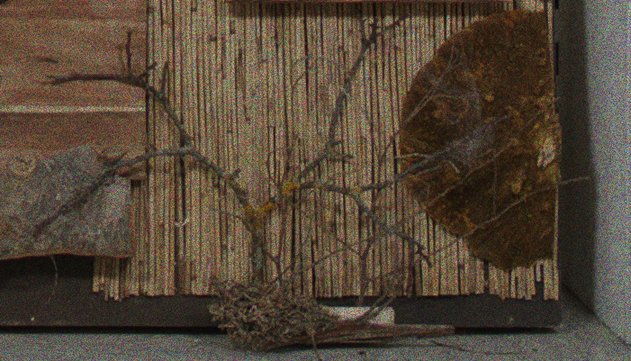}&
			\includegraphics[width=\linewidth,align=c]{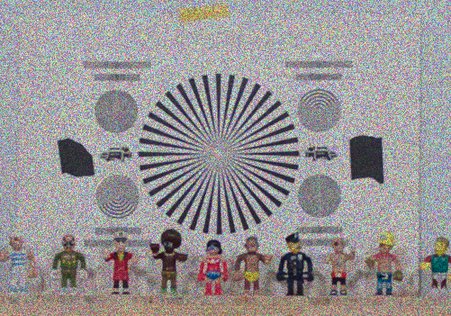} \vspace*{0.01mm}  \\
			(c) WNN \cite{Gu2014} & 
			\includegraphics[width=\linewidth,align=c]{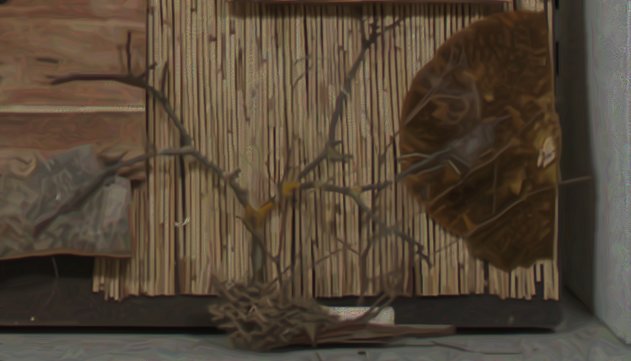}&
			\includegraphics[width=\linewidth,align=c]{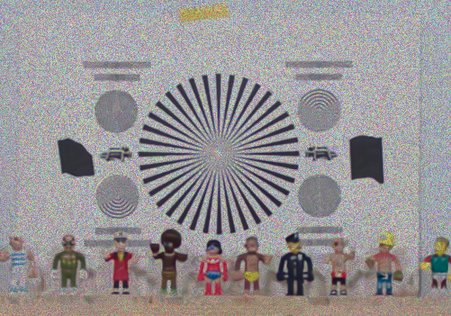} \vspace*{0.01mm}  \\
			(d) CBM3D \cite{Dabov2007a} & 
			\includegraphics[width=\linewidth,align=c]{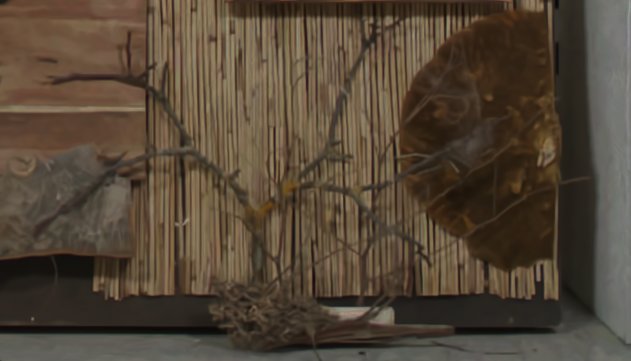}&
			\includegraphics[width=\linewidth,align=c]{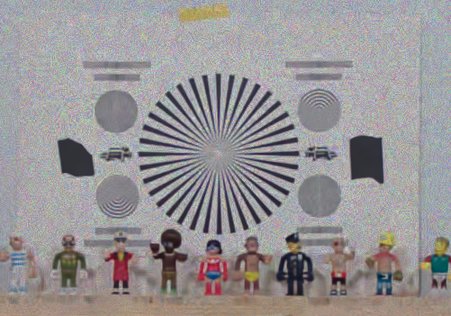} \vspace*{0.01mm}  \\
			(e) RF \cite{Tao2017} with BM3D \cite{Dabov2007} & 
			\includegraphics[width=\linewidth,align=c]{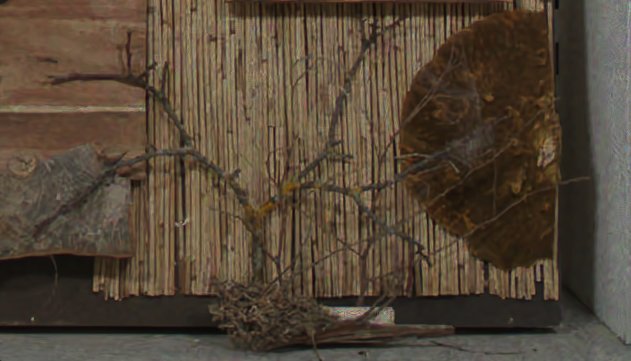}&
			\includegraphics[width=\linewidth,align=c]{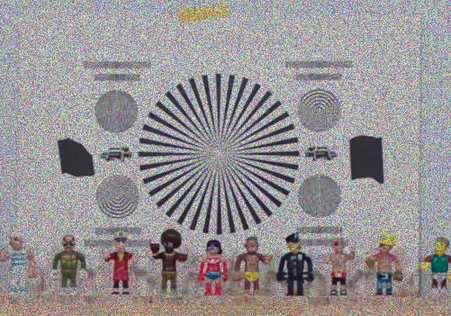} \vspace*{0.01mm} \\
			(e) DnCNN \cite{Zhang2017beyond}& 
			\includegraphics[width=\linewidth,align=c]{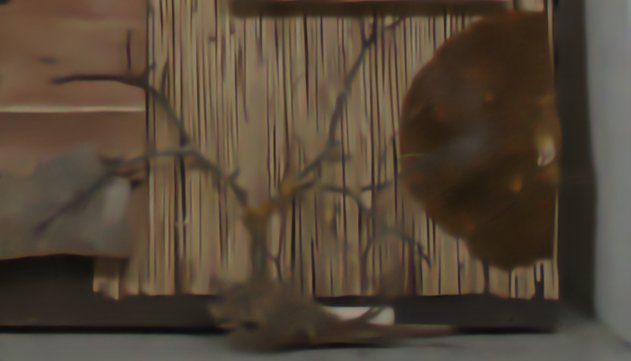}&
			\includegraphics[width=\linewidth,align=c]{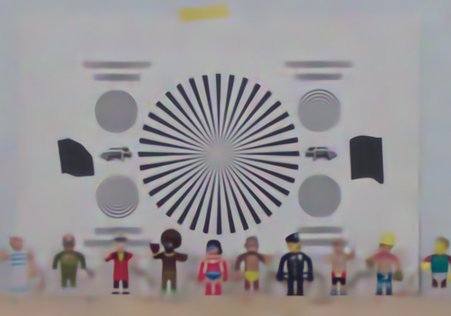} \vspace*{0.01mm} \\
			(f) QuaSI - MC \cite{Schirrmacher2018} & 
			\includegraphics[width=\linewidth,align=c]{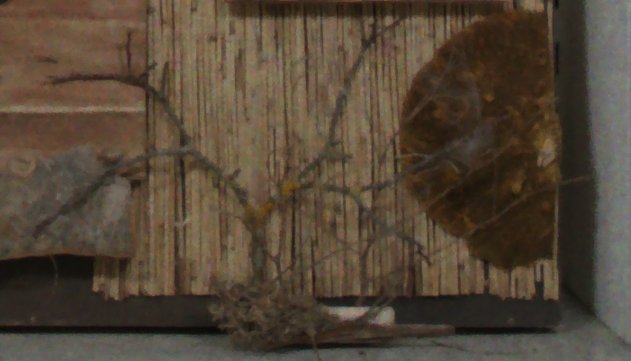}&
			\includegraphics[width=\linewidth,align=c]{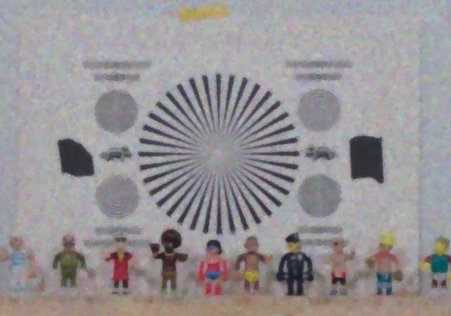} \vspace*{0.01mm}  \\
			(g) \aquasi\ - MC (dynamic guidance) & 
			\includegraphics[width=\linewidth,align=c]{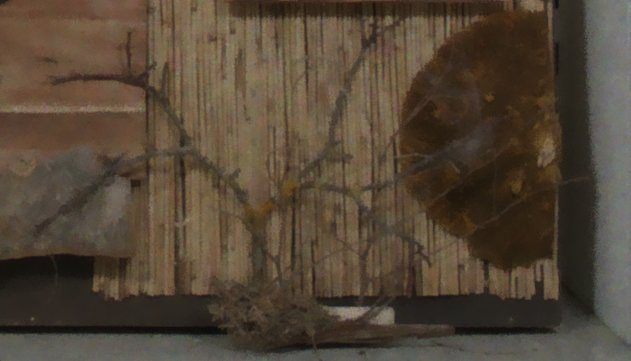}&
			\includegraphics[width=\linewidth,align=c]{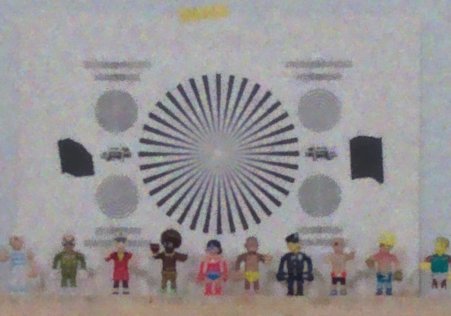} \vspace*{0.01mm} \\
		\end{tabular}
		\caption{RGB/NIR image restoration on image 10 of the ARRI dataset \cite{Luethen2017}. All competing methods perform denoising on the degraded color image only.}
		
	\end{figure}

	\begin{figure}[!tbp]
		\footnotesize
		
		\setlength{\tabcolsep}{1.5pt}
		\begin{tabular}{>{\centering\arraybackslash}p{0.24\linewidth} p{0.24\linewidth} p{0.24\linewidth}}
			(a) Ground truth & 
			\includegraphics[width=\linewidth,align=c]{RGB_flower.jpg}&
			\includegraphics[width=\linewidth,align=c]{RGB_ground.jpg} \vspace*{0.01mm}  \\
			(b) Noisy RGB image & 
			\includegraphics[width=\linewidth,align=c]{noisy_flower.jpg}&
			\includegraphics[width=\linewidth,align=c]{noisy_ground.jpg} \vspace*{0.01mm}  \\
			(c) RED with BM3D \cite{Dabov2007} & 
			\includegraphics[width=\linewidth,align=c]{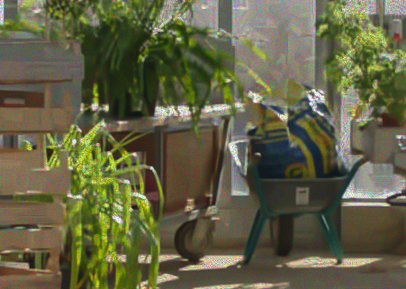} &
			\includegraphics[width=\linewidth,align=c]{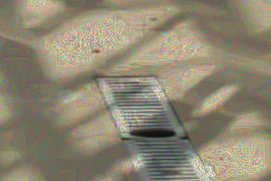} \vspace*{0.01mm} \\
			(d) RED with DnCNN \cite{Zhang2017beyond} & 
			\includegraphics[width=\linewidth,align=c]{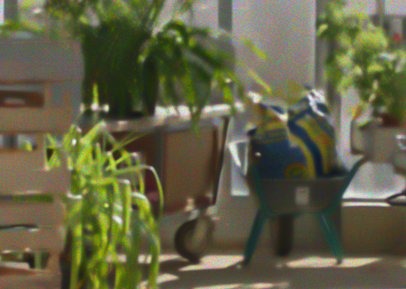} &
			\includegraphics[width=\linewidth,align=c]{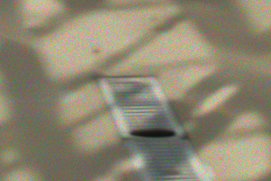} \vspace*{0.01mm} \\
			(e) RED with NLM \cite{Buades2005}& 
			\includegraphics[width=\linewidth,align=c]{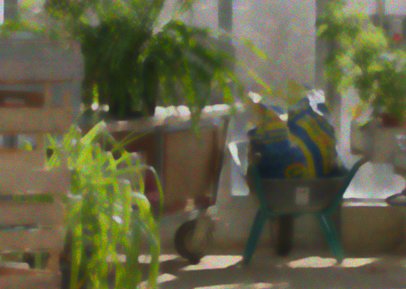} &
			\includegraphics[width=\linewidth,align=c]{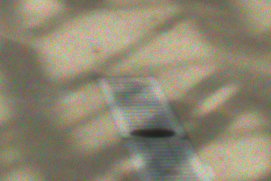} \vspace*{0.01mm} \\
		\end{tabular}
		\caption{RGB/NIR image restoration on image 11 of the ARRI dataset \cite{Luethen2017}. Additional experiments on 3 different denoiser in the RED framework, namely BM3D, DnCNN, and NLM.}
	\end{figure}

	\begin{figure}[!tbp]
		\footnotesize
		
		\setlength{\tabcolsep}{1.5pt}
		\begin{tabular}{>{\centering\arraybackslash}p{0.24\linewidth} p{0.24\linewidth} p{0.24\linewidth}}
			(a) Ground truth & 
			\includegraphics[width=\linewidth,align=c]{RGB_tree.jpg}&
			\includegraphics[width=\linewidth,align=c]{RGB_figures.jpg} \vspace*{0.01mm}  \\
			(b) Noisy RGB image & 
			\includegraphics[width=\linewidth,align=c]{noisy_tree.jpg}&
			\includegraphics[width=\linewidth,align=c]{noisy_figures.jpg} \vspace*{0.01mm}  \\
			(c) RED~\cite{Romano2016a} with BM3D \cite{Dabov2007} & 
			\includegraphics[width=\linewidth,align=c]{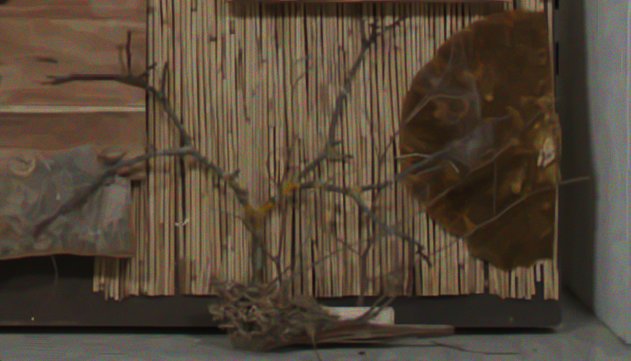} &
			\includegraphics[width=\linewidth,align=c]{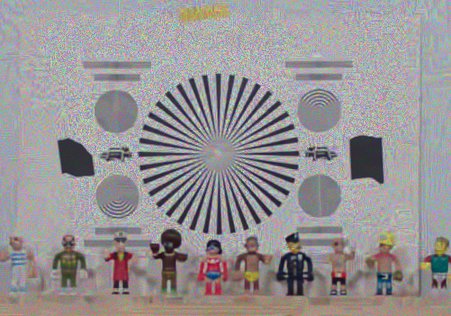} \vspace*{0.01mm} \\
			(d) RED~\cite{Romano2016a} with DnCNN \cite{Zhang2017beyond} & 
			\includegraphics[width=\linewidth,align=c]{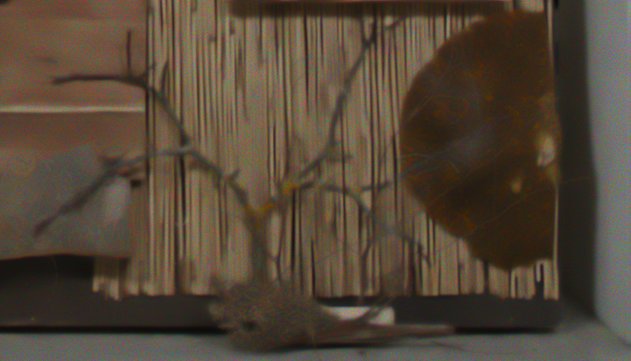} &
			\includegraphics[width=\linewidth,align=c]{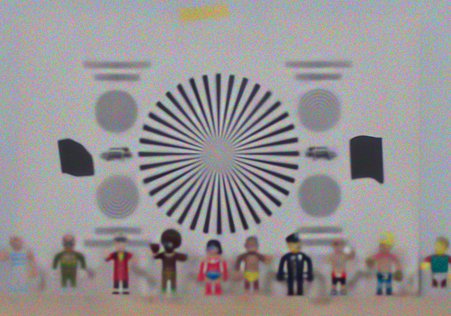} \vspace*{0.01mm} \\
			(e) RED~\cite{Romano2016a} with NLM \cite{Buades2005}& 
			\includegraphics[width=\linewidth,align=c]{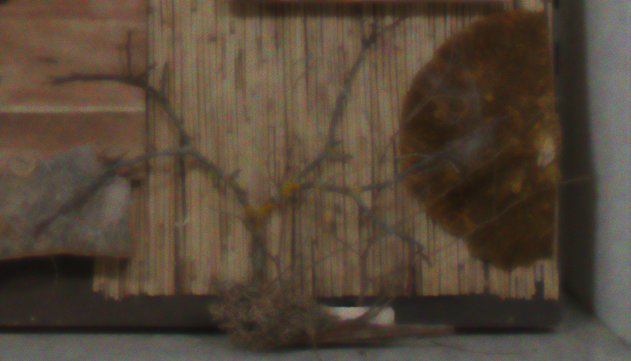} &
			\includegraphics[width=\linewidth,align=c]{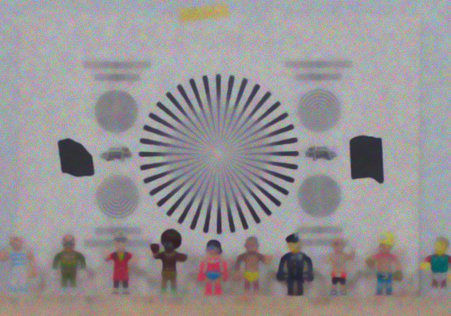} \vspace*{0.01mm} \\
		\end{tabular}
		\caption{RGB/NIR image restoration on image 11 of the ARRI dataset \cite{Luethen2017}. Additional experiments on 3 different denoiser in the RED framework, namely BM3D, DnCNN, and NLM.}
	\end{figure}

	\begin{table}[!tbp]
		
		\centering
		\caption{Averaged PSNR measures of all competing methods on the ARRI dataset. The quantitative results are calculated on the color images.}
		\begin{tabular}{p{5cm} p{1.1cm}c p{1.0cm}c }
			\toprule
			Method												& PSNR  					 \\
			\midrule
			noisy & 18.34  \\	
			CBM3D~\cite{Dabov2007a} & 26.99  \\
			DnCNN\cite{Zhang2017beyond} & 31.08 \\
			Guided filter~\cite{He2013} & 29.89  \\
			QuaSI - MC~\cite{Schirrmacher2018} & 29.37\\
			QuaSI - SC~\cite{Schirrmacher2018} & 27.70  \\
			RED~\cite{Romano2016a} with TNRD~\cite{Chen2017} & 25.98  \\
			RED~\cite{Romano2016a} with BM3D \cite{Dabov2007} & 19.35  \\
			RED~\cite{Romano2016a} with DnCNN ~\cite{Zhang2017beyond} & 19.47  \\
			RED~\cite{Romano2016a} with NLM~\cite{Buades2005} & 19.31\\
			RED~\cite{Romano2016a} with WMF~\cite{Zhang2014} & 21.31  \\
			RF~\cite{Tao2017} with AM & 29.78  \\
			RF~\cite{Tao2017} with BM3D \cite{Dabov2007} & 19.69  \\
			RF~\cite{Tao2017} with GF~\cite{He2013} & 28.59 \\
			RF~\cite{Tao2017} with MF & 13.29 \\
			RF~\cite{Tao2017} with WMF~\cite{Zhang2014} & 22.22 \\
			Scale map ~\cite{Yan2013}& 27.13  \\
			WNN \cite{Gu2014} & 22.97 \\
			\aquasi\ - SC (static guidance) & 30.43  \\
			\aquasi\ - SC (dynamic guidance) & 22.46 \\
			\aquasi\ - MC (static guidance) & 29.99  \\
			\aquasi\ - MC (dynamic guidance) & 29.79\\	
			\bottomrule
		\end{tabular}
		\label{tab:RMSE}
	\end{table}

	\begin{figure}
		\centering
		\subfloat[TV ($\mu = 0.1$)]{\includegraphics[width=0.35\linewidth]{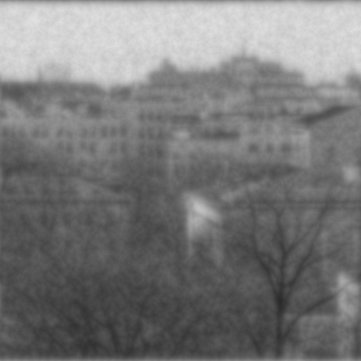}}\
		\subfloat[AQuaSI ($\lambda = 100$) with TV ($\mu = 0.1$)]{\includegraphics[width=0.35\linewidth]{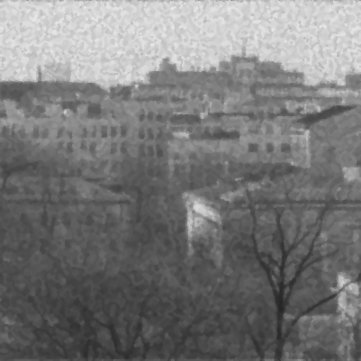}}
		
		\subfloat[TV ($\mu = 4$)]{\includegraphics[width=0.35\linewidth]{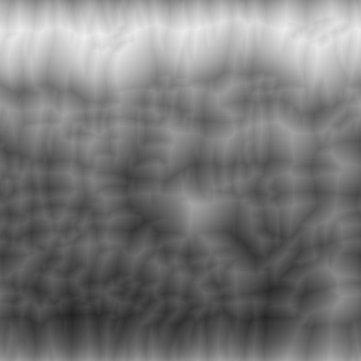}}\
		\subfloat[AQuaSI ($\lambda = 23000$) ]{\includegraphics[width=0.35\linewidth]{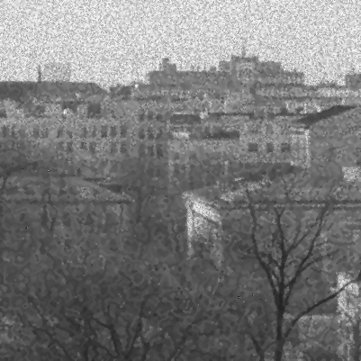}}
		\caption{Additional qualitative results on the comparison of AQuaSI with Total Variation regularization.}	
	\end{figure}

	\begin{figure}[!tbp]
		\centering
		\subfloat[Ground truth]
		{\begin{tikzpicture}[spy using outlines={rectangle,orange,magnification=3, 
				height=1cm, width = 1cm, connect spies, every spy on node/.append style={thick}}]
			\node {\pgfimage[width = 0.21\textwidth]{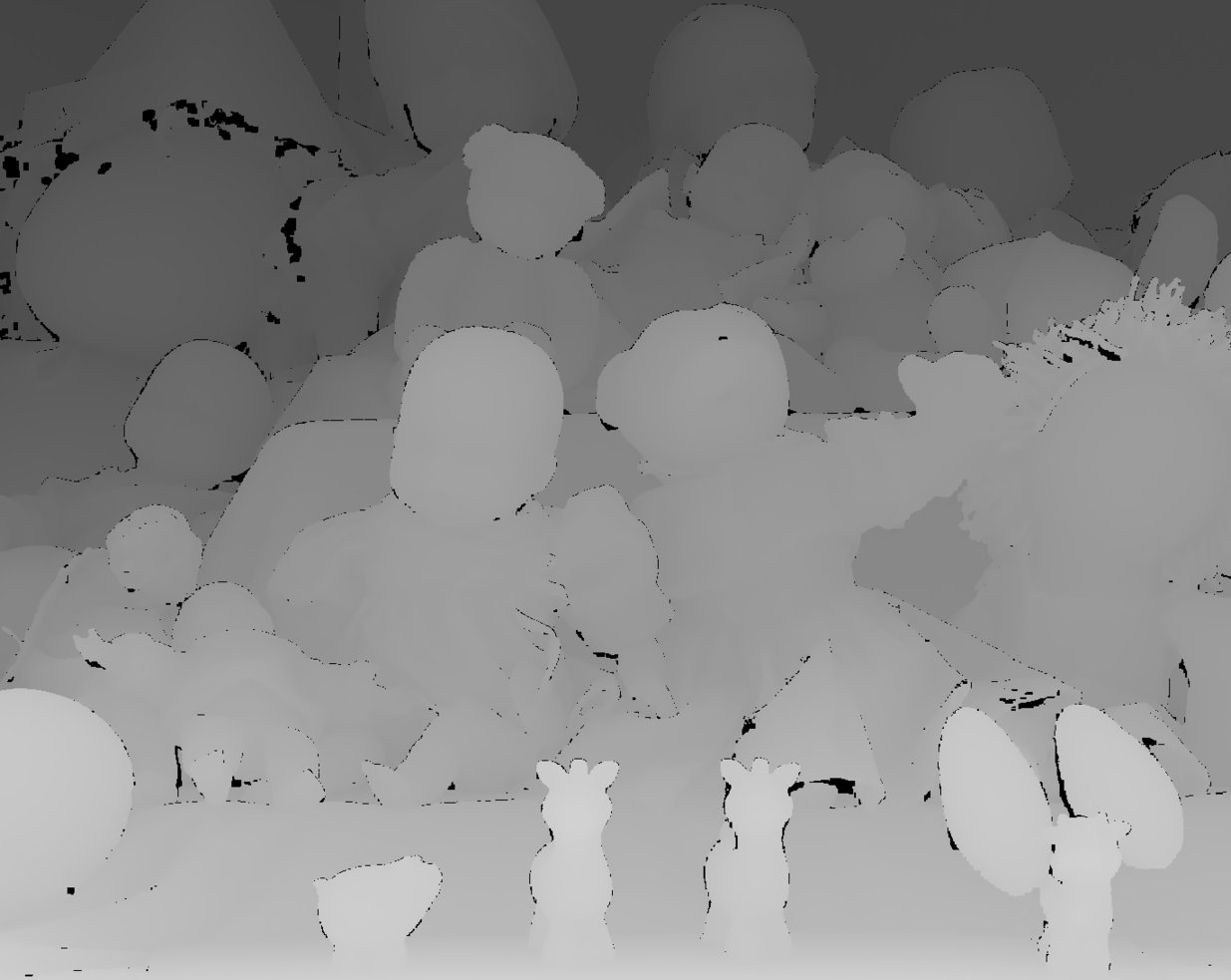}};
			\spy on (-0.1,-0.8) in node [left] at (-0.7, 0.9);		  
			\end{tikzpicture}\label{fig:jointUpsamplingGt}}
		\subfloat[Degraded depth map]
		{\begin{tikzpicture}[spy using outlines={rectangle,orange,magnification=3, 
				height=1cm, width = 1cm, connect spies, every spy on node/.append style={thick}}]
			\node {\pgfimage[width = 0.21\textwidth]{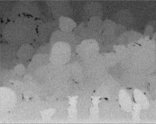}};
			\spy on (-0.1,-0.8) in node [left] at (-0.7, 0.9);		  
			\end{tikzpicture}\label{fig:jointUpsamplingNoisy}}
		\subfloat[Color image]	
		{\begin{tikzpicture}[spy using outlines={rectangle,orange,magnification=3, 
				height=1cm, width = 1cm, connect spies, every spy on node/.append style={thick}}]
			\node {\pgfimage[width = 0.21\textwidth]{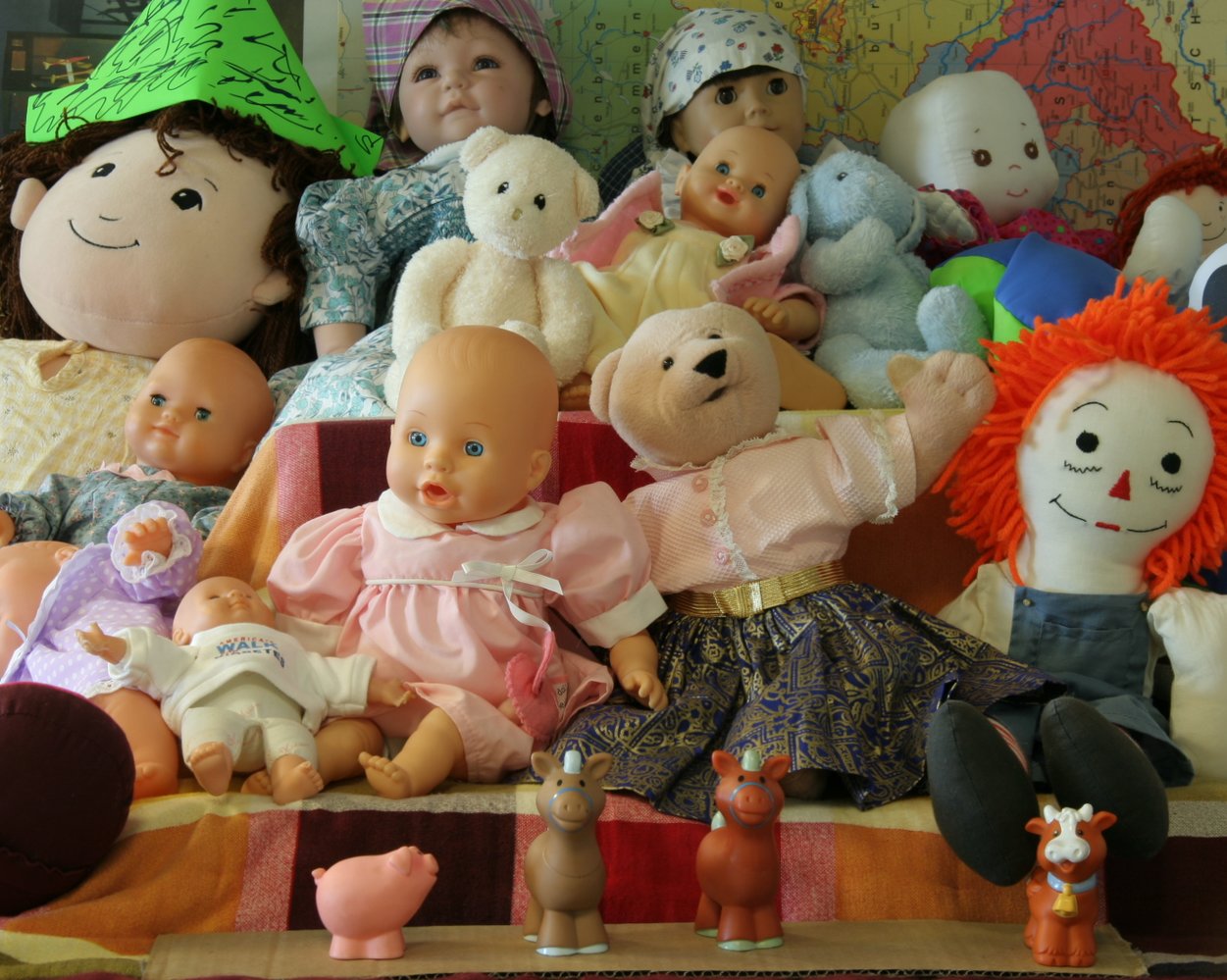}};
			\spy on (-0.1,-0.8) in node [left] at (-0.7, 0.9);
			\end{tikzpicture}\label{fig:jointUpsamplingRGB}}\\	
		\subfloat[MS \cite{Shen2015}\newline  (RMSE:~0.0215 BME: 0.51)]
		{\begin{tikzpicture}[spy using outlines={rectangle,orange,magnification=3, 
				height=1cm, width = 1cm, connect spies, every spy on node/.append style={thick}}]
			\node {\pgfimage[width = 0.21\textwidth]{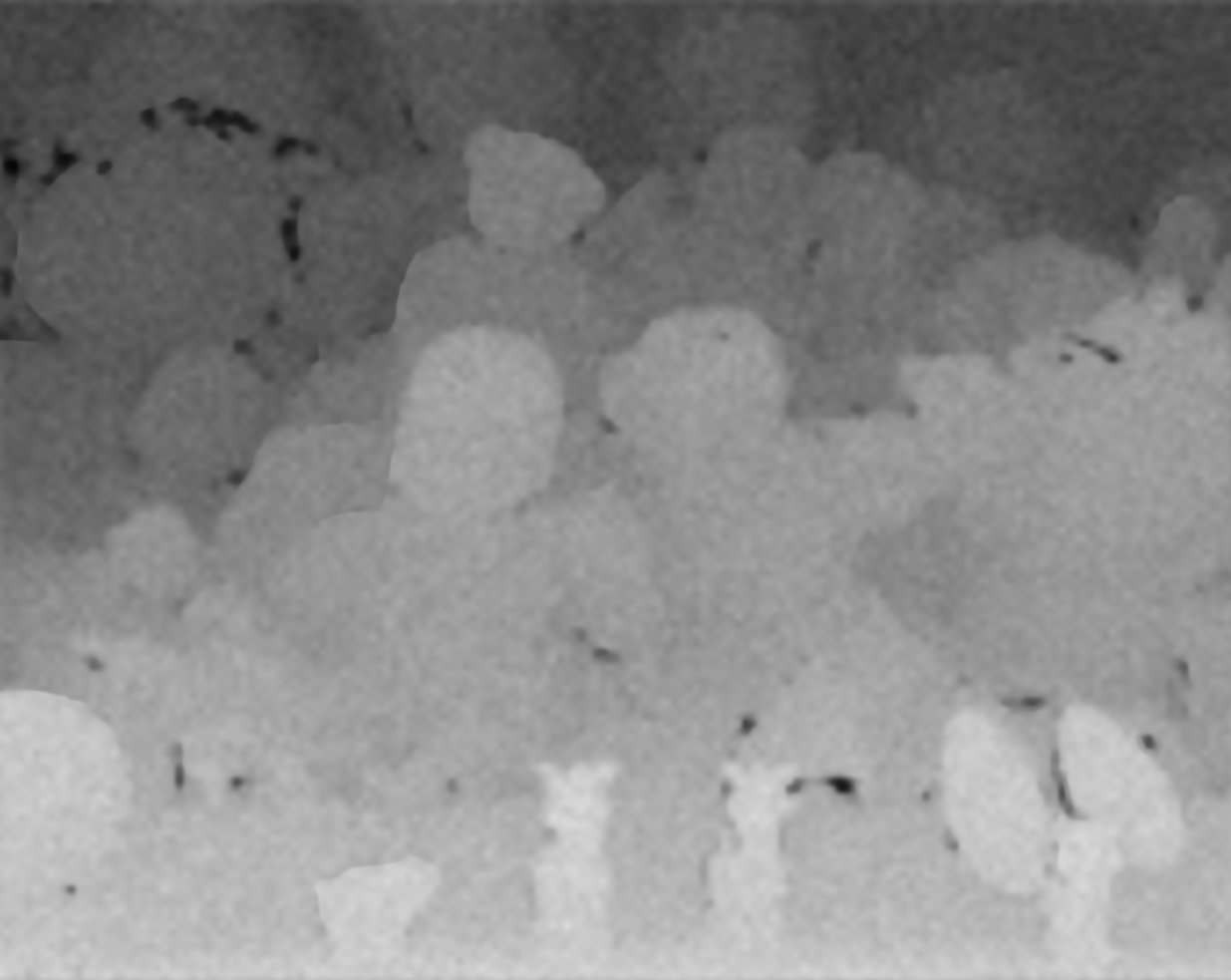}};
			\spy on (-0.1,-0.8) in node [left] at (-0.7, 0.9);
			\end{tikzpicture}\label{fig:jointUpsamplingMS}}
		\subfloat[TGVL2 \cite{Ferstl2013}\newline  (RMSE:~0.0190 BME: 0.47)]
		{\begin{tikzpicture}[spy using outlines={rectangle,orange,magnification=3, 
				height=1cm, width = 1cm, connect spies, every spy on node/.append style={thick}}]
			\node {\pgfimage[width = 0.21\textwidth]{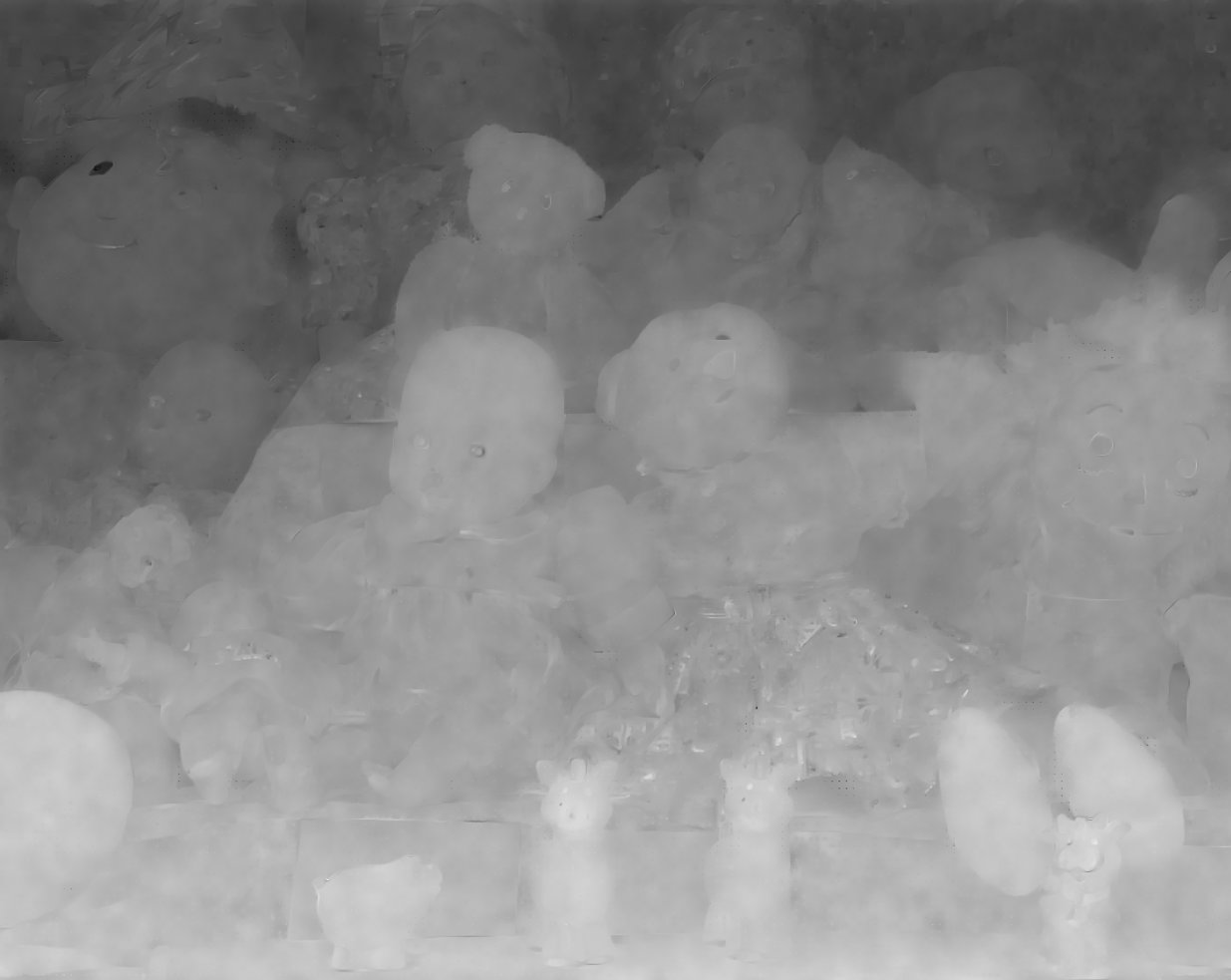}};
			\spy on (-0.1,-0.8) in node [left] at (-0.7, 0.9);  
			\end{tikzpicture}\label{fig:jointUpsamplingTGVL2}}
		\subfloat[DJF \cite{Li2016}\newline  (RMSE:~0.0239 BME: 0.57)]		
		{\begin{tikzpicture}[spy using outlines={rectangle,orange,magnification=3, 
				height=1cm, width = 1cm, connect spies, every spy on node/.append style={thick}}]
			\node {\pgfimage[width = 0.21\textwidth]{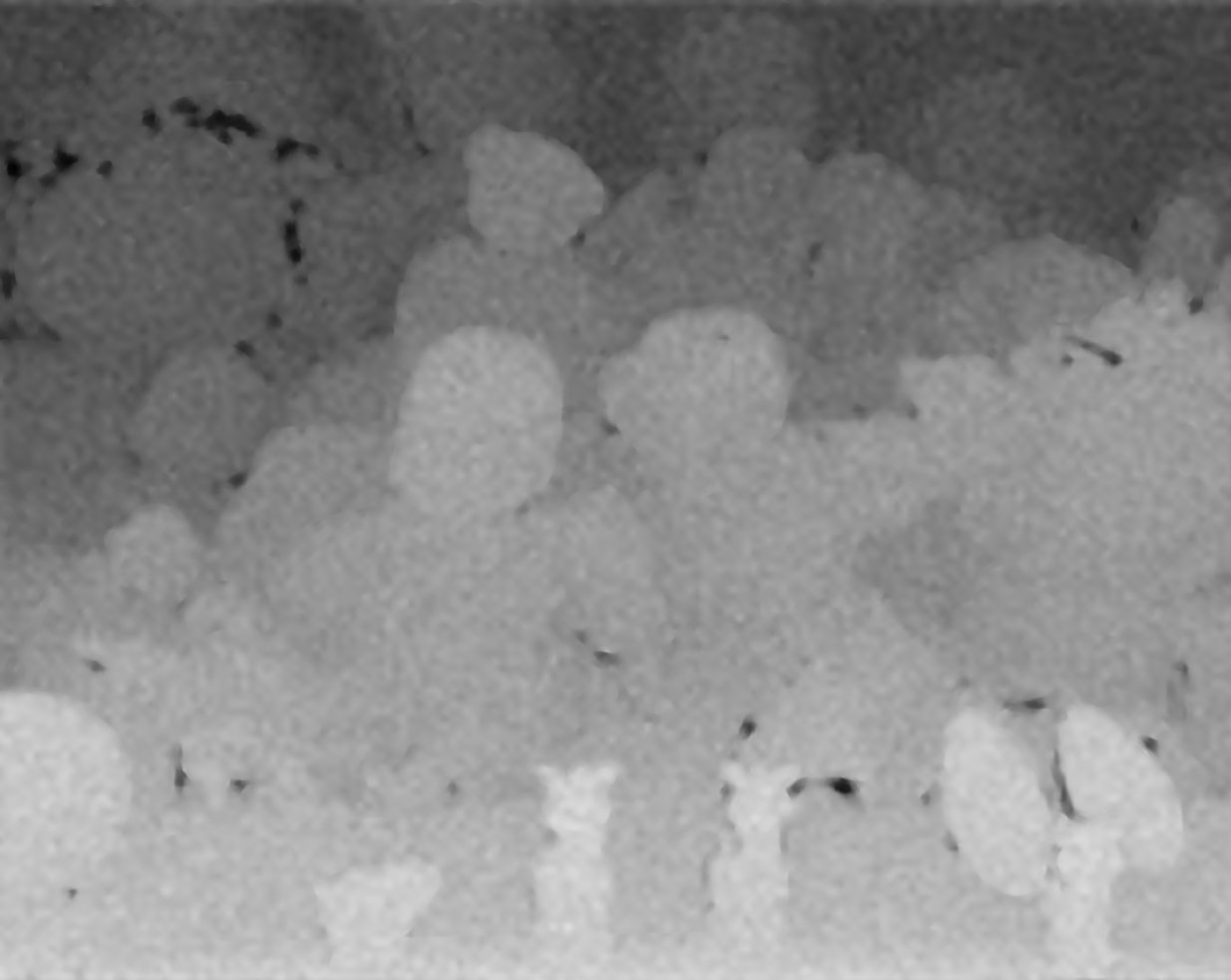}};
			\spy on (-0.1,-0.8) in node [left] at (-0.7, 0.9);		  
			\end{tikzpicture}\label{fig:jointUpsamplingDJF}}
		\subfloat[DJFR \cite{Li2019} \newline (RMSE:~0.0248 BME: 0.58)]
		{\begin{tikzpicture}[spy using outlines={rectangle,orange,magnification=3, 
				height=1cm, width = 1cm, connect spies, every spy on node/.append style={thick}}]
			\node {\pgfimage[width = 0.21\textwidth]{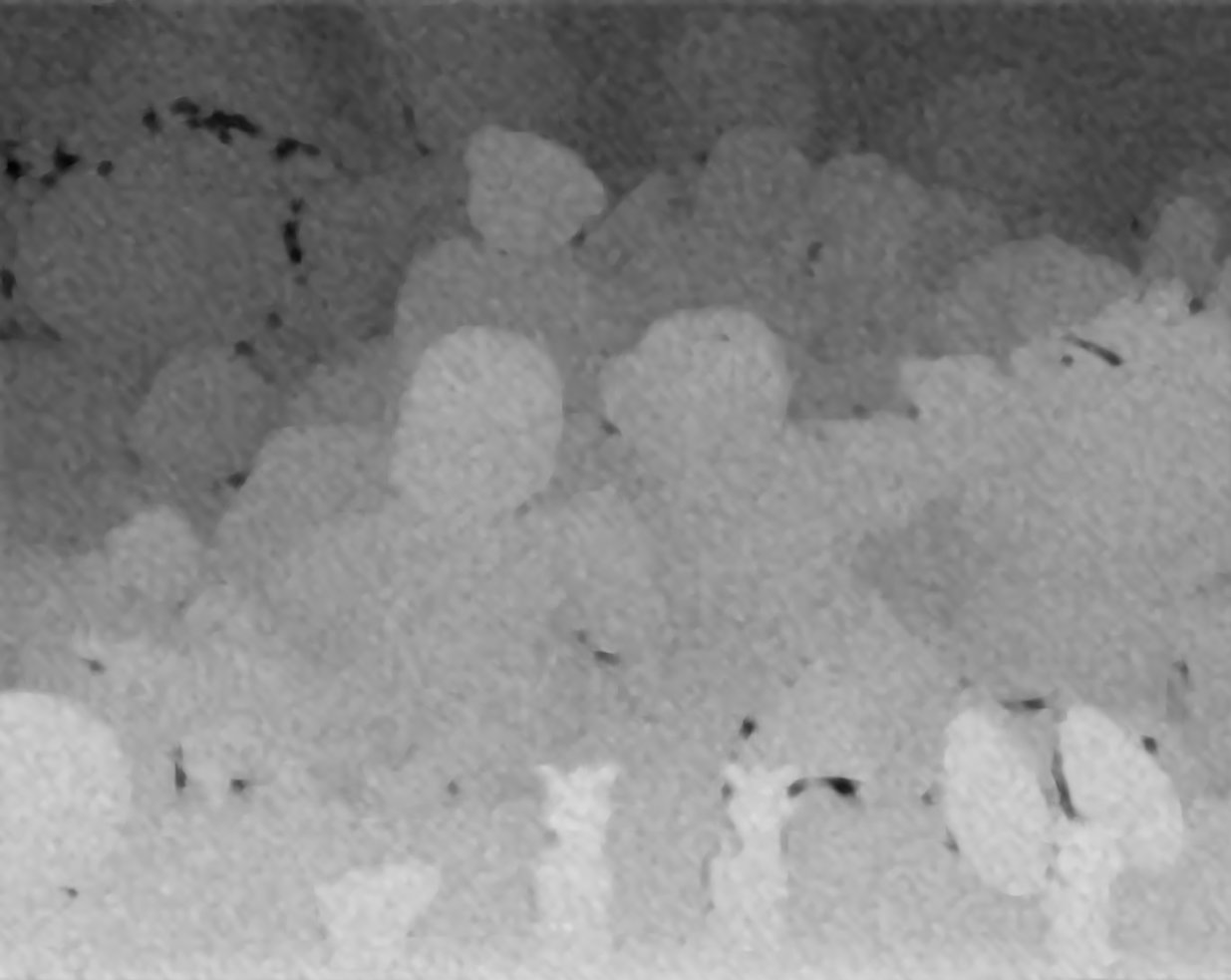}};
			\spy on (-0.1,-0.8) in node [left] at (-0.7, 0.9);			  
			\end{tikzpicture}\label{fig:jointUpsamplingDJFR}}\\
		\subfloat[SD~\cite{Ham2015} \newline w/ Welsch, w/o \aquasi \newline (RMSE:~0.0191 BME: 0.40)]
		{\begin{tikzpicture}[spy using outlines={rectangle,orange,magnification=3, 
				height=1cm, width = 1cm, connect spies, every spy on node/.append style={thick}}]
			\node {\pgfimage[width = 0.21\textwidth]{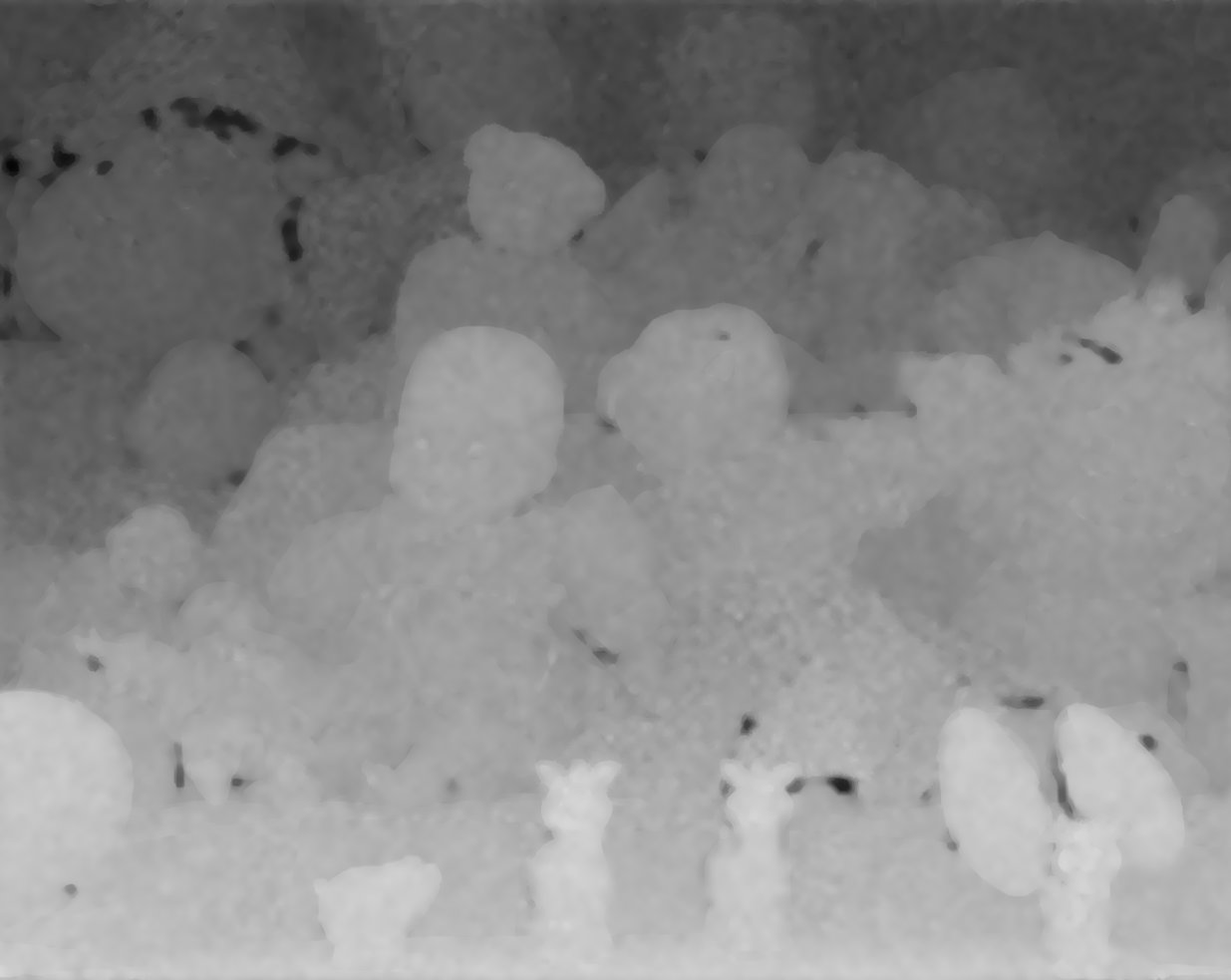}};
			\spy on (-0.1,-0.8) in node [left] at (-0.7, 0.9);		  
			\end{tikzpicture}\label{fig:jointUpsamplingSD}}	
		\subfloat[SD filter~\cite{Ham2015} \newline w/o Welsch, w/ \aquasi  \newline (RMSE:~0.0219 BME: 0.50)]		
		{\begin{tikzpicture}[spy using outlines={rectangle,orange,magnification=3, 
				height=1cm, width = 1cm, connect spies, every spy on node/.append style={thick}}]
			\node {\pgfimage[width = 0.21\textwidth]{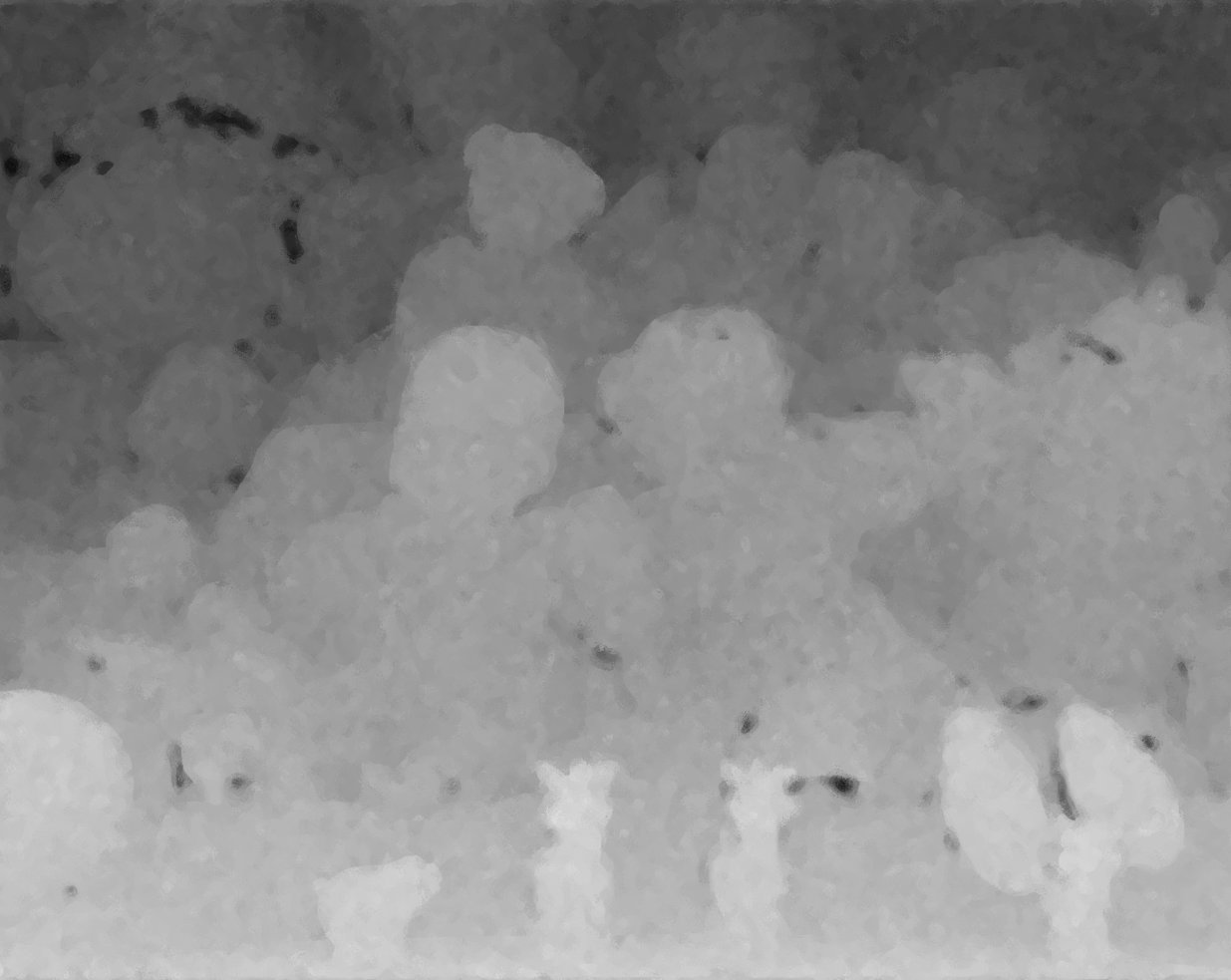}};
			\spy on (-0.1,-0.8) in node [left] at (-0.7, 0.9);		  
			\end{tikzpicture}\label{fig:jointUpsamplingAQuaSIonly}}
		\subfloat[SD filter~\cite{Ham2015}\newline w/ Welsch, w/ QuaSI \newline (RMSE:~0.0160 BME: 0.27)]		
		{\begin{tikzpicture}[spy using outlines={rectangle,orange,magnification=3, 
				height=1cm, width = 1cm, connect spies, every spy on node/.append style={thick}}]
			\node {\pgfimage[width = 0.21\textwidth]{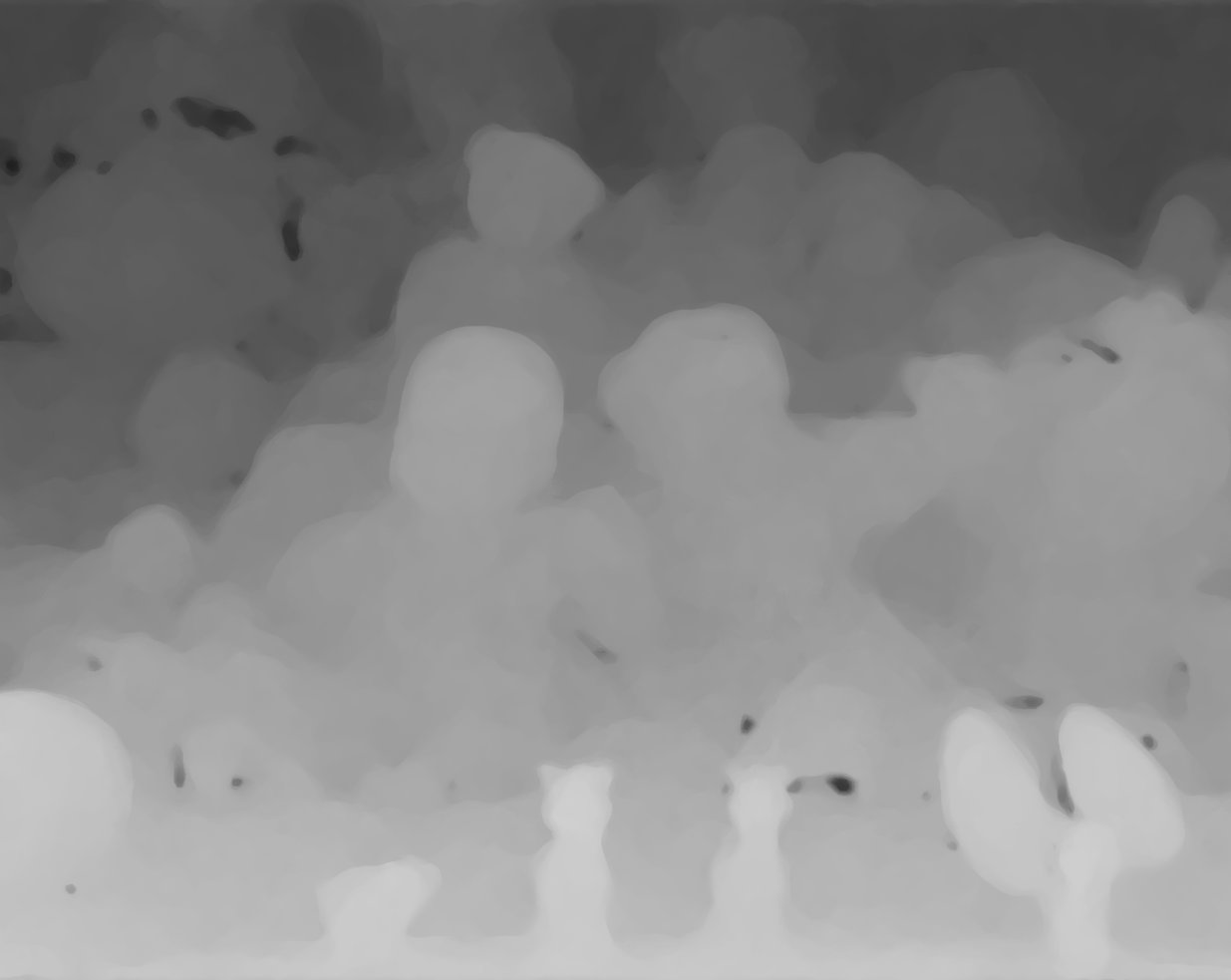}};
			\spy on (-0.1,-0.8) in node [left] at (-0.7, 0.9);		  
			\end{tikzpicture}\label{fig:jointUpsamplingQuaSI}}
		\subfloat[SD filter~\cite{Ham2015}\newline w/ Welsch, w/ \aquasi \newline (RMSE:~0.0166 BME: 0.28)]		
		{\begin{tikzpicture}[spy using outlines={rectangle,orange,magnification=3, 
				height=1cm, width = 1cm, connect spies, every spy on node/.append style={thick}}]
			\node {\pgfimage[width = 0.21\textwidth]{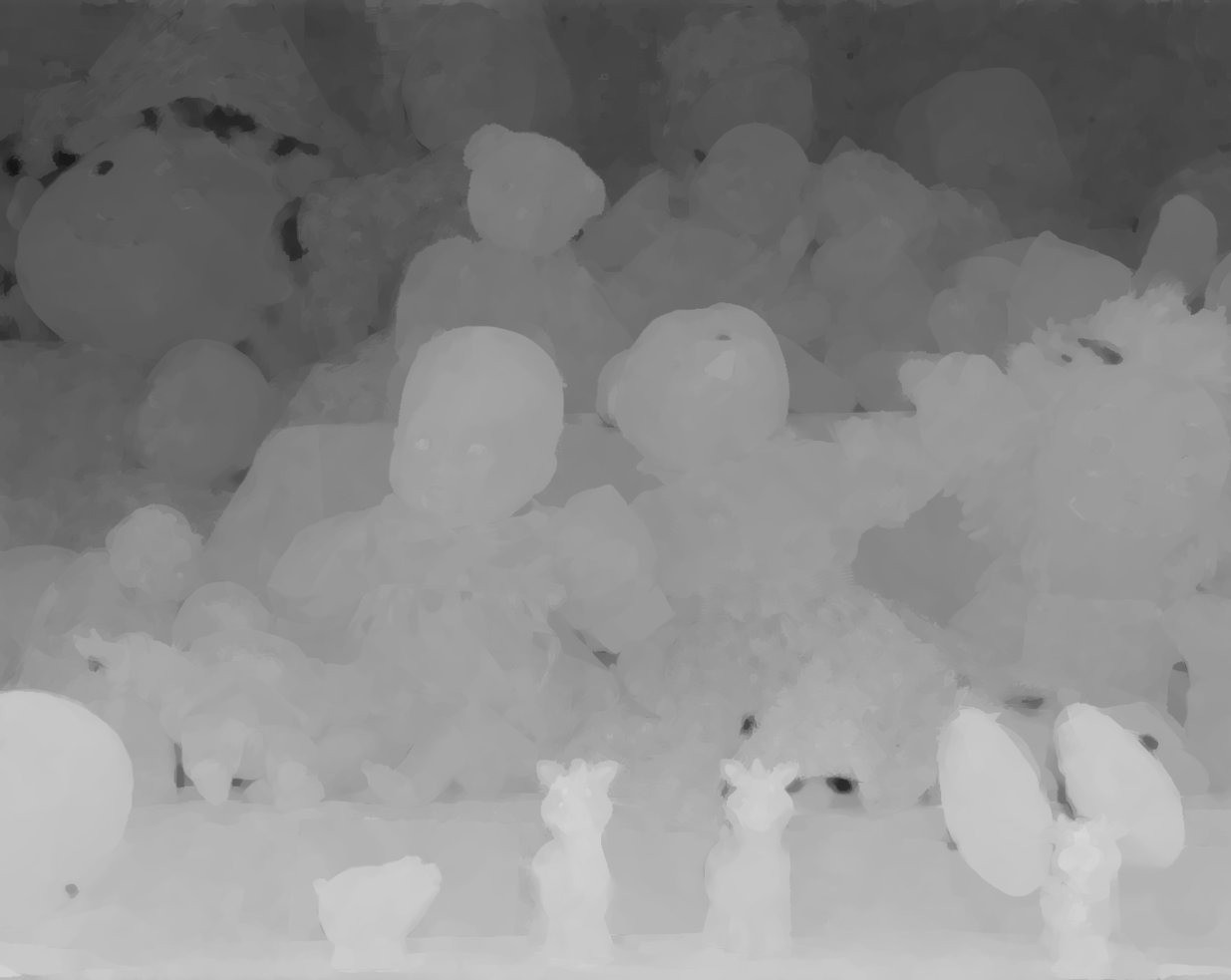}};
			\spy on (-0.1,-0.8) in node [left] at (-0.7, 0.9);		  
			\end{tikzpicture}\label{fig:jointUpsamplingAQuaSI}}
		
		\caption{Joint RGB/depth map upsampling on a Middlebury image pair \textit{Dolls}. \protect\subref{fig:jointUpsamplingGt} Ground truth depth map, \protect\subref{fig:jointUpsamplingNoisy} degraded depth map, \protect\subref{fig:jointUpsamplingRGB} color image, \protect\subref{fig:jointUpsamplingMS} - \protect\subref{fig:jointUpsamplingAQuaSI} mutual structure (MS) filter, TGVL2 upsampling, deep joint filter (DJF), residual-based deep joint filter (DJFR), SD filter w/ Welsch, w/o \aquasi, SD filter w/o Welsch, w/ \aquasi, SD filter w/ Welsch, w/ QuaSI, and, SD filter w/ Welsch, w/ \aquasi.   }
		\label{fig:jointUpsampling}
	\end{figure}

	\clearpage

	\bibliographystyle{IEEEtran}



	%
	
	%
	%
	%
	
	
	
	
\end{appendices}
	
\end{document}